\documentclass[pdflatex,sn-apa]{sn-jnl}

\usepackage{graphicx}%
\usepackage{multirow}%
\usepackage{amsmath,amssymb,amsfonts}%
\usepackage{amsthm}%
\usepackage{mathrsfs}%
\usepackage[title]{appendix}%
\usepackage{xcolor}%
\usepackage{textcomp}%
\usepackage{manyfoot}%
\usepackage{booktabs}%
\usepackage{algorithm}%
\usepackage{algorithmicx}%
\usepackage{algpseudocode}%
\usepackage{listings}%
\usepackage{url}
\usepackage{bm}

\renewcommand{\algorithmiccomment}[1]{\hfill$\triangleright$ #1} 

\theoremstyle{thmstyleone}%
\newtheorem{theorem}{Theorem}
\newtheorem{proposition}[theorem]{Proposition}%
\newtheorem{corollary}[theorem]{Corollary}

\theoremstyle{thmstyletwo}%
\newtheorem{remark}{Remark}%

\theoremstyle{thmstylethree}%

\usepackage{amsmath, amssymb}
\usepackage{xcolor}
\usepackage{url}
\usepackage{color, graphics, graphicx}
\usepackage{subfigure}

\begin{document}

\title{E-ROBOT: a dimension-free method for  robust statistics and machine learning via  Schrödinger bridge}


\author*[1]{\fnm{Davide} \sur{La Vecchia}}\email{davide.lavecchia@unige.ch}
\author[2]{\fnm{Hang} \sur{Liu}}\email{hliu01@ustc.edu.cn}


\affil*[1]{\orgdiv{Geneva School of Economics and Management}, \orgname{University of Geneva}, \orgaddress{\street{Bld. du Pont d'Arve}, \city{Geneva}, \postcode{CH-1211}, \state{Switzerland}}}

\affil[2]{\orgdiv{Department of Statistics and Finance, School of Management}, \orgname{University of Science and Technology of China}, \orgaddress{\street{Jinzhai Rd}, 
\city{Hefei}, \postcode{ 230026}, \state{Anhui Province}, \country{China}}}



\abstract{We propose the Entropic-regularized Robust Optimal Transport (E-ROBOT) framework, a novel method that combines the robustness of ROBOT with the computational and statistical benefits of entropic regularization. We show that, rooted in the Schr\"{o}dinger bridge problem theory, E-ROBOT defines the  robust Sinkhorn divergence $\overline{W}_{\varepsilon,\lambda}$, where the parameter $\lambda$ controls robustness and $\varepsilon$ governs the regularization strength. Letting $n\in \mathbb{N}$ denote the sample size, a central theoretical contribution is establishing that the sample complexity of $\overline{W}_{\varepsilon,\lambda}$ is $\mathcal{O}(n^{-1/2})$, thereby avoiding the curse of dimensionality that plagues standard ROBOT. This dimension-free property unlocks the use of  $\overline{W}_{\varepsilon,\lambda}$ as a loss function in large-dimensional statistical and machine learning tasks. With this regard, we demonstrate its utility through four applications: goodness-of-fit testing;  computation of barycenters for corrupted 2D and 3D shapes; definition of gradient flows; and image colour transfer. From the computation standpoint, a perk of our novel method is that it can be easily implemented by modifying existing (\texttt{Python}) routines. From the theoretical standpoint, our work opens the door to many research directions in statistics and machine learning:  we discuss some of them.}




\maketitle

\section{Introduction}

\subsection{Related work}

The notion of the Schr\"odinger bridge problem (SBP) originates in~\cite{schrodinger1931}, when Schr\"odinger  investigated the most likely evolution of a cloud of independent Brownian particles subject to boundary conditions.

Intuitively, the SBP searches for the most likely joint distribution of initial and final cloud of particles that is consistent with the observed endpoint results (namely, the original and target measures). This entropy-minimization problem exhibits a profound unity across seemingly distinct fields, including statistical and quantum physics, probability, information theory, statistics, machine learning, and artificial intelligence. We refer the reader to~\cite{RigolletWeeds2018,CP19,wang2021deep,liu2022deep,bunne2023schrodinger,pooladian2025plug} and references therein.

One of the main attractive features of the SBP is its link to optimal transport (OT) problem; see \cite{leonard2014survey}. This connection yields two significant advantages. The first advantage is theoretical: the sample complexity of the Sinkhorn divergence, which results from the combination of SBP and OT, scales at a rate better than that of typical unregularized OT distances, which suffer from the curse of dimensionality. Essentially, the problem with standard OT is that Wasserstein distance computed between two samples converges very slowly to its population counterpart;
see~\cite{genevay2019sample}. Moreover,~\cite{feydy2019} show that entropic-regularized OT (E-OT) interpolates between the Wasserstein distance and the Maximum Mean Discrepancy (MMD, see e.g. \cite{sriperumbudur2011universality}). Specifically, it preserves the appealing geometric properties of OT losses, and, at the same time, it benefits from the low sample complexity of MMD norms. The second advantage is computational: the Sinkhorn algorithm accelerates the computation of an approximate transport plan, significantly expanding the range of OT applications, particularly in machine learning; see~\cite{cuturi2013,CP19}.

However, a well-known limitation of both OT and E-OT is their sensitivity to anomalous records and their requirement of finite moments---both issues stem from the use of an unbounded cost function in the transportation cost definition; see e.g.~\cite{mukherjee2021outlier,NGC22,Maetal25}. To address these problems,~\cite{Maetal25} introduce robust optimal transport (ROBOT) and study its statistical properties, proving its robustness to outliers. Despite these advantages and its good performance in many statistical and machine learning tasks, ROBOT still suffers from the curse of dimensionality and exhibits multi-scale behavior.  Similarly to OT-based distances (see e.g.~\cite{weed2019sharp}), the sample complexity of ROBOT-based distances depends on the data dimension; see Theorem 10 in~\cite{Maetal25}. Given the widespread use of OT in statistics, time series analysis, and machine learning (see e.g.~\cite{CP19,H22,la2022some,hallin2023rank,hallin2022cent, HallinLiu2024}), this limitation significantly restricts the applicability of ROBOT-based methods.

In the OT literature, solutions exist to mitigate the curse of dimensionality. For instance, the sliced-Wasserstein metric projects higher-dimensional distributions into one-dimensional representations and computes the distance as a functional of the Wasserstein distances between these projections; see e.g.~\cite{kolouri2017optimal,CP19} for a review. Alternatively, Gaussian-smoothed OT applies isotropic Gaussian smoothing to the original measures, alleviating the curse of dimensionality while preserving structural properties of the Wasserstein distance; see~\cite{goldfeld2020gaussian,nietert2021smooth}. Unfortunately, to the best of our knowledge, no robustness guarantees are available for these methods: although applicable in high dimensions, they still rely on unbounded cost functions and thus remain sensitive to outliers. This limits their applicability in the presence of anomalous records or when distributions lack finite moments.

\subsection{Our contributions and a preview of some results} \label{Sec:Preview}

In this paper, we address both the dimensionality and robustness issues of OT: we propose a novel method that simultaneously handles both challenges. Our approach thus bridges a critical gap in the OT literature within machine learning, while also providing a significant contribution to multivariate and robust statistics. We consider both theoretical and computational aspects and we illustrate the ease-of-implementation of our methodology. By building on the existing OT literature, our theoretical developments yield the needed guarantees, ensuring that the method's robustness and scalability are well-principled and not merely empirical.

To help the reader navigating through the paper, below we provide an overview of our contributions---we flag that Appendix A and Appendix B, available in the \textcolor{blue}{Supplementary Material}, contain all proofs and additional numerical results. 

In \S\ref{Sec2} and \S\ref{Sec:Main}, building on theory of SBP, we introduce the entropic-regularized robust optimal transport (E-ROBOT) framework, which combines the robustness of ROBOT with the computational and statistical benefits of entropic regularization. In \S\ref{Sec:Main}, we derive key theoretical and methodological aspects of E-ROBOT. Specifically, we provide the functional form of its potentials and their properties (Propositions~\ref{Prop1},~\ref{lemma2},~\ref{LemmaUCP}), its dual formulation (Proposition~\ref{PropDual}), and the convergence behavior of the transport plan as the regularization term vanishes (Proposition~\ref{conv1}) and as the sample size increases (Proposition~\ref{propD}). We show how E-ROBOT defines a truncated robust Sinkhorn divergence ($\overline{W}_{\varepsilon,\lambda}$) that metrizes convergence in law (Proposition~\ref{propopMetrics}) and how the Schr\"odinger potentials define a Bregman-type divergence (Proposition~\ref{GradEROBOT}). One of our main contributions is Theorem~\ref{Thm1} and Corollary~\ref{corollary}, where we derive the sample complexity of the robust Sinkhorn divergence and show that it achieves a dimension-free rate, similar to non-robust E-OT. Finally, we justify the use of E-ROBOT as a loss function for statistical inference and machine learning tasks (Proposition~\ref{PropDec}). While some results follow directly from existing SBP and OT theory, others require careful adaptation to the E-ROBOT setting, revealing interesting theoretical implications, such as those related to new MMDs discussed after Proposition~\ref{propopMetrics} and after Corollary~\ref{corollary} in Remark~\ref{Remark}.

In \S\ref{SecNum}, we illustrate the applicability and performance of E-ROBOT in various settings and tasks in statistics and machine learning. Readers primarily interested in computational aspects may proceed directly to that section. Our experiments demonstrate that implementing our method requires some modifications of existing \texttt{Python} libraries and \texttt{R} routines. Indeed, one needs to replace the unbounded cost matrix used in standard OT with the bounded cost matrix from~\eqref{Eq.newcost}. This yields ready-to-use tools that are reliable in the presence of outliers and in high dimensions. Furthermore, our experiments show that E-ROBOT-based procedures enable inference in large-dimensional settings where the underlying distribution lacks finite moments---even without outliers---such as multivariate $t$-distributions with small degrees of freedom ($df$). This is not possible with standard OT and E-OT based methods like those relying on the $p$-Wasserstein distance $W_p$, which all require finite moments of order $p \geq 1$.

In \S\ref{conclusion}, we discuss how this work lays the foundation for several promising research avenues. These include new theoretical developments, such as the derivation of parametric inference based on our robust divergence, as well as methodological ones, such as the joint selection of the hyperparameters $\lambda$ and $\varepsilon$. We regard these developments as natural and essential extensions of the framework presented here, though their detailed treatment constitutes a separate research program. The primary goal of this paper is to establish the theoretical and practical foundation of the E-ROBOT framework itself, thereby demonstrating its viability and advantages over existing methods.

To illustrate some of the advantages of E-ROBOT and the shortcomings of existing methods, we preview some of our results. Figure~\ref{Figtdist} shows outcomes for a (simple) hypothesis test of distribution equality; see~\cite{hallin2021multivariate} for background. Testing distribution equality is relevant in statistics (e.g., goodness-of-fit) and machine learning (e.g., generative model training). We compare our $\overline{W}_{\varepsilon,\lambda}$ (continuous blue curve) with the $W_1$-based test (dotted red curve) from~\cite{hallin2021multivariate}; see \S\ref{GoF} for details. To study the effect of moment existence and dimensionality, we use a large-dimensional setting: for a sample size $n=50$, we generate samples from a $d$-variate $t$-distribution, with $d=50$, and we consider different $df$. We fix the level at $5\%$: the results show that the $W_1$-based test lacks power for $df = 1, 2$, gains some power at $df = 3$ (where first and second moments are finite), but is consistently outperformed by our test across all $df$ values and alternatives. This demonstrates that standard procedures can struggle with moment requirements and dimensionality even in a simple setting (simple hypothesis and  no outliers). 

In the following pages, we detail the construction of the E-ROBOT framework and provide its theoretical underpinnings. We then illustrate its application not only in the same simple hypothesis testing problem of Figure~\ref{Figtdist},  but also in more complex tasks, such as barycenters computation in 2D and 3D, gradient flows, and image color transfer.

\begin{figure}[!h]
    \centering
    \begin{tabular}{ccc}  
        \includegraphics[width=0.31\textwidth, height=0.375\textwidth]{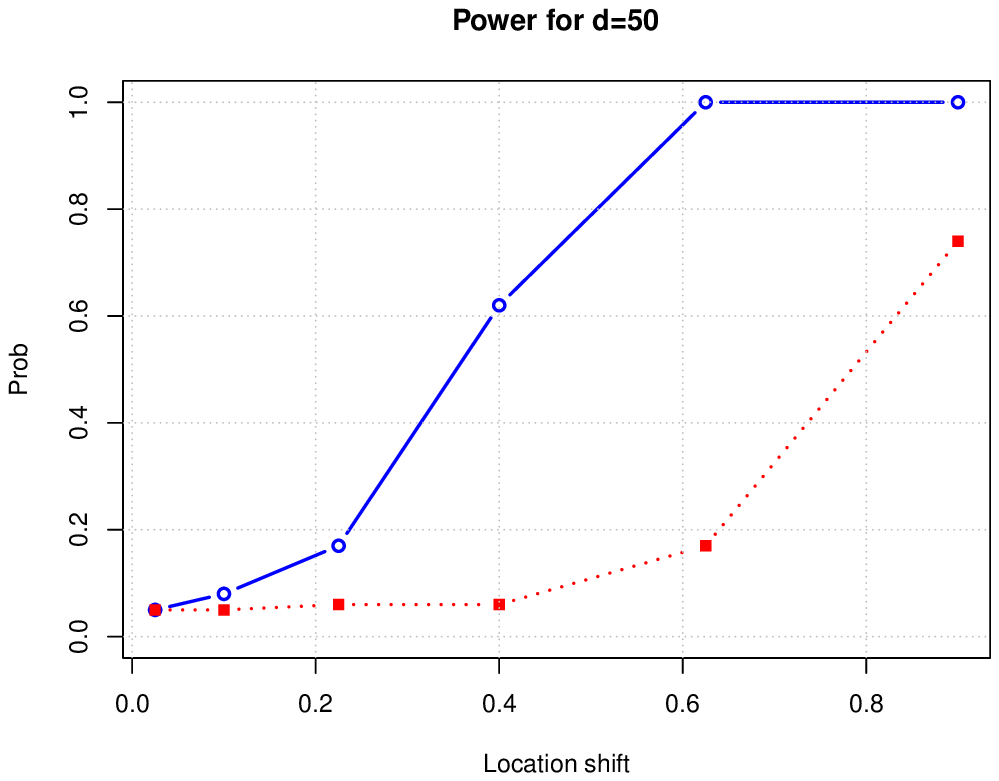} &
        \includegraphics[width=0.31\textwidth, height=0.375\textwidth]{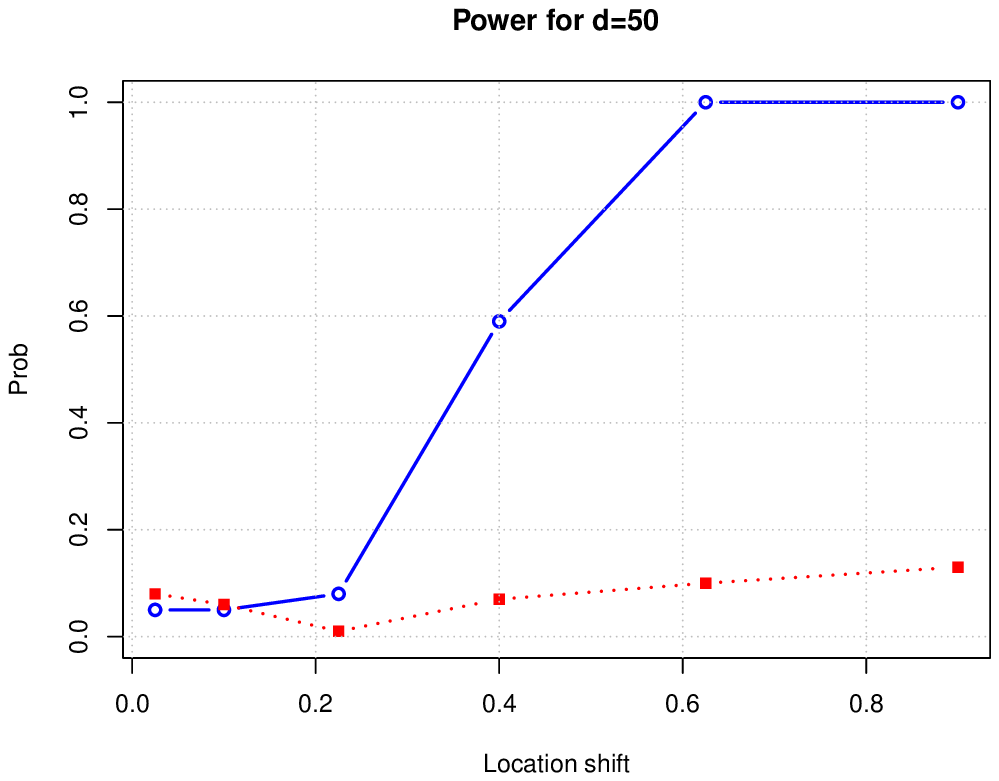} &
        \includegraphics[width=0.31\textwidth, height=0.375\textwidth]{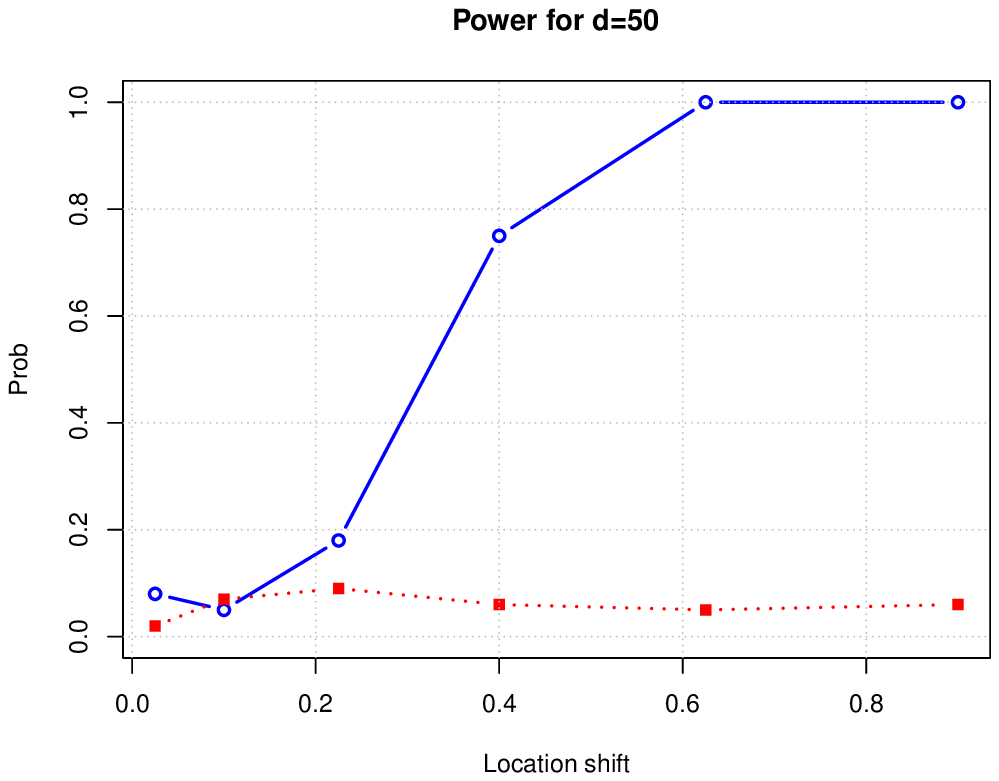}
    \end{tabular} 
    \caption{Power curves for testing the null hypothesis of a $d$-variate $t$-distribution with different degrees-of-freedom ($df$), location zero, and identity scale matrix. The continuous (blue) 
    curve with circles corresponds to our new test statistics based on $\overline{W}_{\varepsilon,\lambda}$, with $\varepsilon=0.05$ and $\lambda=3$; the dotted (red) 
    curve with squares corresponds to the test statistics based on $W_1$. Left plot is for $df=3$, middle plot for $df=2$, right plot for $df=1$.}
    \label{Figtdist}
\end{figure}


%
%

\section{SBP, ROBOT, and E-ROBOT} \label{Sec2}

In \S \ref{SBP} and \S \ref{ROBOT} we recall the key theoretical aspects of SBP  and ROBOT that are needed for our developments; we refer to  
 \cite{leonard2014survey} and to \cite{Maetal25} for more details.
Then, in \S \ref{Sec:EROBOT} we explain how SBP and ROBOT can be blended  to obtain the E-ROBOT.

\subsection{SBP } \label{SBP}

Let $(X, \mu)$ and $(Y, \nu)$ be separable probability spaces, and let $\mathcal{P}(X)$ denote the set of all probability measures on $X$. Let $R \in \mathcal{P}(X \times Y)$ be a reference measure. The goal is to find a coupling $\pi^* \in \Pi(\mu, \nu)$ minimizing the relative entropy with respect to $R$:
\begin{equation}
\pi^* = \arg\min_{\pi \in \Pi(\mu, \nu)} H(\pi \| R), \label{SB}
\end{equation}
where $\Pi(\mu, \nu)$ is the set of couplings with fixed marginals $\mu$ and $\nu$ and \( H(\pi \| R) = \mathbb E_\pi[ \ln (\pi / R)] \) denotes the relative entropy (or Kullback-Leibler divergence) of \( \pi \) with respect to \( R \). If $\Pi_{\mathrm{fin}}(\mu, \nu) := \{\pi \in \Pi(\mu, \nu) : H(\pi \| R) < \infty\} \neq \emptyset$, then there exists a unique minimizer $\pi^*$. When $R $ is absolutely continuous with respect to $ \mu \otimes \nu$ (the product measure), the optimal $\pi^*$ satisfies
\begin{equation}
\frac{d\pi^*}{dR} = e^{\varphi^*(x) + \psi^*(y)} \quad R\text{-a.s.},
\end{equation}
for some measurable functions $\varphi^* : X \to \mathbb{R}$ and $\psi^* : Y \to \mathbb{R}$, called Schrödinger potentials. These are unique up to an additive constant. Moreover,
let $c : X \times Y \to (-\infty, \infty]$ be a function and  define 
\begin{equation}
e^{-c(x, y)} = \frac{dR}{d(\mu \otimes \nu)}(x, y).
\end{equation}
Then, the optimal potentials $(\varphi^*, \psi^*)$ solve the Schr\"odinger system:
\begin{align}
\varphi^*(x) &= -\ln \int_Y e^{\psi^*(y) - c(x, y)} \, \nu(dy), \label{SE1} \\
\psi^*(y) &= -\ln \int_X e^{\varphi^*(x) - c(x, y)} \, \mu(dx). \label{SE2}
\end{align}

We notice that the Schrödinger potentials $\phi^*, \psi^*$ satisfy the fixed-point equations in \eqref{SE1} and \eqref{SE2}, which correspond to a {SoftMax} (log-sum-exp) operation. We will come back to this aspect in \S \ref{Sec:EROBOT}.

\subsection{ROBOT} \label{ROBOT}

The ROBOT framework, as defined in \cite{mukherjee2021outlier}, modifies the classical Kantorovich formulation of optimal transport (OT) by incorporating robustness to outliers via a total variation (TV) regularization. The primal formulation introduces an auxiliary perturbation $s$ to the source measure $\mu$, leading to the constrained problem:
\begin{eqnarray*}
&	\min_{\pi, s} \int_X \int_Y c(x,y) \pi(x,y) dx dy + \lambda \|s\|_{\mathrm{TV}} \\
&	\quad \text{s.t.} \quad \int_Y \pi(x,y) dy = \mu(x) + s(x), \; \int_X \pi(x,y) dx = \nu(y), \; \int_X s(x)dx = 0,
\end{eqnarray*}
where $\lambda > 0$ controls the influence of outliers and $c(x,y)=d(x,y)$, with $d(x,y)=\Vert x-y\Vert$ being the Euclidean distance. This formulation effectively eliminates data points $x$ such that $\mu(x) + s(x) = 0$, identifying them as outliers. A computationally efficient equivalent formulation is given by:
\begin{equation*}
	\inf \left\{ \int_{{X} \times {Y}} c_{\lambda}(x,y) \, d\pi(x,y) : \pi \in \Pi(\mu,\nu) \right\},
\end{equation*}
{with } $c_{\lambda}(x,y) := \tilde{d}_{\lambda}(x,y) :=  \min\{ d(x,y), 2\lambda \}$, which replaces the original cost $c = d$ with a trimmed version $c_\lambda$ that bounds the transport cost and introduces robustness.  \cite{Maetal25} illustrate that the dual formulation of ROBOT, unlike classical OT, imposes a bounded range constraint on the Kantorovich potential $\psi$:
\begin{equation*}
	\sup \left\{ \int \psi \, d\mu - \int \psi \, d\nu : \psi \in \mathcal{C}_b({X}), \; |\psi(x) - \psi(y)| \leq d(x,y), \; \mathrm{range}(\psi) \leq 2\lambda \right\},
\end{equation*}
where for the sake of notation we omit the integration domain and $ \mathcal{C}_b$ denotes the set of bounded and continuous functions. Moreover, by noticing that $\tilde{d}_{\lambda}(x,y)$ is a metric, one can construct the {robust Wasserstein distance}:
\begin{equation}
	W^{(\lambda)}(\mu, \nu) := \inf_{\pi \in \Pi(\mu,\nu)} \int c_{\lambda}(x,y) d\pi(x,y), \label{EqWlambda}
\end{equation}
which is shown to be a proper metric on the space of probability measures $\mathcal{P}(X)$. Unlike standard Wasserstein distances, $W^{(\lambda)}$ remains well-defined for all probability measures, even without moment conditions, making it suitable for robust inference for heavy-tail data distributions. Additionally,
 $W^{(\lambda)}$ is continuous and  monotonically non-decreasing with respect to $\lambda \in [0,\infty)$, and, if $ W_1(\mu,\nu)  $ exists, 
$\lim_{\lambda \rightarrow \infty} W^{(\lambda)}(\mu,\nu) =W_1(\mu,\nu).$ For a Polish space   $ ({X },d) $,  if we suppose that $ \mu_k$, (resp.  $\nu_k$)  converges weakly to $ \mu$, (resp. $\nu$)  in $ \mathcal{P}(X) $ as $k \rightarrow \infty$, then $W^{(\lambda)}(\mu_k,\nu_k) \rightarrow W^{(\lambda)}(\mu,\nu)$. Finally, Theorem 10 in \cite{Maetal25}  proves that the sample complexity of $W^{(\lambda)}$ is:
\[
	\mathbb{E}[W^{(\lambda)}(\mu,\hat\mu_n)] = 
	\begin{cases}
		\mathcal{O}(n^{-1/2})
		 &\text{ when }d=1\enspace,\\
		\mathcal{O}(\frac{\ln n}{n^{1/2}}) 
		&\text{ when }d=2\enspace,\\
		\mathcal{O}(n^{-1/d})
		&\text{ when }d\geqslant 3, \enspace
	\end{cases}
\]
which implies that ROBOT suffers from the curse of dimensionality: for $d>2$ the dimension of the underlying space has an impact on the mean rate. 

\subsection{E-ROBOT} \label{Sec:EROBOT}

Let  the product measure \(P= \mu \otimes \nu \in \mathcal{P}(X \times Y) \) be the reference measure. Then the entropic optimal transport problem with regularization parameter \( \varepsilon > 0 \) is defined via the minimization over $\pi \in \Pi(\mu, \nu)$ of
\[
C_\varepsilon(\mu, \nu, c, \pi) :=  \int c(x, y) \, d\pi(x, y) + \varepsilon H(\pi \| P).
\]
To proceed further, let  us consider the  ROBOT cost function $c_\lambda$.
For $\varepsilon >0$, we define the rescaled and truncated cost function
\begin{equation}
 \varepsilon^{ -1} c_\lambda(x,y) = \varepsilon^{ -1} \min(d(x,y), 2\lambda), \label{Eq.newcost}
\end{equation} 
which allows 
to define the entropic regularized ROBOT (E-ROBOT) problem. Omitting the arguments for the sake of notation, we introduce the  truncated Laplacian kernel 
\begin{equation}
k_{\varepsilon,\lambda} =  e^{- \varepsilon^{-1} c_\lambda} \label{EqGibbs}
\end{equation}
 (more generally, this is a Gibbs kernel) and define the reference (Gibbs) joint distribution
\begin{equation*}
dR_\varepsilon := \frac{1}{\beta}k_{\varepsilon,\lambda} dP, \quad \text{where } \beta := \int k_{\varepsilon,\lambda} dP.
\end{equation*}
For any \( \pi \in \mathcal{P}(X \times Y) \) with  $\pi$ absolutely continuous w.r.t. $R_\varepsilon$, namely \( \pi \ll R_\varepsilon \), we have the equalities
\begin{eqnarray*}
H(\pi \| R_\varepsilon) &=& \int \ln\left (\frac{d\pi}{dR_\varepsilon} \right)  d\pi =  \int \ln \left(\frac{d\pi}{dP} \frac{d P}{dR_\varepsilon}  \right) d\pi \\
                &=& \int \ln \left(\frac{d\pi}{dP} \right) d\pi +  \int \ln \left(  \frac{d P}{dR_\varepsilon}  \right) d\pi \\
                &=&H(\pi \| P) + \ln  \beta + \int  \varepsilon^{-1} c_\lambda d 
\pi,
\end{eqnarray*}
so
\begin{equation}
C_\varepsilon(\mu, \nu, c_\lambda, \pi) =  \int c_\lambda \, d\pi + \varepsilon H(\pi \| P) = \varepsilon H(\pi \| R_\varepsilon) -  \varepsilon \ln \beta.
\label{CepsLam}
\end{equation}
The E-ROBOT problem is to minimize $C_\varepsilon$ in (\ref{CepsLam}), that is
\begin{equation}
\inf_{\pi \in \Pi(\mu, \nu)}   C_\varepsilon(\mu, \nu, c_\lambda,  \pi) 
= \inf_{\pi \in \Pi(\mu, \nu)} \varepsilon H(\pi \| R_\varepsilon) -  \varepsilon \ln \beta, \label{EROBOT}
\end{equation}
which is a (static) Schr\"odinger bridge problem for $R_\varepsilon$, similarly to the problem stated in \eqref{SB}.  When $ \varepsilon > 0$, the E-ROBOT problem  is strongly convex, so that the optimal plan is unique. Then, the next propositions can be proved along the same lines as in Thm. 4.2 and Thm. 4.7 in \cite{N22}, to which we refer for the mathematical detail.

\begin{proposition}  \label{Prop1}
Let  \( c_\lambda \) be the ROBOT cost function and let $(X,\mu)$ and $(Y,\nu)$ 
be separable. Then:
\begin{itemize}
  \item[(i)]  there is a unique minimizer \( \pi_{\varepsilon}^* \in \Pi(\mu, \nu) \) for the E-ROBOT problem in \eqref{EROBOT};
    \item[(ii)]There exist measurable functions \( \varphi^* : X \to \mathbb{R} \), \( \psi^* : Y \to \mathbb{R} \)  (those are potentials which in fact depend on both $\varepsilon$ and $\lambda$) such that
  \[
  \frac{d\pi_{\varepsilon}^*}{d(\mu \otimes \nu)} = e^{\varphi^* + \psi^* - \varepsilon^{-1} c_\lambda} \quad \mu \otimes \nu\text{-a.s.}
  \]
  The potentials \( \varphi^*, \psi^* \) are unique up to an additive constant,  \( \varphi^* \in L^1(\mu) \), \( \psi^* \in 
L^1(\nu) \) and
 \begin{eqnarray}
  \varphi^{*}(x)  =  -\varepsilon \ln \int e^{ \psi^*(y) -  \varepsilon^{-1} c_\lambda(x,y)}   \nu(dy) \ \ \mu-\text{a.s.} \label{mySE1}\\
  \psi^{*}(y)       =  -\varepsilon \ln \int e^{ \varphi^*(x) -  \varepsilon^{-1} c_\lambda(x,y)}   \mu(dx) \ \ \nu-\text{a.s.} \label{mySE2} 
 \end{eqnarray}

   Conversely, if \( \bar \pi  \in \Pi(\mu, \nu) \) admits a density of the form
  \[
  \frac{d\bar \pi }{d(\mu \otimes \nu)} = e^{\varphi + \psi -  \varepsilon^{-1} c_\lambda } \quad \mu \otimes \nu\text{-a.s.}
  \]
  for measurable 
functions \( \varphi, \psi \) satisfying \eqref{mySE1} and \eqref{mySE2}, then \( \bar \pi = \pi_{\varepsilon}^* \).
\end{itemize}
\end{proposition}
%

Eq. \eqref{mySE1} and \eqref{mySE2} make explicit the link between the E-ROBOT and the Schr\"odringer equations, as in \eqref{SE1} and \eqref{SE2}. Moreover,
since $c_\lambda \in L^{1}(\mu \otimes \nu)$, we state 

\begin{proposition}[E-ROBOT Dual problem] \label{PropDual}
The  $C_\epsilon$  in \eqref{CepsLam} and the related E-ROBOT in \eqref{EROBOT} are such that:
\begin{equation}
\inf_{\pi \in \Pi(\mu, \nu)}   C_\varepsilon  = \sup_{\varphi \in L^1(\mu), \psi \in L^1(\nu)} \left\{ \int  \varphi \, d\mu + \int  \psi \, d\nu -  \varepsilon \int e^{\varphi + \psi - \varepsilon^{-1} c_\lambda} \, d(\mu \otimes \nu) +  \varepsilon \right\}, \label{Dualsup}
\end{equation}
and the supremum is attained by the E-ROBOT potentials \( \varphi^*, 
\psi^* \in L^1(\mu) \times L^1(\nu) \), with
\begin{equation}
\inf_{\pi \in \Pi(\mu, \nu)}  C_\varepsilon =  \varepsilon \left(\int \varphi^* \, d\mu + \int \psi^* \, d\nu\right). \label{Eq.dual}
\end{equation}
The maximizers are a.s. unique up to an additive constant.
\end{proposition}

In the next section, equipped with the results of Propositions \ref{Prop1} and \ref{PropDual}, we now study the main properties of the E-ROBOT problem and  of some of its related notions.

\section{Main results: key properties of E-ROBOT} \label{Sec:Main}

\subsection{Potentials: uniform boundedness and uniform convergence}

The solution to the SBP described in Eq. \eqref{SE1} and \eqref{SE2} illustrates the pivotal role played by Schr\"odinger potentials. Propositions \ref{Prop1} and  \ref{PropDual} further demonstrate that these potentials are fundamental to the solution of the E-ROBOT primal and dual problem. This leads to  natural questions regarding the limiting behavior of the potentials in \eqref{mySE1} and \eqref{mySE2},  as the regularization parameter approaches zero or as the sample size diverges. The next two propositions answer these questions. Specifically,  Proposition \ref{lemma2} shows that the potentials are Lipschitz continuous functions, uniformly bounded.

\begin{proposition} \label{lemma2}
Let \( X,Y \subset \mathbb{R}^d \) 
and \( c_\lambda : X \times Y \to \mathbb{R} \) be the continuous and bounded ROBOT cost function. Let \( \mu \in \mathcal{P}(X), \nu \in \mathcal{P}(Y) \) be probability measures, and let \( \varphi^*, \psi^* \) be the optimal dual potentials for the entropic optimal transport problem with cost \( c_\lambda \) and regularization parameter \( \varepsilon > 0 \). Then \( \varphi^* \) and \( \psi^* \): (i) are Lipschitz continuous functions on \( X \) and \( Y \), respectively; (ii) \( \varphi^* \in L^\infty(X) \) and \( \psi^* \in L^\infty(Y) \), namely they are uniformly bounded.
\end{proposition}

Besides being uniformly bounded, in the next proposition, we show also that the potentials converge uniformly as the sample size $n$ diverges.

\begin{proposition} \label{LemmaUCP}
Let $\mu_n, \nu_n$ be empirical measures based on $n$ i.i.d. samples from compactly supported probability measures $\mu, \nu$ on a compact sets $X,Y \subset \mathbb{R}^d$. Let $\phi^*_{n}, \psi^*_{n}$ be the Schrödinger potentials associated with the entropic optimal transport plan $\pi_n^* \in \Pi(\mu_n, \nu_n)$, and $\phi^*, \psi^*$ the potentials associated with the true plan $\pi^* \in \Pi(\mu, \nu)$, both for cost $c_\lambda$ and regularization parameter $\varepsilon > 0$. Let  $c_\lambda$ be the ROBOT cost function.
%
Then, as $n \to \infty$,
\[
\sup_{x \in X} |\phi^*_{n}(x) - \phi^*(x)| \to 0, \quad \sup_{y \in Y} |\psi^*_{n}(y) - \psi^*(y)| \to 0
\]
i.e., the potentials converge uniformly.
\end{proposition}

\subsection{Optimal regularized transport plan}

It is easy to conjecture that the structure of the E-ROBOT potentials and their limiting behavior have implications on the limiting behavior of the related transportation plan. In this section, we study that aspect.  More precisely, for fix $\lambda$, we discuss the limiting behavior as $\varepsilon \to 0$ of the E-ROBOT problem. The  optimal transport plan $\pi_0$ of the corresponding unregularized OT problem is defined via:
\begin{equation}
 C_0 = \inf_{\pi \in \Pi(\mu,\nu)} \int c_\lambda \, d\pi.
\label{EROBlim}
\end{equation}


Since the value function $\inf_{\pi \in \Pi(\mu, \nu)} C_\varepsilon(\pi)$ of E-ROBOT decreases monotonically as $\varepsilon \downarrow 0$, we have
$$
\lim_{\varepsilon \to 0} \inf_{\pi \in \Pi(\mu, \nu)} C_\varepsilon(\pi) \geq \inf_{\pi \in \Pi(\mu, \nu)} C_0(\pi).
$$
We now show this inequality is an equality, meaning the E-ROBOT value converges to the ROBOT value as regularization vanishes (Proposition \ref{conv1}(i)). Moreover, the corresponding optimizer converges weakly: the optimal plan \(\pi^*_\varepsilon\) for E-ROBOT converges to an optimal plan \(\pi^*_0\) for the original ROBOT problem with cost \(c_\lambda\) (Proposition \ref{conv1}(ii)).

\begin{proposition}  \label{conv1}

 Let \( c_\lambda \) be the cost associated to the ROBOT problem for two measures $\mu$ and $\nu$, and let $\pi_0$ be the optimal transport plan. If 
 \( H(\pi \| \mu \otimes \nu  ) < \infty \), then:
 \begin{itemize}
\item[(i)] we have that
\begin{equation}
\lim_{\varepsilon \to 0} \inf_{\pi \in \Pi(\mu, \nu)} C_\varepsilon =   C_0.
\label{ContC}
\end{equation}
\item[(ii)] 
if $\varepsilon_n \to 0$ and $\lim_{n \to \infty}  \pi^{*}_{\varepsilon_n} = \pi_0$ weakly, we have 
that $\pi_0  \in \Pi(\mu,\nu) $ is the unique unconstrained ROBOT plan and  it follows that  $\lim_{\varepsilon \to 0} \pi^{*}_{\varepsilon} = \pi_0  $ weakly.
\end{itemize}

\end{proposition}



Finally, we study the large sample behaviour ($n \to \infty$) of the regularized optimal transport plan. To this end, we state:

\begin{proposition}  \label{propD}
Let \( X,Y \subset \mathbb{R}^d \) be compact 
and \( \mu_n  \in \mathcal{P}(X), \nu_n \in  \mathcal{P}(Y)   \) be empirical measures based on i.i.d. samples from compactly supported probability measures \( \mu, \nu\), such that \( H(\pi \| \mu \otimes \nu  ) < \infty \). Let \( \pi_n^* \in \Pi(\mu_n, \nu_n) \) be the entropic optimal transport plan with cost \( c_\lambda \) and regularization parameter \( \varepsilon > 0 \), and let \( \pi^* \in \Pi(\mu, \nu) \) be the corresponding optimal plan for the true marginals. Then $ \pi_n^* \to \pi^*$ weakly as  $n \to \infty.$
\end{proposition}

\subsection{Robust Sinkhorn divergence and MMD} \label{DiffMeasures}

Given the cost function $c_\lambda(x, y)$ defined on $X \times Y$, with $Y = X$, and a regularization parameter $\varepsilon > 0$, the entropic regularized optimal transport cost between two probability measures $\mu$ and $\nu$ is denoted $W_{\varepsilon,\lambda}(\mu, \nu)$ and is defined as:
\[
W_{\varepsilon,\lambda}(\mu, \nu) = \inf_\pi  \int c_\lambda \, d\pi + \varepsilon H(\pi  \| \mu \otimes \nu).
\]
Due to the regularization, $W_{\varepsilon,\lambda}(\mu, \mu) $ is not granted to be zero and this entails the entropic bias. To correct for this bias and ensure that the loss vanishes when $\mu = \nu$, we propose the following modified unbiased version:
\begin{equation}
\overline{W}_{\varepsilon,\lambda}(\mu, \nu) := W_{\varepsilon,\lambda}(\mu, \nu) - \frac{1}{2} \left( W_{\varepsilon,\lambda}(\mu, \mu) + W_{\varepsilon,\lambda}(\nu, \nu) \right), \label{Wel}
\end{equation}
and we call it {the robust Sinkhorn loss}.  When $\lambda \to \infty$, $c_\lambda \to c$ so we obtain the usual Sinkhorn loss, which is commonly referred to as the {Sinkhorn divergence} in the OT literature that has been studied in \cite{genevay2018learning, genevay2019sample}.   %

The following proposition states that  the  robust Sinkhorn divergence $\overline{W}_{\varepsilon,\lambda}$ defines a symmetric and positive definite loss function that is convex in each of its input variables. Moreover, it   metrizes the convergence in law.

\begin{proposition} \label{propopMetrics}
Let $X \subset \mathbb{R}^d$ 
and consider the cost function $c_\lambda(x,y)$. Then, 
  for all probability measures $\mu$ and $\nu$ on $X$, $\overline{W}_{\varepsilon,\lambda}$ is such that:
\begin{itemize}  
\item[(i)]  $0= \overline{W}_{\varepsilon,\lambda}
(\nu, \nu) \leq \overline{W}_{\varepsilon,\lambda}
(\mu, \nu), $
\item[(ii)]  $\mu = \nu \iff \overline{W}_{\varepsilon,\lambda} (
 \mu, \nu) = 0, $
 \item[(iii)]  $\lim_{n \to \infty } {\mu_n}= \mu \ \text{weakly} \ \iff \overline{W}_{\varepsilon,\lambda} (\mu_n, \mu) \to 0 $.  
\item[(iv)] (Limiting behavior for $\varepsilon \to 0$) $\overline{W}_{\varepsilon,\lambda}(\mu, \nu) \to  W_\lambda (\mu, \nu),$ as $ \varepsilon \to 0.$ 

\end{itemize}  
\end{proposition}
In addition to the limiting behavior for $\varepsilon \to 0$ discussed in Proposition \ref{propopMetrics},  $\overline{W}_{\varepsilon,\lambda}$ has connections to  MMD (see e.g. \cite{gretton2006kernel}) and Bregman-type divergence (see e.g. \cite{pardo2018statistical} for book-length introduction). \cite{gretton2006kernel} introduced MMD  in machine learning and since then they have been applied for different tasks, like e.g. comparing distributions via distribution-free tests \cite{gretton2012kernel}, generative models \cite{li2015generative}, gradient flow and neural network optimization \cite{arbel2019maximum}.  To illustrate the connection, let us consider 
\begin{equation}
\|\mu - \nu\|^2_{- c_\lambda}
:= \iint -c_{\lambda}(x,y) \, d(\mu-\nu)(x) \, d(\mu-\nu)(y), 
\end{equation}
which is the MMD with kernel $- c_\lambda$. We refer to Lemma 12 
in Appendix A 
for the properties of this kernel, here we remark that expanding the last expression, we obtain
\begin{equation}
\|\mu - \nu\|^2_{- c_\lambda}
=  \iint - c_\lambda\, d\mu(x)\, d\mu(y)
+ \iint - c_\lambda\, d\nu(x)\, d\nu(y)
- 2 \iint - c_\lambda\, d\mu (x)\, d\nu(y). \label{MMD}
\end{equation}
From the definition of $W_{\lambda,\varepsilon}(\mu, \nu)$,
when $\varepsilon \to +\infty$, the entropic term $\varepsilon \, H(\pi \,\|\, \mu \otimes \nu)$ dominates.
The solution to the E-ROBOT problem is then close to the independent coupling and the transport cost term becomes $
\iint c_\lambda(x,y) \, d\mu(x) \, d\nu(y).$
So, as $\varepsilon \to \infty$, for
each $W_{\lambda,\varepsilon}$  in \eqref{Wel} we have 
\[
\begin{aligned}
W_{\lambda,\varepsilon}(\mu, \nu) &\to \iint c_\lambda(x,y)\, d\mu(x) d\nu(y), \\
W_{\lambda,\varepsilon}(\mu, \mu) &\to \iint  c_\lambda(x,y)\, d\mu(x) d\mu(y), \\
W_{\lambda,\varepsilon}(\nu, \nu) &\to \iint  c_\lambda(x,y)\, d\nu(x) d\nu(y).
\end{aligned}
\]
Thus, as  $\varepsilon \to \infty$ we have 
\begin{equation}
 \overline{W}_{\varepsilon,\lambda} (\mu,\nu) \to \iint  c_\lambda(x,y)\, d\mu(x) d\nu(y) - \frac12 \left(\iint  c_\lambda(x,y)\, d\mu(x) d\mu(y) +  \iint  c_\lambda(x,y)\, d\nu(x) d\nu(y) \right), \label{W2}
 \end{equation}
and, comparing \eqref{W2} to \eqref{MMD}, we conclude that 
\begin{equation}
\lim_{\varepsilon \to \infty} \overline{W}_{\varepsilon,\lambda} (\mu,\nu) =  \frac12 \|\mu - \nu\|^2_{- c_\lambda} \label{limMMD}
\end{equation}
namely, in the large-$\varepsilon$ limit, $\overline{W}_{\varepsilon,\lambda} (\mu,\nu) $ becomes half the squared MMD norm
with kernel $- c_\lambda$. 
This results provide an interesting interpretation for the E-ROBOT: by changing $\varepsilon$, the E-ROBOT interpolates between the ROBOT and the MMD norm as obtained using the truncated cost as kernel. 

In addition to this property, making use of  the  positive and $c$-universal kernel $k_{\varepsilon,\lambda}$ (see Lemma 12 in Appendix A), 
we define the MMD:
\begin{equation}
\Vert \mu \Vert^2_{k_{\varepsilon,\lambda}} = \iint k_{\varepsilon,\lambda}  d\mu(x) d\mu(y) = \iint e^{- \frac{1}{\varepsilon} c_\lambda(x,y)} \, d\mu(x) d\mu(y), \label{MMDkel}
\end{equation}
and we define also the E-ROBOT negentropy $F_{\varepsilon, \lambda}(\mu)$ as a mapping from $\mathcal{P}(X)$ to $\mathbb{R}$:
\begin{equation}\label{EqNegEntr}
F_{\varepsilon, \lambda}: \mu \mapsto - \frac{1}{2} W_{\varepsilon,\lambda}(\mu, \mu) = - \frac{1}{2} \inf_{\pi \in \Pi(\mu, \mu)}  C_\varepsilon(\mu, \mu, c_\lambda, \pi)
\end{equation}
Working along the same lines as Proposition 4 in \cite{feydy2019} (see Appendix A, Lemma~2), we can prove that
\begin{equation}
	\frac{1}{\varepsilon} F_{\varepsilon, \lambda}(\mu) + \frac{1}{2} = \inf_{\xi \in \mathcal{P}(X)} \left\{ \int \ln \left( \frac{d\mu}{d\xi} \right) d\mu + \frac{1}{2} 
	\Vert \mu \Vert^2_{k_{\varepsilon,\lambda}} \right\},
	\label{my18}
\end{equation}
where $F_{\varepsilon, \lambda}$ is a strictly convex functional on $\mathcal{P} (X)$.  Moreover, $F_{\varepsilon, \lambda}$ is differentiable in the following sense.

Let $\mathcal{C}(X)$ denote the set of continous function on $X$, and define the operator $\langle \cdot  ,  \cdot \rangle:  \mathcal{P}(X) \times \mathcal{C}(X) \rightarrow \mathbb{R}$ as the mapping  $(\mu, f) \mapsto \int f(x) d\mu(x)$. 
Recall that a function $F: \mathcal{P}(X) \rightarrow \mathbb{R}$ is {differentiable} at $\mu \in \mathcal{P}(X)$ if there exists a continuous function (called {gradient}) $\nabla F(\mu) \in \mathcal{C}(X)$ such that for any $\xi = \nu_1 - \nu_2$ with $\nu_1, \nu_2 \in  \mathcal{P}(X)$, we have
$$F(\mu + t\xi) = F(\mu) + t \langle \xi, \nabla F(\mu) \rangle + o(t).$$ Then, moving along the same lines as the proof in Appendix B2 of \cite{feydy2019}, we have the following proposition, which implies that  the E-ROBOT negentropy $F_{\varepsilon, \lambda}$ is differentiable everywhere in $\mathcal{P} (X)$ with gradient $\nabla F_{\varepsilon, \lambda}(\mu)(x) = -\varphi^*(x)/2$ for $x \in X$.


\begin{proposition} \label{GradEROBOT}
Let \( \mu, \nu \in \mathcal{P}(X) \) be probability measures on a compact set \( X \subset \mathbb{R}^d \), and let \( c_\lambda : X \times X \to \mathbb{R} \) be the ROBOT cost function. Then the entropic cost
$W_{\varepsilon,\lambda}(\mu, \nu)$
is weak-* continuous and differentiable over $\mathcal{P}(X) \times \mathcal{P}(X)$,  with the gradient given by the pair of Schrödinger potentials:  
\[
\nabla W_{\varepsilon,\lambda}(\mu, \nu) = (\varphi^*, \psi^*).
\]
\end{proposition}


In Section~\ref{sec.GradFlow}, we show that the formula of the gradient $\nabla F_{\varepsilon, \lambda}(\mu)$ can be applied to model the evolution of a distribution along the E-ROBOT gradient flow. Also, it allows us to define the E-ROBOT Hausdorff divergence 
\begin{equation}
H_{\varepsilon, \lambda}(\mu, \nu) := \frac{1}{2} \left\langle \mu - \nu, \nabla F_{\varepsilon, \lambda}(\mu) - \nabla F_{\varepsilon, \lambda}(\nu) \right\rangle, \label{Hd}
\end{equation}
which is a  Bregman-type divergence induced by the strictly convex functional \( F_{\varepsilon, \lambda} \) and is therefore a positive definite quantity.

\subsection{Sample complexity}

The results of \S \ref{DiffMeasures} illustrate the properties of $\overline{W}_{\varepsilon,\lambda}$ as a tool to measure the proximity between two (probability) measures. This is similar to $W_p$, $W^{\lambda}$, and $W_{\varepsilon}$. However, the results of \S \ref{DiffMeasures} are stated at the population level. In applications, $W_p$ and $W^{\lambda}$ are estimated from samples. A well-known issue is that the error of these empirical estimates suffers from a serious dependence on dimension: the rate at which $W_p(\hat{\mu}_n, \mu)$ and $W^{(\lambda)}(\hat{\mu}_n, \mu)$ converge to 0 scales as $n^{-1/d}$ under mild moment conditions for $d \geq 3$. Thus, this rate (also called sample complexity) deteriorates poorly with dimension. As shown in \cite{genevay2019sample}, $W_{\varepsilon}$ does not suffer from the same problem: its sample complexity scales with $n^{-1/2}$ in any dimension $d$. In Theorem \ref{Thm1} and Corollary \ref{corollary}, we show that $\overline{W}_{\varepsilon,\lambda}$ enjoys the same desirable property.

Before presenting the statements, we clarify an important theoretical point. Although our results on the sample complexity of $\overline{W}_{\varepsilon,\lambda}$ are consistent with those obtained for the Sinkhorn divergence in E-OT by \cite{genevay2019sample} and for Gaussian-smoothed OT in \cite{nietert2021smooth}, our proof requires a fundamentally different derivation and cannot directly build upon existing strategies. For instance, the proofs in \cite{genevay2019sample} require that the cost function $c$ is smooth, a condition not satisfied by our truncated cost $c_\lambda$. The proofs in \cite{nietert2021smooth} hinge on the specific idea of convolving measures with an isotropic Gaussian density, which is not a feature of our $\overline{W}_{\varepsilon,\lambda}$ framework. Therefore, to establish the sample complexity of $\overline{W}_{\varepsilon,\lambda}$, we must resort to different mathematical tools (empirical process theory), leveraging the special structure of the E-ROBOT potentials.

The novelty of our approach lies in this application of empirical process theory to the E-ROBOT potentials, whose regularity properties are guaranteed by the entropic regularization. To provide intuition for this strategy, recall that $\overline{W}_{\varepsilon,\lambda}(\mu,\nu)$ depends on ${W}_{\varepsilon,\lambda}(\mu,\nu)$, which in turn depends on the entropic regularization term $\varepsilon H(\pi \| \mu \otimes \nu)$. Now, recall that
\[
H(\pi \| \mu \otimes \nu) = \int \ln\left( \frac{d\pi}{d(\mu \otimes \nu)} \right)  d\pi,
\]
where the optimal plan $\pi^*$ is:
\[
\frac{d\pi^*}{d(\mu \otimes \nu)}(x,y) = \exp\left( \varphi^*(x) + \psi^*(y) - \frac{1}{\varepsilon} c_\lambda(x,y) \right),
\]
and similarly for $\pi_n^*$ with potentials $\varphi_n^*, \psi_n^*$ and empirical marginals $\mu_n, \nu_n$. Substituting into the entropy expression, we obtain:
\[
H(\pi^* \| \mu \otimes \nu) = \int \left( \varphi^*(x) + \psi^*(y) - \frac{1}{\varepsilon} c_\lambda(x,y) \right)  d\pi^*(x,y).
\]
Now, since $\varphi_n^*, \psi_n^* \to \varphi^*, \psi^*$ uniformly (see Lemma~\ref{LemmaUCP}), and $c_\lambda$ is bounded and Lipschitz, the integrand
$
(x,y) \mapsto \varphi_n^*(x) + \psi_n^*(y) -  c_\lambda(x,y)/{\varepsilon}$
is itself uniformly bounded and Lipschitz. Moreover, from Proposition~\ref{conv1}, the plans $\pi_n^* \to \pi^*$ weakly, and the domain is compact. By standard empirical process theory (uniform convergence for Lipschitz function classes, see Sections 2.2 and 2.5.1 in \cite{vandervaart1996weak}), we obtain $\mathbb{E} \left[ \left| H(\pi_n^* \| \mu_n \otimes \nu_n) - H(\pi^* \| \mu \otimes \nu) \right| \right] = \mathcal{O}(n^{-1/2})$.
This follows because the Schrödinger potentials $\varphi_n^*, \psi_n^*$ are uniformly bounded and Lipschitz (see Proposition~\ref{lemma2}), and their uniform convergence (see Proposition~\ref{LemmaUCP}) ensures that the integrand in the entropy expression is itself Lipschitz and bounded. Therefore, $H(\pi_n^* \| \mu_n \otimes \nu_n) - H(\pi^* \| \mu \otimes \nu)$ behaves like an empirical process indexed by a class of functions that is both Glivenko–Cantelli and Donsker, and its expectation converges at the rate $n^{-1/2}$.

The above derivation provides the basic intuition for how the SBP, the ROBOT, and the empirical process theory can be combined nicely to establish the sample complexity of $\overline{W}_{\varepsilon,\lambda}(\mu,\nu)$. With this regard, we state:

\begin{theorem} \label{Thm1}
Let $\mu_n, \nu_n$ be empirical measures based on $n$ i.i.d. samples from compactly supported probability measures $\mu, \nu$ on  compact set $X \subset \mathbb{R}^d$. Let $c_\lambda$ be a bounded Lipschitz cost function and $\varepsilon > 0$ a fixed regularization parameter. Then the expected deviation of the robust Sinkhorn loss satisfies:
\[
\mathbb{E}\left[ \left| \overline{W}_{\varepsilon,\lambda}(\mu_n, \nu_n) - \overline{W}_{\varepsilon,\lambda}(\mu, \nu) \right| \right] =  \mathcal{O}(n^{-1/2}).
\]
\end{theorem}

\begin{corollary}  \label{corollary}
Under the same assumptions as Theorem \ref{Thm1}, we have 
$
\mathbb{E}\left[ \overline{W}_{\varepsilon,\lambda}(\mu_n, \mu) \right] = \mathcal{O}(n^{-1/2}).
$
\end{corollary}

\begin{remark}  \label{Remark}
The proofs of these results reveal an important aspect. A key distinction between E-ROBOT and ROBOT lies in the structure of their respective dual potentials. In E-ROBOT, the Schrödinger potentials $\phi^*, \psi^*$ satisfy fixed-point equations in \eqref{SE1} and \eqref{SE2}, which correspond to a {SoftMax} operation; see \cite{CP19}, Ch. 4. This structure induces smoothness and regularity, ensuring that the potentials are uniformly bounded and Lipschitz continuous. As a result, the function class indexed by these potentials has finite entropy and is P-Donsker under compact support.  In contrast, the ROBOT framework lacks entropic regularization, and its Kantorovich potentials arise from a linear program with Lipschitz and range constraints. While these potentials are bounded and Lipschitz, they do not enjoy the smoothing effects of the SoftMax structure. Consequently, uniform convergence and Donsker-type properties are not guaranteed in ROBOT  when $d \geq2$.
\end{remark}

%


\subsection{Truncated Laplace deconvolution}

\newcommand{\PiSet}{\Pi}

\cite{RigolletWeeds2018} give a statistical interpretation of E-OT for $W_2$ by showing that
performing maximum-likelihood estimation for Gaussian deconvolution corresponds to
calculating a projection with respect to the entropic optimal transport distance. The projection estimator as obtained using $W_2$ has been employed in the machine learning community 
as a smoothed version of a minimum Kantorovich distance estimator (MKE, \cite{bassetti2006minimum,bassetti2006asymptotic}) more suitable for automatic differentiation in generative models; see \cite{MMC16},\cite{genevay2018learning, genevay2019sample}.  We now prove that this connection between EOT and maximum-likelihood deconvolution can be  derived also for the E-ROBOT, considering the deconvolution with a truncated Laplace distribution. 

To elaborate, we start by recalling that a class $\mathcal{P}$ of probability measures is said to be closed under domination, if $\mu_1 \ll \mu_2$ for some $\mu_2 \in  \mathcal{P}$ implies that $\mu_1 \in \mathcal{P}$. Moreover,
let $\mathcal{\mu}$ contain probability distributions over $\mathbb{R}^d$  and let $\mu^*$ be an unknown distribution of an i.i.d. sample $X_1, \ldots, X_n$.  The deconvolution problems consists in estimating $\mu^*$ using the corrupted random observations $(Q_1, \ldots, Q_n)$, where
\begin{equation}
Q_i=X_i+Z_i, \quad i=1, \ldots, n \label{LaplMech}
\end{equation}
and the errors $Z_1, \ldots, Z_n$ are independent of $X_1, \ldots, X_n$. In what follows, the random variables $\left\{Z_i\right\}$ are assumed to be independent copies of a random variable $Z$ with known truncated Laplace distribution: $Z \sim \mathcal{L}\left(0, \varepsilon, \lambda\right)$, where the location is zero, the scale is $\varepsilon$, and the truncation parameter is $\lambda$.

In this context, the distribution of $Q_i$ has density $f_{\left(0, \varepsilon, \lambda\right)} \star \mathrm{d} \mu^*$, where, for any $\mu \in \mathcal{P}$, we define
$$
f_{\left(0, \varepsilon, \lambda\right)} \star \mathrm{d} \mu(y)=\int f_{\left(0, \varepsilon, \lambda\right)} (y-x) \mathrm{d} \mu(x)
$$
and $f_{\left(0, \varepsilon, \lambda\right)}$ denotes the density of $Z \sim \mathcal{L}\left(0, \varepsilon, \lambda\right)$. Under these assumptions, we call (\ref{LaplMech}) the truncated Laplacian deconvolution model. The maximum-likelihood estimator (MLE) $\hat{\mu}$ defined by
$$
\hat{\mu}=\underset{\mu \in \mathcal{P}}{\operatorname{argmax}} \sum_{i=1}^n \ln f_{\left(0, \varepsilon, \lambda\right)} \star \mathrm{d} \mu\left(Q_i\right)
$$
is a natural candidate to estimate $\mu^*$. 

Equipped with these definitions, we state a proposition that makes the link between the E-ROBOT  and the truncated Laplacian deconvolution model, showing that E-ROBOT is in fact implementing $\hat{\mu}$.

\begin{proposition} \label{PropDec}
Let $\nu_n = {\sum_{i=1}^n \delta_{q_i}}/{n}$ be an empirical measure of the observations $(q_1,q_2,...,q_n)$. Let $\mathcal{P}$ be a convex class of probability measures that is closed under domination.
The maximum-likelihood estimator for the Laplace convolution model $Q = X + Z$, where the noise $Z$ has truncated Laplace density $f_{\left(0, \varepsilon, \lambda\right)} (z) \propto \exp(-\varepsilon^{-1} c_\lambda(0, z))$, is given by 

$$
\hat{\mu} = \arg \min_{\mu \in \mathcal{P}} W_{\lambda,\varepsilon}(\mu, \nu_n).
$$
\end{proposition}

Proposition \ref{PropDec} implies that  the maximum-likelihood estimator $\hat{\mu}$ is the projection of the empirical measure $\nu_n$ onto $\mathcal{P}$ with respect to  $W_{\lambda,\varepsilon}$. Differently from the $W_2$ case of \cite{RigolletWeeds2018}, the E-ROBOT framework corresponds to a deconvolution problem with a specific, robust noise model: the noise variable $Z$ is assumed to follow a {truncated Laplacian distribution}. This distribution has the following properties:  (i) for $\|z\| \leq 2\lambda$, $f_{0,\varepsilon, \lambda}(z) \propto \exp(-\|z\|/\varepsilon)$, identical to a standard Laplace distribution; (ii) for $\|z\| > 2\lambda$, $f_{0,\varepsilon, \lambda}(z) \propto \exp(-2\lambda/\varepsilon)$, a constant value. This implies a uniform distribution on the tails beyond the radius $2\lambda$; (iii) the log-probability of any large value is bounded below by $-{2\lambda}/{\varepsilon}$, making the model robust to outlying observations. The parameter $\lambda$ controls the robustness (truncation point), while $\varepsilon$ controls the scale (dispersion) of the core Laplace distribution component. 
Proposition \ref{PropDec} justifies the use of E-ROBOT for inference and prediction tasks, like those described in \cite{Maetal25}---e.g. minimum Kantorovich distance estimation, generative models, domain adaptation, and outliers detection.

\section{Numerical illustrations} \label{SecNum}

We illustrate the benefits and ease-of-use of E-ROBOT with four examples, covering key statistical inference issues and typical machine learning problems. All calculations were performed on a standard laptop with a 2.4 GHz 8-Core Intel Core i9 processor, with each exercise requiring only a few minutes. Code to replicate our results is available on GitHub at \url{https://github.com/dvdlvc/E-ROBOT} and can be combined with the ROBOT code available at \url{https://github.com/dvdlvc/Robust-optimal-transportation}.

A methodological note on implementation. The application of E-ROBOT requires selecting the hyper-parameters $\lambda$ (from ROBOT) and $\varepsilon$ (from E-OT). A theoretically grounded, general-purpose procedure for this joint selection remains a fundamental open challenge---rather than a limitation specific to this work. The literature, in fact, offers scant guidance even on selecting these parameters in isolation, with some exceptions like the ROBOT  of  \cite{Maetal25}. We regard the derivation of such a joint selection criterion as an \textit{essential but separate} line of theoretical research---see \S \ref{conclusion}. 
In practice,  we emphasize that E-ROBOT is highly operable. We demonstrate that small values of $\varepsilon$ (e.g., on the order of $1E^{-2}$), consistent with standard E-OT procedures (see e.g. \cite{CP19}), yield excellent  results across all our experiments (e.g., the power curves in \S \ref{GoF} and barycenter calculations in \S \ref{Bary}) even for large dimensions. Furthermore, inheriting the stability of its ROBOT predecessor, the method performs effectively across a wide range of $\lambda$ values. For implementation, we advise analysts to inspect the truncated cost matrix: the distribution of its entries provides immediate, empirical insight into the scale of outlier-induced costs and a suitable range for $\lambda$.

\subsection{Nonparametric tests for equality of multivariate distributions} \label{GoF}
%

The $\overline{W}_{\lambda,\epsilon}$ and/or the related MMD or divergences described in \S \ref{DiffMeasures} can be applied to devise testing procedures aimed to determine whether given two distribution are the same. More generally, they can be used as a loss functions in various statistics machine learning tasks such as density estimation, domain adaptation, and generative models.  Among these different uses, we consider the Goodness-of-Fit (GoF) test problem discussed in \S 2 of \cite{hallin2021multivariate} uses the empirical Wasserstein distance between the empirical distribution $\hat{\mu}_n$ and a fully specified null distribution $\mu_0$. Their test statistic is based on $W_p^p(\hat{\mu}_n, \mu_0)$, with critical values determined via Monte Carlo simulation under the null. While this approach is fully nonparametric, it is sensitive to outliers and suffers from the curse of dimensionality. To address these limitations, we propose replacing the classical Wasserstein distance with the robust Sinkhorn distance $\overline{W}_{\varepsilon,\lambda}$ from the E-ROBOT framework. Specifically, we define the test statistic as:
\begin{equation}
\overline{T}_n := \overline{W}_{\varepsilon,\lambda}(\hat{\mu}_n, \mu_0). \label{GoFtest}
\end{equation}
This modification inherits the robustness properties of ROBOT and the statistical regularity of entropic optimal transport, including dimension-independent convergence rates and uniform convergence of the associated Schr\"{o}dinger potentials.

Under the conditions of Proposition \ref{propopMetrics}, one can prove a consistency result analogous to Proposition 1 in \cite{hallin2021multivariate}: for any fixed alternative $\mu \neq \mu_0$, the test based on $\overline{T}_n$ rejects the null hypothesis with probability tending to one as $n \to \infty$:
$\lim_{n \to \infty} \mathbb{P}(\overline{T}_n > c_n(\alpha)) = 1,$
where $c_n(\alpha)$ is the Monte Carlo critical value at level $\alpha$.   This follows from the convergence of $\overline{W}_{\varepsilon,\lambda}(\hat{\mu}_n, \mu)$ to $\overline{W}_{\varepsilon,\lambda}(\mu, \mu_0)$ and the fact that $\overline{W}_{\varepsilon,\lambda}(\mu, \mu_0) > 0$ under the alternative.

To implement the test, we compute $\overline{T}_n$ as in \eqref{GoFtest} using an adaptation of the Sinkhorn algorithm  (see e.g. \cite{CP19}) for the ROBOT setting, as in the pseudocode available in Algorithm \ref{MySink}. Then, we apply it to the mentioned GoF problem. Specifically,
%
we test two null hypotheses: first, $\mathcal{H}_0: \mu_0 = \mathcal{N}(0_d, I_d)$ (a $d$-variate standard normal); second, $\mathcal{H}_0: \mu_0 = {t}(0_d,  I_d, 1)$, namely a $d$-variate $t$-distribution with $df=1$, location $0_d$ (the $d$-dimensional vector of zeros), and scale $d \times d$-matrix $ I_d$. To examine the role of dimensionality $d$, we study power curves for $d \in \{2,10,15\}$, sample size $n=50$, significance level $5\%$, and a sequence of local alternatives obtained by shifting the location parameter equally in each dimension. For comparison, we also include the test statistic based on the $W_1$ distance from \cite{hallin2021multivariate}. Note that in this setting, $\overline{W}_{\lambda,\varepsilon}$ is well-defined for all considered distributions, whereas $W_1$ is well-defined for the Gaussian cases but not for the $t$-distributions (where the first moment does not exist).

\renewcommand{\algorithmiccomment}[1]{\hfill$\triangleright$ #1} 

\begin{algorithm}
\caption{E-ROBOT Sinkhorn Algorithm}\label{MySink}
\begin{algorithmic}[1]
\State \textbf{Input:}
\State \quad $1_n = (1, \ldots, 1)^\top$
\State \quad Source marginal: $\mu \in \mathbb{R}_{+}^{n}$, where $\mu^\top 1_n = 1$
\State \quad Target marginal: $\nu \in \mathbb{R}_{+}^{n}$, where $\nu^\top 1_n = 1$
\State \quad Regularization parameter: $\varepsilon > 0$
\State \quad Robustness parameter: $\lambda > 0$
\State \textbf{Output:}
\State \quad Approximate optimal transport plan: $\pi^{(t)}$
\State \quad Final scaling vectors: $u^{(t)}, v^{(t)}$

\Procedure{EROBOT\_Sinkhorn}{$\mu, \nu, \varepsilon, \lambda$}
    \State // Precompute the Gibbs kernel matrix $K$
    \For{$i = 1$ \textbf{to} $n$}
        \For{$j = 1$ \textbf{to} $n$}
            \State $c_\lambda(i,j) \gets \min(\|x_i - y_j\|, 2\lambda)$
            \Comment{Eq. (\ref{Eq.newcost}): ROBOT cost}
            \State $K(i,j) \gets \exp(-c_\lambda(i,j) / \varepsilon)$
            \Comment{Eq. (\ref{EqGibbs}): E-ROBOT kernel}
        \EndFor
    \EndFor

    \State // Initialize scaling vectors
    \State $u^{(0)} \gets 1_n$

    \State // Main Sinkhorn iteration loop
    \For{$t = 0, 1, 2, \ldots$} \Comment{Iterate until convergence}
        \State $v^{(t)} \gets \nu / (K^\top u^{(t)})$
        \Comment{Element-wise division}
        \State $u^{(t+1)} \gets \mu / (K v^{(t)})$
        \Comment{Element-wise division}
    \EndFor

    \State // Form the approximate optimal transport plan
    \State $\pi^{(t)} \gets \operatorname{diag}(u^{(t)}) \, K \, \operatorname{diag}(v^{(t)})$

    \State \textbf{return} $\pi^{(t)}, u^{(t)}, v^{(t)}$
\EndProcedure
\end{algorithmic}
\end{algorithm}

\begin{figure}[!h]
    \centering
    \begin{tabular}{ccc}  
        \includegraphics[width=0.31\textwidth, height=0.375\textwidth]{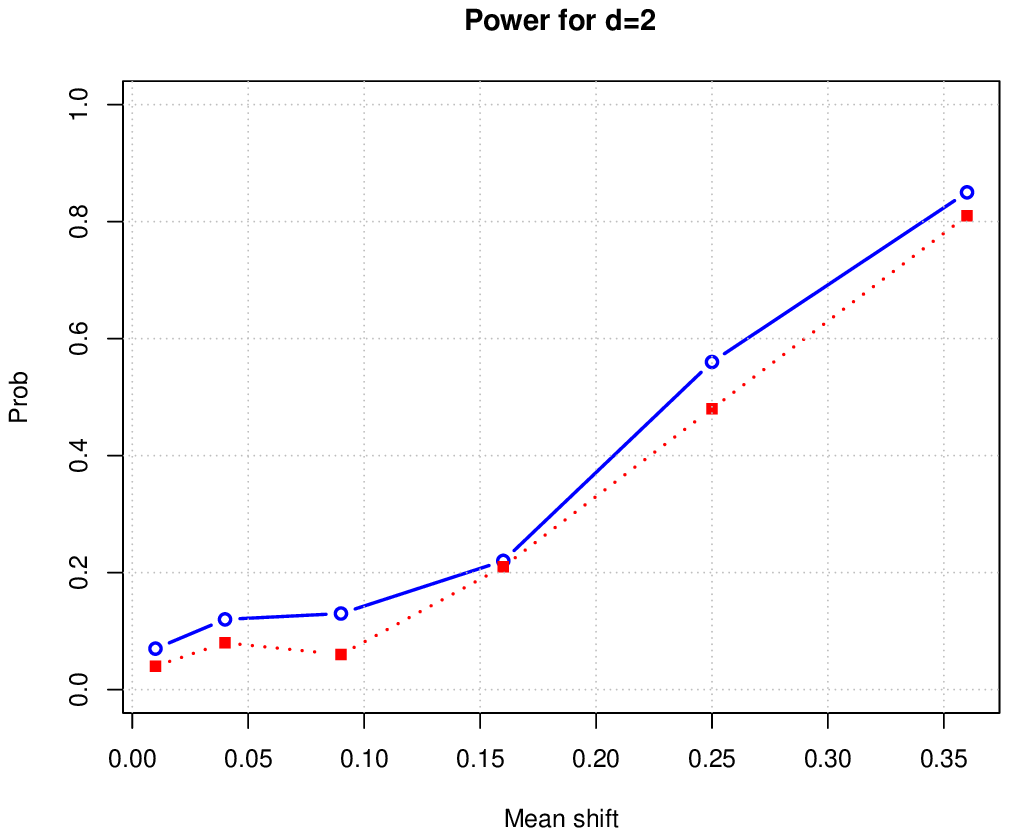} &
        \includegraphics[width=0.31\textwidth, height=0.375\textwidth]{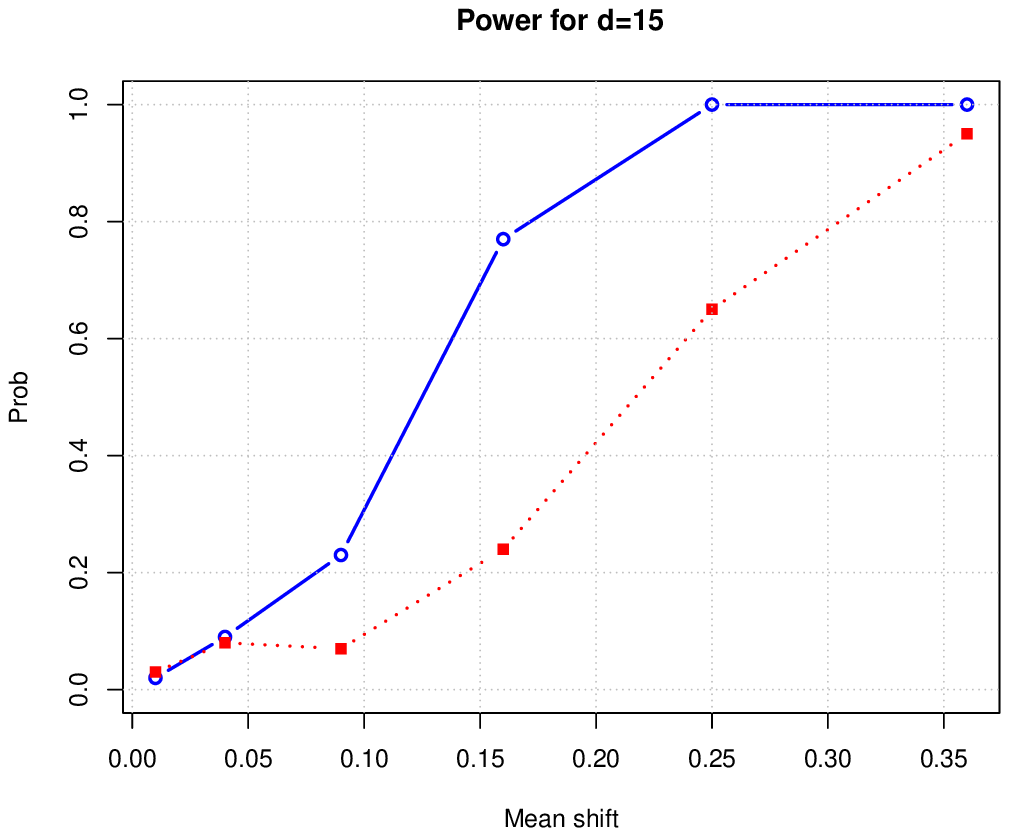} &
        \includegraphics[width=0.31\textwidth, height=0.375\textwidth]{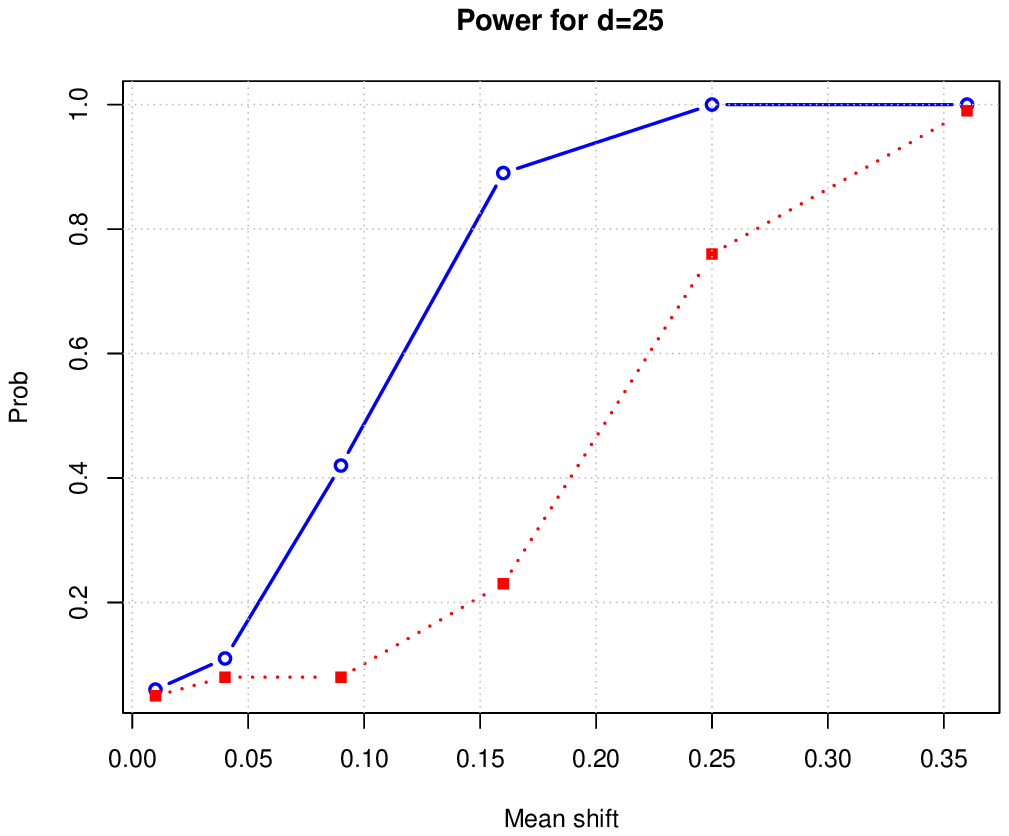}
    \end{tabular} 
    \caption{GoF power curves for testing the null of a $d$-variate standard normal. The continuous (blue) curve with circles corresponds to the $\overline{T}_n$ statistic using the entropic regularized Sinkhorn distance with $\varepsilon=5$ and $\lambda=10$; the dotted (red) curve with squares corresponds to the test based on $W_1$.}
    \label{FigNorm}
\end{figure}

Figure \ref{FigNorm} displays the results for the multivariate normal case. The left plot shows that our test and the $W_1$-based test have similar power for $d=2$. However, as $d$ increases, the power of $\overline{T}_n$ exceeds that of the $W_1$-based test. For the $t$-distribution (Figure \ref{Figtdist}), the $W_1$-based test performs near its level, while $\overline{T}_n$ maintains good power across all dimensions. These results complement the preview of our findings that we provided in \S \ref{Sec:Preview}.

\begin{figure}[!h]
    \centering
    \begin{tabular}{ccc}  
        \includegraphics[width=0.31\textwidth, height=0.375\textwidth]{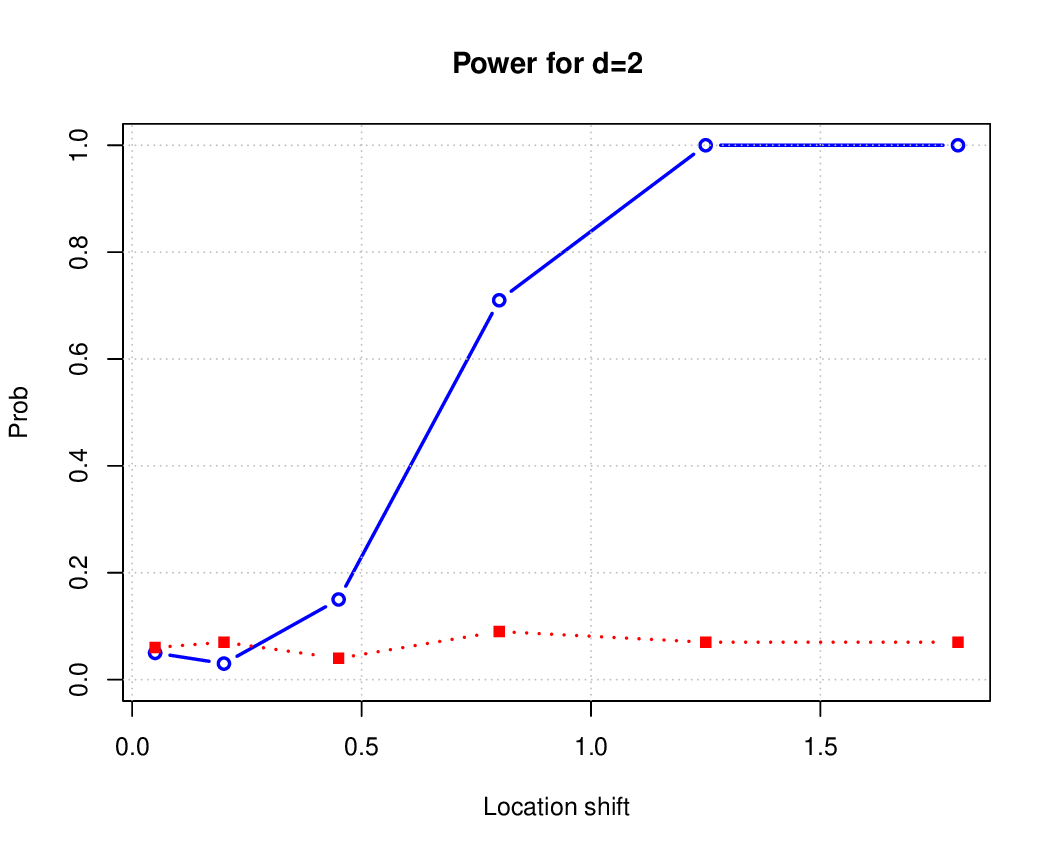} &
        \includegraphics[width=0.31\textwidth, height=0.375\textwidth]{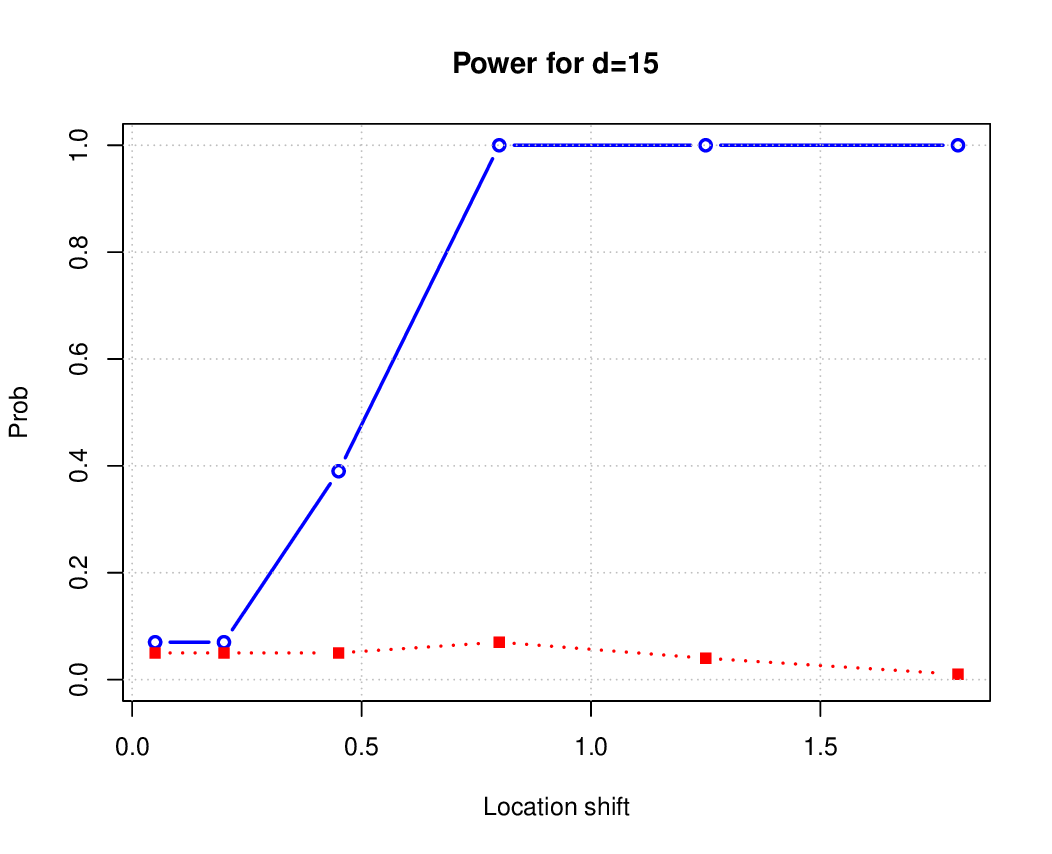} &
        \includegraphics[width=0.31\textwidth, height=0.375\textwidth]{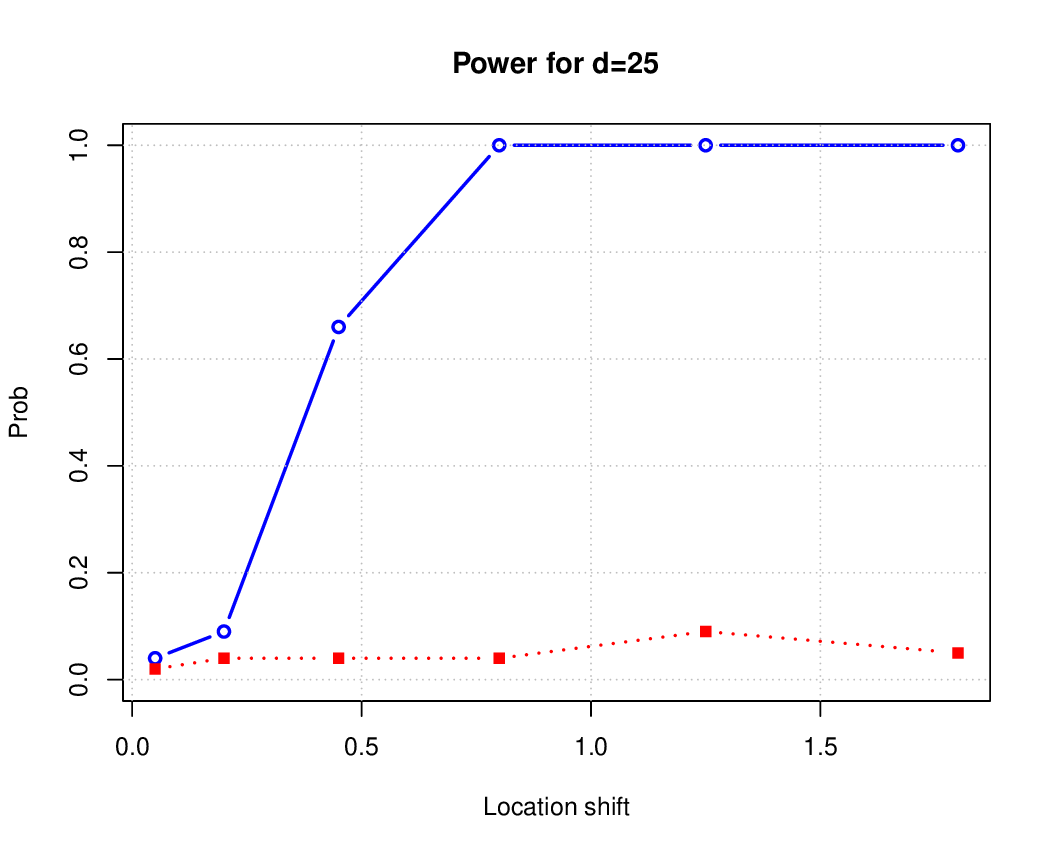}
    \end{tabular} 
    \caption{GoF power curves for testing the null of a $d$-variate $t$-distribution with one degree of freedom, location zero, and identity scale matrix. The continuous (blue) curve with circles corresponds to the $\overline{T}_n$ statistic using the entropic regularized Sinkhorn distance with $\varepsilon=0.05$ and $\lambda=3$; the dotted (red) curve with squares corresponds to the test based on $W_1$.}
    \label{Figtdist}
\end{figure}

\subsection{E-ROBOT barycenters}

\subsubsection{IBP for E-ROBOT} \label{IBP-method}

 Computing Wasserstein barycenters is a fundamental task in OT, with machine learning applications ranging from shape analysis and image synthesis to generative  modeling. It provides a principled way to define a central representative of a collection of probability measures, generalizing the notion of the Fréchet mean to the Wasserstein space.     
 
 Let $\Sigma_n := \left\{ \mu \in \mathbb{R}_+^n \;:\; \mu^\top 1_n  = 1 \right\}$ denote the simplex in $\mathbb{R}^n$.  In the classical entropic barycenter formulation, the barycenter $\mu \in \Sigma_n$ of input measures $\mu_m \in \Sigma_n, m = 1, \ldots, M$, is defined as the solution to
$\min_{\mu \in \Sigma_n} \sum_{m=1}^M \alpha_m W_\varepsilon(\mu_m, \mu)$. 
Given $(\mu, \nu) \in \Sigma_n \times \Sigma_n$, the polytope of couplings between $\mu$ and $\nu$ is defined as
\[
\Pi(\mu, \nu) := \left\{ \gamma \in \mathbb{R}_+^{n \times n} \;:\; \gamma {1}_n = \mu, \; \gamma^\top {1}_n = \nu \right\}.
\]
As expressed in \eqref{CepsLam}, the E-OT problem can be re-cast in the form $\min _{\gamma \in \mathcal{S}} \text{KL}(\gamma \mid \xi) $
where $\xi$ is a given point in $\mathbb{R}_{+}^{n \times n}$, and $\mathcal{S}$ is an intersection of  two closed convex sets, namely
$ \mathcal{S}= \mathcal{S}_{1} \bigcap   \mathcal{S}_{2}$, such that $\mathcal{S}$ has nonempty intersection with $\mathbb{R}_{+}^{n \times n}$, where  $ \mathcal{S}_1 = \left\{ \boldsymbol\gamma = (\gamma_m)_{m=1}^M \in (\Sigma_n )^M : \gamma^\top_m {1}_n = \mu_m \quad \forall m \right\}$
and $ \mathcal{S}_2 = \left\{ \boldsymbol\gamma = (\gamma_m)_{m=1}^M \in (\Sigma_n )^M : \exists \mu \in \mathbb{R}_+^{n} , \gamma_m {1}_n = \mu \quad \forall m \right\}$.
We focus on the  case where the convex sets $\mathcal{S}_{\ell}$, for $\ell=1,2$ are affine subspaces. In this case, it is possible to solve the 
KL -minimization problem by simply using iterative KL-projections. For the sake of completeness, we recall that,  given a convex set $D$ and a reference measure $\xi$, the 
KL-projection of $\xi$ onto $D$ is defined as 
$$
\mathrm{P}_{D}^{\mathrm{KL}}(\xi) := \arg\min_{\gamma \in D} \mathrm{KL}(\gamma \| \xi), \quad \mathrm{KL}(\gamma \| \xi) = \sum_{i,j} \gamma(i,j) \left( \ln\left( \frac{\gamma(i,j)}{\xi(i,j)} \right) - 1 \right). 
$$
To compute the KL-projection, we start from $\gamma^{(0)}=\xi$, and compute
$\forall t>0$,  $\gamma^{(t)} := P_{\mathcal{S}_{t\text{mod}2}}^{\mathrm{KL}}\left(\gamma^{(t-1)}\right)$. One can 
show that $\gamma^{(t)}$ converges towards the unique solution $
\gamma^{(t)} \rightarrow P_{\mathcal{S}}^{\mathrm{KL}}(\xi)$  as  $t \rightarrow \infty$; see \cite{benamou2015iterative}.

The computation of the barycenter in the E-ROBOT setting, requires  the computation of a Wasserstein barycenter of input probability measures $\mu_1, \dots, \mu_M \in \Sigma_n$ with weights $\alpha_1, \dots, \alpha_M$ (such that $\sum_m \alpha_m = 1$), using  $W_{\lambda,\varepsilon}$. To achieve this goal,
we resort on KL-projections using the truncated cost matrix $C_\lambda \in \mathbb{R}^{N \times N}$ whose $(i,j)$-entry is 
$C_\lambda(i,j) = \min(C(i,j), 2\lambda)$, where $C({i,j})= d(x_i, y_j)$, and the associated Laplace kernel $
\xi_k =  \xi= k_{\lambda, \varepsilon}$ as in \eqref{EqGibbs}, for every $m=1,2,...,M$. Indeed, as in the E-OT case, the E-ROBOT barycenter problem can
be re-formulated as:
\begin{equation}
\min_{ \boldsymbol\gamma} \sum_{m=1}^M \alpha_m  \, \mathrm{KL}(\gamma_m \| \xi_m), \quad \text{s.t.} \quad   \boldsymbol\gamma \in \mathcal{S}_1 \cap  \mathcal{S}_2.
\label{RobBar}
\end{equation}
Then, the Iterative Bregman Projection (IBP) scheme (see \cite{benamou2015iterative}) can be applied. 
More specifically, we first compute the projection onto the constraint set $\mathcal{S}_1$  for each $k$, then we compute the projection onto the constraint set $\mathcal{S}_2$, which enforces a shared left marginal $\mu$ across all couplings. This procedures leads to iterates $ \boldsymbol\gamma^{(t)} = (\gamma_{m}^{(t)})_{m}$ which satisfy, for each \( m \),
$\gamma_{m}^{(t)} = \mathrm{diag}(u_{m}^{(t)}) \, \xi \, \mathrm{diag}(v_{m}^{(t)})$
for two vectors \( (u_{m}^{(t)}, v_{m}^{(t)}) \in \mathbb{R}^n \times \mathbb{R}^n \), initialized as \( v_{m}^{(0)} = {1}_n \) for all \( m \), and computed with the iterations:
\[
u_{m}^{(t)} = \frac{\mu^{(t)}}{\xi v_m^{(t)}}, 
\quad 
v_m^{(t+1)} = \frac{\mu_m}{\xi^\top u_m^{(t)}},
\]
where \( \mu^{(t)} \) is the current estimate of the barycenter obtaied as:
$
\mu^{(t)} = \prod_{m=1}^N \left( u_m^{(t)} \odot (\xi v_m^{(t)}) \right)^{\alpha_m}.
$
Operations are to be interpreted element-wise, namely, for vectors $(a, b) \in \mathbb{R}^n \times \mathbb{R}^n$, we denote entry-wise multiplication and division as
$$
a \odot b := (a_i b_i)_i \in \mathbb{R}^n \quad \text{and} \quad \frac{a}{b}  :=  (a_i/b_i)_i \in \mathbb{R}^n.
$$
These alternating projections  are repeated until convergence: the resulting shared left marginal $\mu$ is the robust entropic barycenter. To summarize the described methodology,  in Algorithm \ref{Algm2}, we provide the pseudo code for E-ROBOT barycenters via IBP calculation.

\begin{algorithm}
\caption{E-ROBOT Barycenter via Iterative Bregman Projections (IBP)}
\label{Algm2}
\begin{algorithmic}[1]
\State \textbf{Input:}
\State \quad Input measures: $\mu_1, \mu_2, \ldots, \mu_M \in \Sigma_n$
\State \quad Weights: $\alpha_1, \alpha_2, \ldots, \alpha_M$ where $\sum_{m=1}^M \alpha_m = 1$
\State \quad Truncated cost matrix: $C_\lambda \in \mathbb{R}^{n \times n}$, where $C_\lambda(i,j) = \min(C(i,j), 2\lambda)$
\State \quad Regularization parameter: $\varepsilon > 0$
\State \quad Convergence threshold: $\delta > 0$
\State \textbf{Output:}
\State \quad Robust entropic barycenter: $\mu^* \in \Sigma_n$

\Procedure{IBP\_EROBOT\_Barycenter}{$\{\mu_m\}, \{\alpha_m\}, C_\lambda, \varepsilon, \delta$}
\State Precompute the Gibbs kernel:
\State $K \gets \exp(-C_\lambda / \varepsilon)$ \Comment{Element-wise exponentiation}

\State Initialize scaling vectors:
\For{$m = 1$ to $M$}
    \State $v_m \gets 1_n$ \Comment{Initialize to vector of ones}
\EndFor

\State $\mu \gets 1_n / n$ \Comment{Initialize barycenter uniformly}
\State $\Delta \gets \infty$  \Comment{Initialize $\Delta$ to a very large number}
\State $t \gets 0$

\While{$\Delta > \delta$}
    \State $t \gets t + 1$
    
    \State First projection: enforce marginal constraints
    \For{$m = 1$ to $M$}
        \State $u_m \gets \mu_m \slash (K v_m)$ \Comment{Element-wise division}
    \EndFor
    
    \State Second projection: enforce shared barycenter constraint
    \For{$m = 1$ to $M$}
        \State $\tilde{v}_m \gets \mu \slash (K^\top u_m)$ \Comment{Temporary update (element-wise division)}
    \EndFor
    
    \State Update barycenter estimate:
\State $\mu_{\text{new}} \gets 1_n$ \Comment{Initialize to ones}
\For{$m = 1$ to $M$}
    \State $\mu_{\text{new}} \gets \mu_{\text{new}} \odot \left(u_m \odot (K \tilde{v}_m)\right)^{\alpha_m}$ \Comment{Element-wise operations}
\EndFor
\State $\mu_{\text{new}} \gets \mu_{\text{new}} / \sum \mu_{\text{new}}$ \Comment{Normalize}

\State Update scaling vectors:
\For{$m = 1$ to $M$}
    \State $v_m \gets \mu_m \slash (K^\top u_m)$ \Comment{Direct update from constraint (element-wise division)}
\EndFor

\State $\Delta \gets \|\mu_{\text{new}} - \mu\|_1$  \Comment{Compute $L_1$ norm}
\State $\mu \gets \mu_{\text{new}}$
\EndWhile

\State $\mu^* \gets \mu$ \Comment{Assign final estimate to output variable}
\State \Return $\mu^*$
\EndProcedure
\end{algorithmic}
\end{algorithm}

In the next two subsections, we illustrate how the E-ROBOT barycenters perform on 2D and 3D shapes in the presence of anomalous records. To implement our method, we resort on the \texttt{Python} library \texttt{ot}, which contains the routine
\texttt{bregman} that computes Bregman projections for E-OT. That routine requires as an input a user-specified cost matrix. Therefore, to implement our E-ROBOT method we need to input in \texttt{ot.bregman} the matrix resulting from the application of the truncated cost  function $c_\lambda$ to the 2D and 3D data, computing
the distance via the routine \texttt{cdist} and trimming the entries of the resulting matrix via $2\lambda$.

\subsubsection{Barycenters for corrupted 2D and 3D shapes} \label{Bary}


In the next numerical example,
we illustrate the use of the IBP and E-ROBOT in the computation of barycenters for 2D images. We consider two shapes.  \textit{Shape 1 (Source)} is a red circle with a radius of $4.5$ pixels which we contaminate with 10 outliers in the top-right corner. \textit{Shape 2 (Target)} is a blue square with a side length of 9 pixels which we contaminate with 10 outliers in the bottom-right corner. The resulting images are normalized to represent probability distributions.

\begin{figure}[!h]
     \centering
     \begin{subfigure}
         \centering
         \includegraphics[width=0.89\textwidth, height=0.265\textwidth]{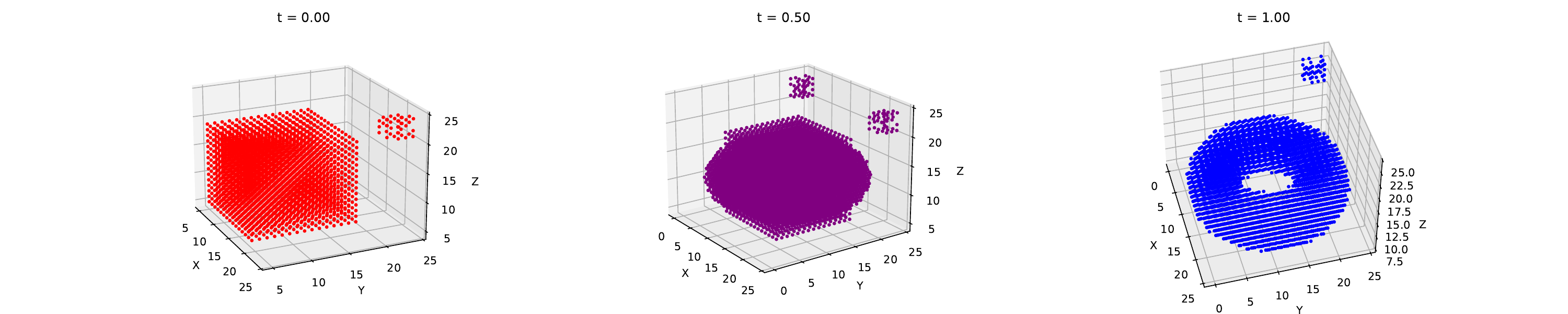}
     \end{subfigure} \\
     \begin{subfigure}
         \centering
         \includegraphics[width=0.89\textwidth, height=0.265\textwidth]{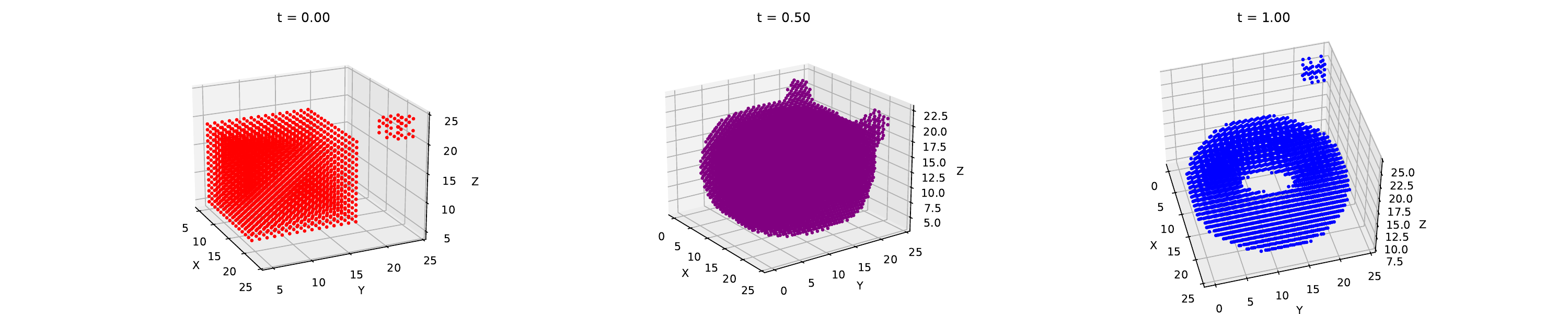}
     \end{subfigure}
        \caption{E-ROBOT Barycenters via IBP for 2D shapes:  top panels for $\lambda=4$ and bottom for $\lambda=4 \times 10^7$. Shapes are on a $32 \times 32$ grid. The entropic regularization parameter is $\varepsilon=0.15$.}
       \label{FigBary}
\end{figure}

We compute entropic barycenters, which can be interpreted as interpolated shapes between the source and target distributions. For 
the sake of visualization, we consider weights $t=0.25,0.5, 0.75$, with corresponding the weights for the barycenter calculation being $\{1-t, t\}$. We consider the E-ROBOT and the EOT. In  the EOT case, the cost matrix  is based on the Euclidean distance between pixel coordinates as implemented in the Python routine \texttt{entropic\_barycenter} in the \texttt{ot} library. To implement the E-ROBOT  barycenters, we modify this routine introducing a truncation parameter $\lambda$, which trims the entires of  this matrix. This simple modification is central to the implementation. Indeed, it allows for the comparison between two distinct methods: the standard EOT  and  our novel {E-ROBOT}.  In Figure \ref{FigBary},  we display two sets of plots, one for each  method.

The five top plots, corresponding to the E-ROBOT with a small $\lambda = 4$, show a clean barycenter computation process: the red circle smoothly transforms into the blue square, while the outliers appear to simply fade in and out at their fixed locations without being transported. This demonstrates that E-ROBOT successfully isolates the barycenters computation of the main shapes from the influence of the outliers. Differently, the five bottom plots, which is for the E-ROBOT with a large $\lambda$, illustrate that the outliers significantly impact 
the barycenter calculation: as the circle becomes the square, the red outliers from the top-right corner are transported to the location of the blue outliers in the bottom-right corner. This results in particularly noticeable at $t=0.5$, where a blurred mass appears to be streaking between the outlier locations. This is due to the fact that  
the E-ROBOT with a large $\lambda$  behaves like the standard EOT approach: it attempts to transport all mass, including the anomalous records, as part of the overall transformation.

\begin{figure}[!h]
     \centering
     \begin{subfigure}
         \centering
         \includegraphics[width=0.9\textwidth, height=0.25\textwidth]{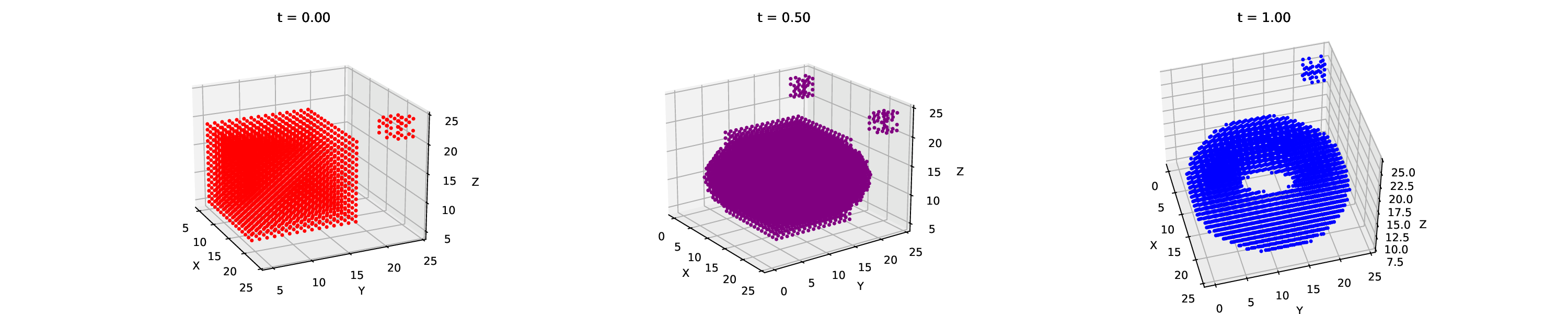}
     \end{subfigure} \\
     \begin{subfigure}
         \centering
         \includegraphics[width=0.9\textwidth, height=0.25\textwidth]{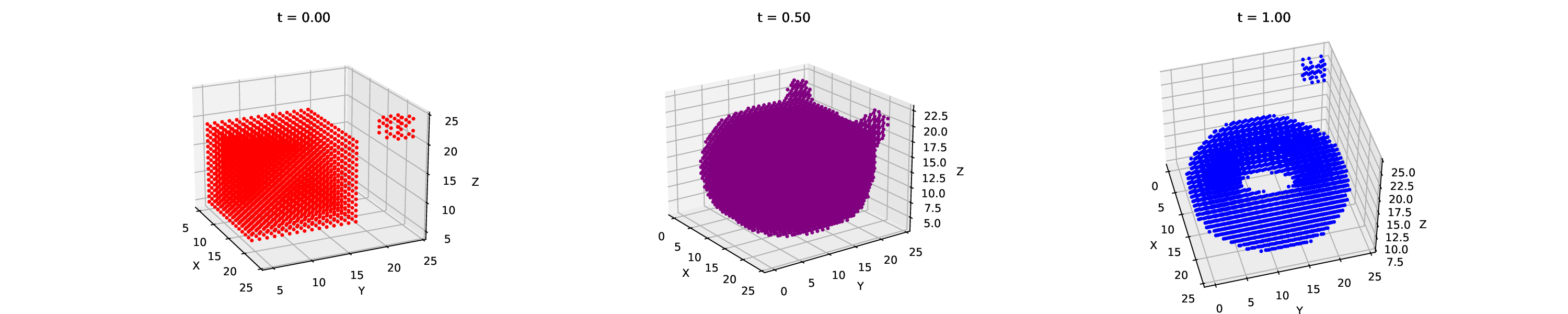}
     \end{subfigure}
        \caption{ E-ROBOT barycenters via IBP for 3D shapes, for weights $t=0,0.5,1$: top 3 panels are for $\lambda=0.1$ and bottom 3 panels are $\lambda=10$). Shapes are on a $22 \times 22 \times 22$ grid. The entropic regularization parameter is $\varepsilon=0.05$.}
       \label{FigBary3D}
\end{figure}

Similar considerations hold for the case of 3D shapes, where we compute  the barycenter for values of $t=0,0.5,1$, between a cube and torus. The original shapes are contaminated by outliers, mimicking the logic as in the 2D case. In Figure \ref{FigBary3D}, we display the results for the E-ROBOT with small $\lambda$  (three top plots) and the E-ROBOT with large $\lambda$ (three bottom panels). Also in the 3D case, the key point is that in the E-ROBOT with large $\lambda$, the IBP algorithm considers the transport of both the shapes and the outliers, treating them as part of a single, coherent distribution. In contrast, the E-ROBOT  with small $\lambda$  limits the maximum cost of transporting mass, particularly between the far-apart main shapes and the outliers. This is evident comparing the top and bottom middle plots. 

\subsection{Gradient flows for corrupted 2D shapes}\label{sec.GradFlow}

Let $\mu \in \mathcal P(X)$ be a probability measure on a compact set $X \subset \mathbb R^d$.
Gradient flows are the continuous-time analogue of gradient descent: they describe how a probability distribution $\mu_t$ evolves with time $t$. 
In optimal transport, this evolution takes place in the Wasserstein space of probability measures. Such flows appear naturally in machine learning, for example in {generative modeling} (evolving a model distribution toward data), {domain adaptation} (aligning source and target), and {clustering/density estimation} (regularizing or smoothing empirical measures).

To compare the de-biased robust Sinkhorn divergence $\overline{W}_{\varepsilon,\lambda}$ with the standard Sinkhorn-type loss for $W_1$, a natural experiment is to let a model distribution $\mu_t$ evolve along the E-ROBOT gradient flow of a data-fit loss that drives it toward a target distribution (\cite{S15}). This non-parametric fitting setup makes explicit the role of the robustness parameter $\lambda$ in the presence of outliers. Classical EOT flows suffer from well-documented artefacts: an {entropic shrinkage bias} (working on the entropic regularized OT alone pulls $\mu$ toward an over-concentrated measure), and -- on the other end of the spectrum -- kernel MMDs can show {vanishing gradients} near the support’s extremes; see  \cite{feydy2019}. These issues motivate the use of a robust, regularized: $\overline{W}_{\varepsilon,\lambda}$. 

More in detail,  recall the E-ROBOT negentropy functional $F_{\varepsilon, \lambda}(\mu)$ in \eqref{EqNegEntr}. This functional is strictly convex and differentiable; when $\mu=\nu$, the Schrödinger potentials coincide, $\phi^* = \psi^*$. Consequently, the (first-variation) gradient of the E-ROBOT negentropy reads
$\nabla F_{\varepsilon, \lambda}(\mu) = -\phi^*/2,$
and the corresponding Wasserstein gradient flow is $
{d\mu_t}/{dt} = -\,\nabla F_{\varepsilon, \lambda}(\mu_t).$
To simulate this continuous-time PDE, one may resort on  the explicit (forward) Euler scheme with step size $\tau>0$, where
the current distribution is pushed forward by a small displacement along the {negative} gradient field. If $\mu_k = n^{-1}\sum_{i=1}^n \delta_{x_i^k}$ is an empirical measure at the $k$-th step, 
this reduces to particle updates of the familiar gradient descent form $
 x_i^{k+1} =  x_i^{k} - \tau\,\nabla_{x_i} F_{\varepsilon, \lambda}(\mu_k),$ for $i=1,\dots, n.$
In \texttt{Python} code (as in the \texttt{GeomLoss} package), one evaluates the loss with parameters $(\varepsilon,\lambda)$, back-propagates to obtain $\nabla_{x_i}F_{\lambda,\varepsilon}$, and applies the Euler step above to all particles. The step size $\tau$ controls stability and small values of $\tau$ imply slower but more faithful to the continuous flow. This explicit discretization exactly mirrors the procedure used in Section~4 of \cite{feydy2019}, which contrasts kernel MMD, biased entropic OT, and the de-biased Sinkhorn divergence on toy registration and gradient flow tasks. 

\begin{figure}[!h]
     \centering
     \begin{subfigure}
         \centering
         \includegraphics[width=0.97\textwidth, height=0.6\textwidth]{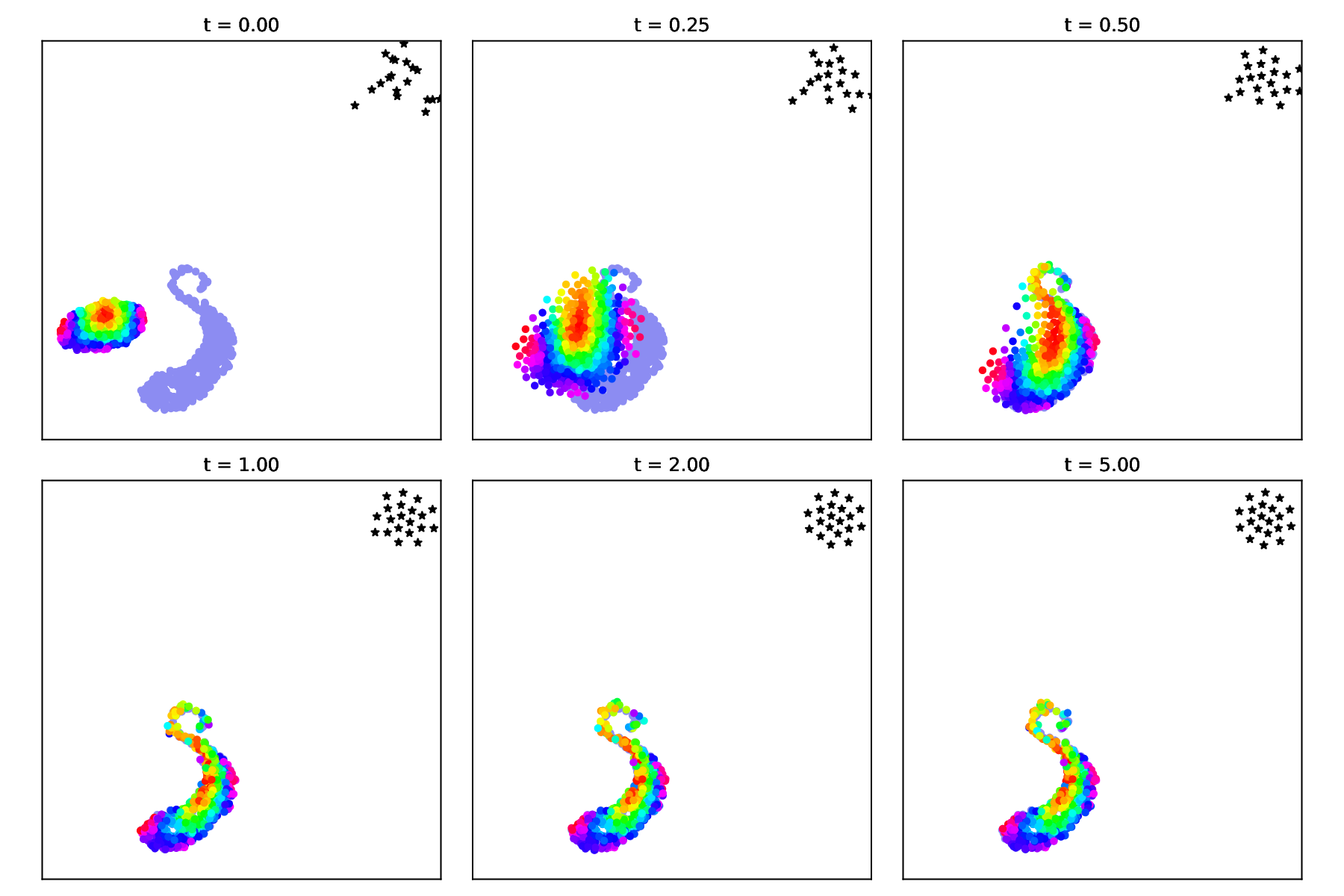}
     \end{subfigure} \\
     \begin{tabular}{cc}
         \includegraphics[width=0.45\textwidth, height=0.35\textwidth]{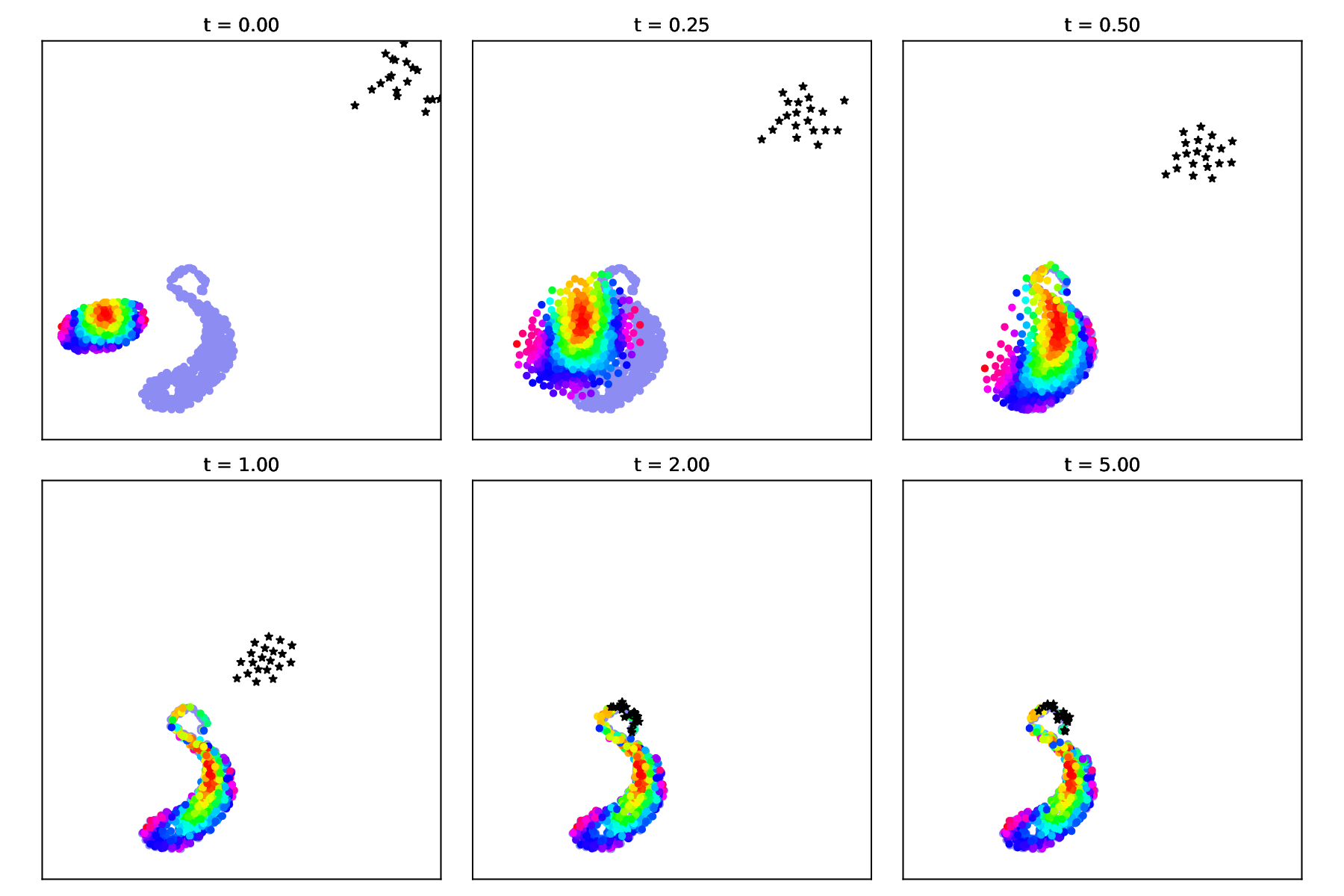} &
         \includegraphics[width=0.45\textwidth, height=0.35\textwidth]{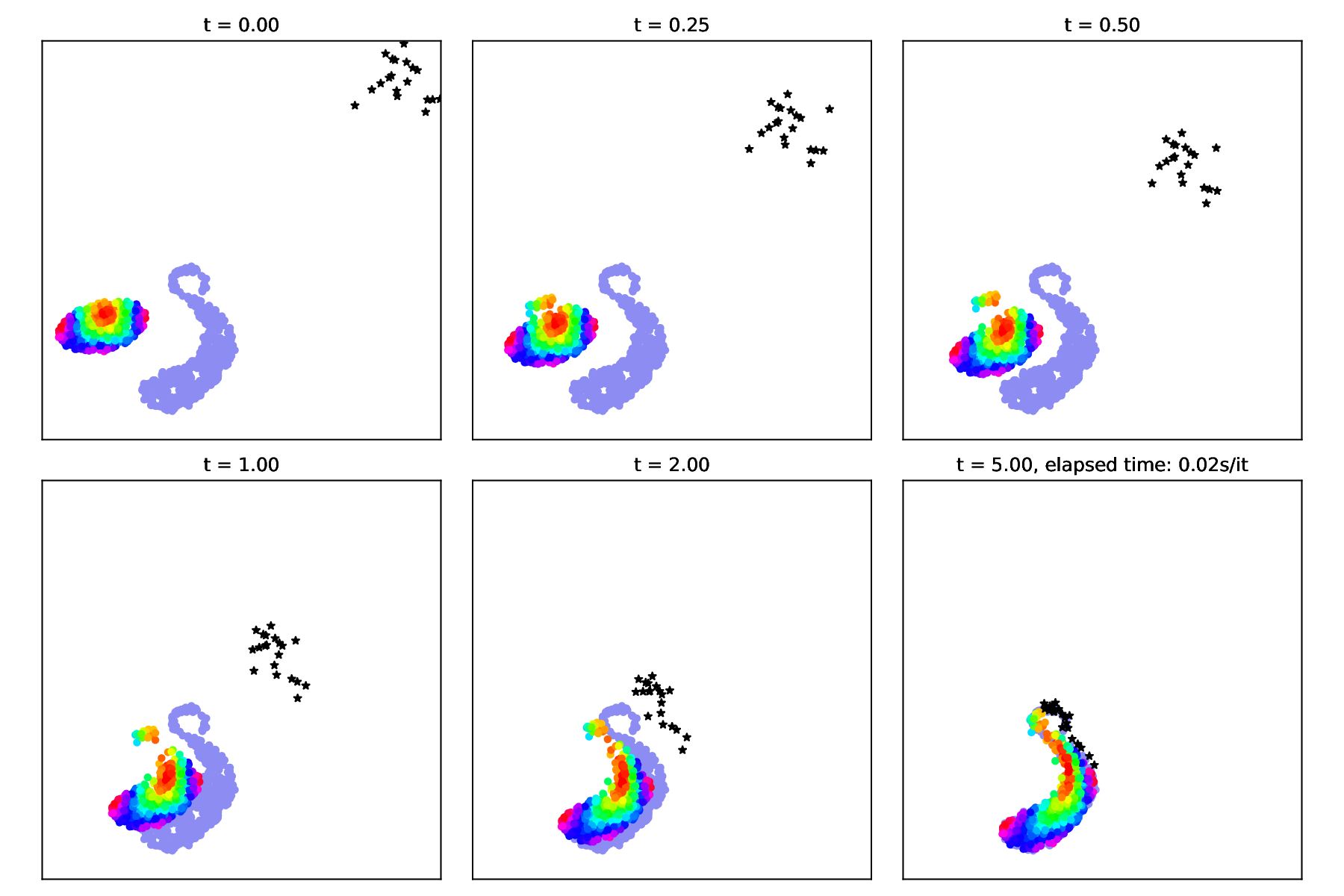}
     \end{tabular}    
        \caption{Gradient flows for 2D shapes via entropic regularized OT. Top panels:  E-ROBOT with $\lambda=0.6$, bottom  left 6 panels E-ROBOT  with $\lambda=6$, and bottom right 6 panels 
        EOT with $W_2$. The entropic regularization parameter is $\varepsilon=0.05$. The learning rate, i.e. time step $\tau$, is set equal to $0.05$, similarly to the 
        default value in \texttt{GeomLoss}.}
       \label{FigFlows}
\end{figure}

For the sake of visualisation and following the existing examples available in the \texttt{GeomLoss}, we focus on 2D shapes. In Figure \ref{FigFlows} we display the  gradient flows for the E-ROBOT with a small value of $\lambda$ (top 6 panels) and a with large $\lambda$ (bottom panels).
The plots showcase how the cost truncation parameter $\lambda$ controls the E-ROBOT method's robustness to outliers.
For a large value of $\lambda$ (bottom panels), the E-ROBOT approach behaves similarly to standard E-OT because the cost truncation is effectively removed. In this case, the gradient flow illustrates that the outliers (black stars) are transported and embedded into the central blob of mass as the shapes evolve. This is visually evident as the black stars, initially separate, are drawn into the evolving mass for $t\geq 0.5$ (they are getting closer and closer to the blobs). This aspect highlights the sensitivity of standard EOT to anomalous data points. This is evident also in the bottom panels, which are for the EOT with $W_2$, as obtained using the command \texttt{gradient\_flow} in \texttt{GeomLoss}---which
remains the faster approach, as indicated by the time for each iteration. In contrast, for a small $\lambda$ (top panels), the truncated cost function limits the maximum transport cost between far-apart points. The gradient flow for small $\lambda$ demonstrates the E-ROBOT method's robustness: the flow of the main blob proceeds without being influenced by the distant outliers. 
The gradient flow effectively excludes these outliers from the optimal transport plan, allowing the algorithm to focus on the smooth transformation of the core blobs. This behaviour confirms that the E-ROBOT method successfully mitigates the influence of outliers by preventing them from being part of the transport. However,
an important remark is in order. A closer inspection of the gradient flow in the top panels of Figure \ref{FigFlows} reveals a subtle but important phenomenon: the outliers are not perfectly stationary from the very beginning. For early timesteps ($t=0.25$ or $t=0.5$), they exhibit some minor movement before settling into a fixed configuration. This initial movement is a natural consequence of the {gradient descent} optimization process. The algorithm, starting from the initial configuration at $t=0$, is in the process of finding the optimal transport plan. At this early stage, there is a weak but non-zero influence from the distant points (the main blob) on the outliers. The gradient flow is essentially taking small, exploratory steps in the direction of steepest descent for the overall system. However, the core mechanism of the E-ROBOT method -- the {truncated cost function} -- quickly takes precedence. The algorithm realises that the cost of moving the outliers is prohibitively high due to the small $\lambda$ value: this leads to outliers motion becoming negligible and the gradient flow for the outliers effectively vanishes as the algorithm converges on the optimal, outlier-excluding solution. 

Referring to the plots in Appendix B, 
the key conclusion from the experiment is that the use of $\overline{W}_{\lambda,\varepsilon}$ is preferable to the MMDs for gradient flow.
 Indeed, $\overline{W}_{\lambda,\varepsilon}$ provides a more reliable, namely a more geometry aware, flow and more robust gradient signal for moving probability distributions toward each other, especially when they are initially far apart and contain outliers.

\subsection{Image color transfer}

\begin{figure}[h!]
    \centering
    \begin{subfigure}
        \centering
        \includegraphics[width=0.8\textwidth, height=0.45\textwidth]{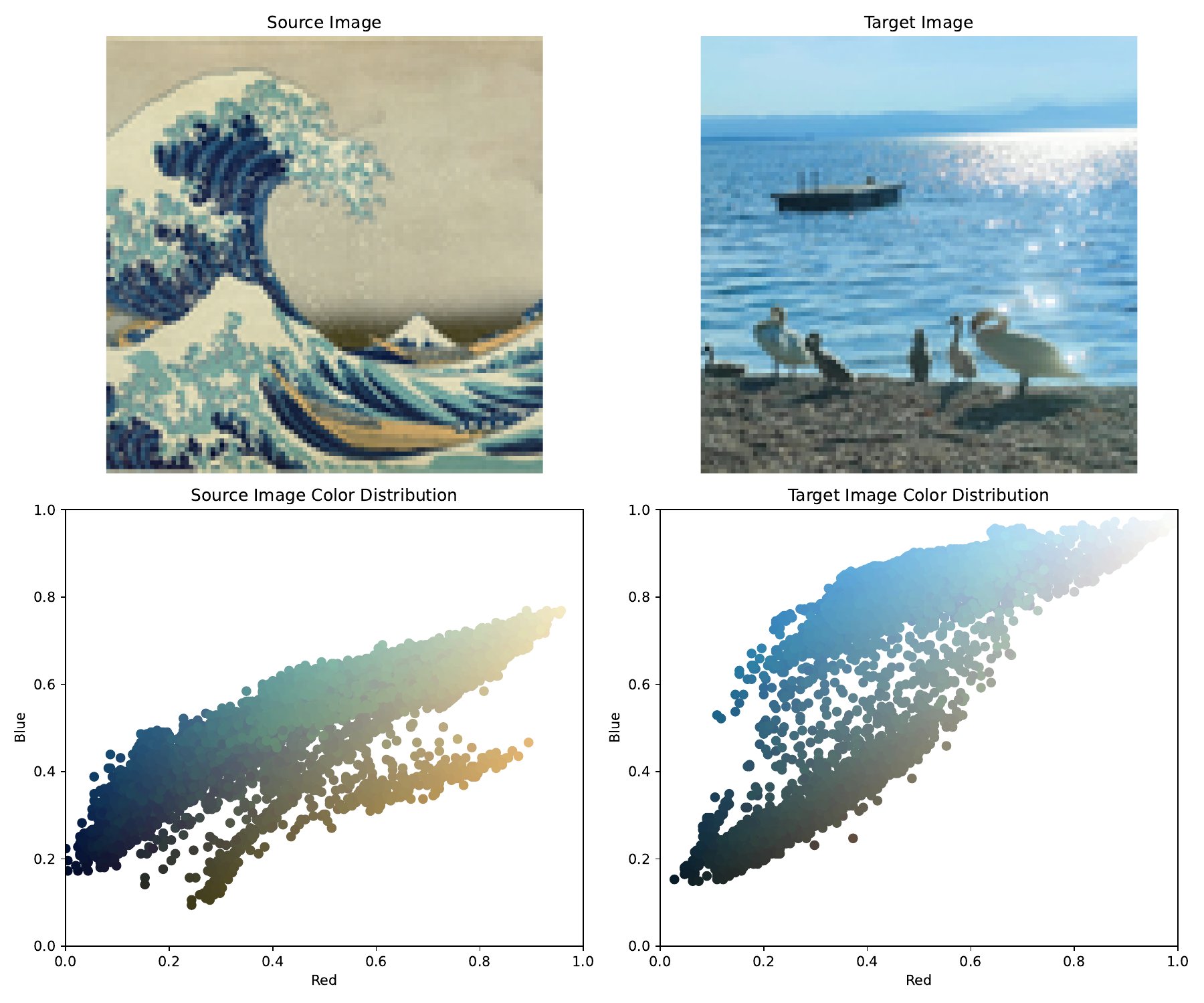}
        \caption{Source (left) and target (right) distributions. Source: \textit{The Great Wave off Kanagawa} by Hokusai, picture credit: Wikipedia. Target: A scene on 
        Lac Léman, Switzerland, picture credit: Nadia La Vecchia.}
        \label{fig:originals}
    \end{subfigure}
    \hfill
            \centering
    \begin{subfigure}
    \centering
    
        \includegraphics[width=0.8\textwidth,  height=0.45\textwidth]{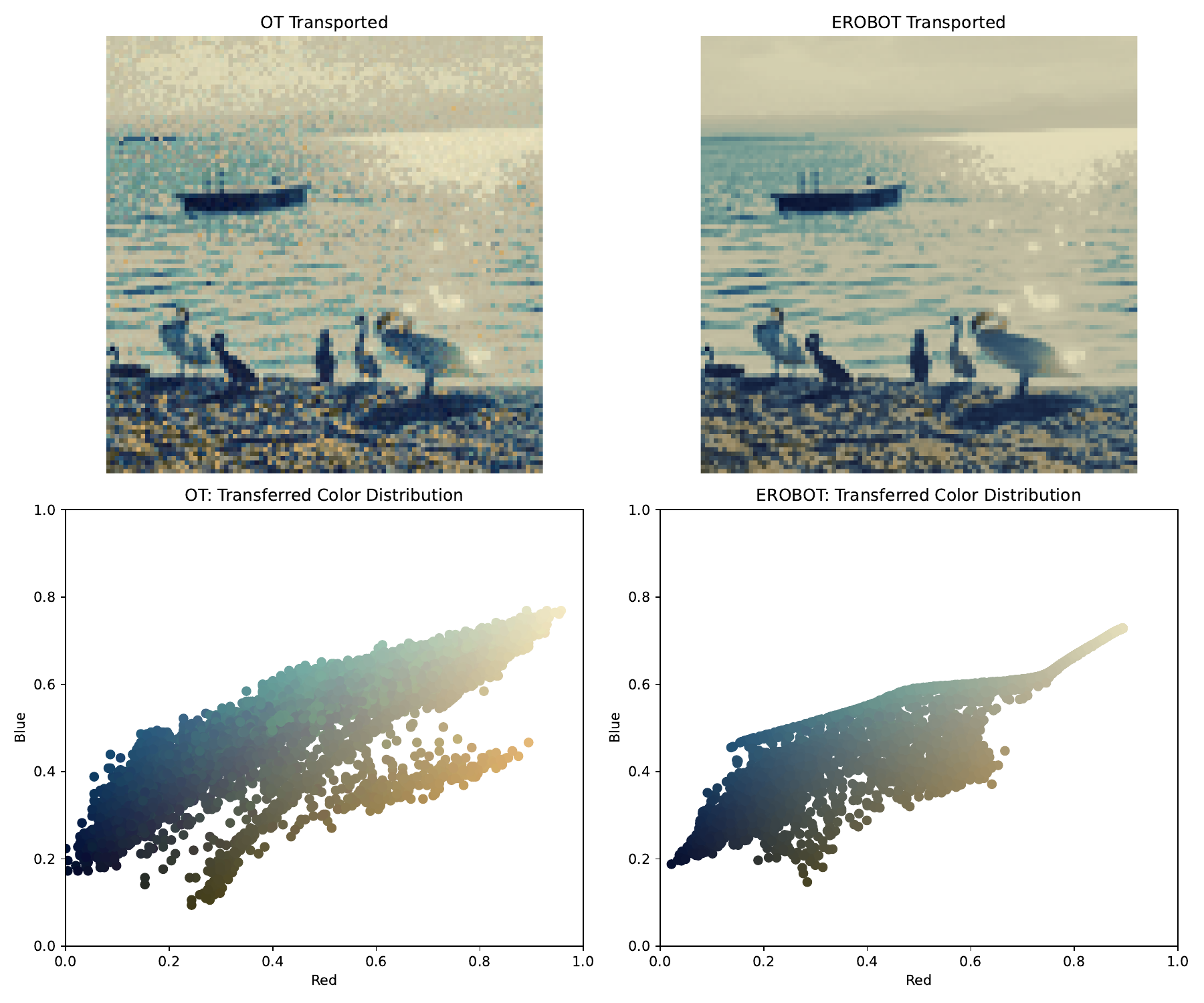}
        \caption{Results of color transfer. Top panel: Output images for OT (left) and E-ROBOT (right), with $\lambda=20$ and $\varepsilon=0.01$. Bottom: Their corresponding color distributions.}
        \label{fig:transferred}
    \end{subfigure}
    \label{fig:color_transfer}
\end{figure}
Color transfer is a fundamental and highly applicable task in computer vision and computer graphics. Its importance stems from its role as a core problem in image manipulation and its wide range of practical applications. For instance, it can be used for aesthetic enhancement of photographs and for accessibility purposes, such as modifying color schemes for color-blind viewers or restoring the faded color palette of historical images. 

In machine learning, from a computational perspective, color transfer serves as a canonical example of distribution alignment. Indeed, the problem of matching the color distribution of a source image to a target is analogous to aligning the feature distributions of two datasets. Methods that solve color transfer robustly and in large-dimensions, like E-ROBOT, provide valuable insights and tools for this machine learning task. With this regard, it is well-known that the regularization of the transport plan helps to remove colorization artifacts due to noise amplification. More precisely, both OT and ROBOT map between complicated densities can be non smooth. As a result, using directly their transport plans to perform color transfer  amplifies the noise in flat areas of the image (namely, it creates colour artifacts). We refer to \cite{ferradans2014regularized} for a discussion.

We demonstrate the practical efficacy of E-ROBOT on a color transfer task, adapting the palette of Hokusai's \textit{The Great Wave off Kanagawa} (source image) to that of a scene on Lake Léman (target image), in the Geneva (Switzerland) area; see Figure \ref{fig:originals} for the pictures (top panels), with the corresponding color distributions (bottom panels), as obtained using a Red-Blue representation.  The ultimate goal  is to impose on the target image the colours of Hokusai's paint, while preserving the smoothness of both images and avoiding the creation of artifacts.

The bottom plots of Figure \ref{fig:originals} highlight that the source image  contains some {colours} that pose significant challenges for traditional OT. For instance, there are  extreme pigment variations (the vibrant Prussian blue used in the woodblock print creates saturated blue values that fall outside the natural color distribution of the target landscape photograph) or compositional elements (the distinctive white foam splashes and the yellow boat in Hokusai's work create isolated color clusters that don't correspond to any elements in the target image).

The results in Figure~\ref{fig:transferred} reveal striking differences in how OT and E-ROBOT handle these challenges. The OT result exhibits notable artifacts, including a speckled noise pattern indicative of unstable pixel assignments, evidenced by a persistent secondary peak in the red region of its distribution. This secondary peak represents OT's attempt to {overfit to red/yiellow colours}, forcibly matching anomalous source colours to inappropriate target regions. The resulting distribution shows issues in the red channel, where the artificial peak creates unnatural tones and distorted color balances.

In contrast, the E-ROBOT transformation (right panels) demonstrates remarkable robustness. The resulting color distribution illustrates that the robust Sinkhorn divergence effectively {downweights the influence of extreme  colours}. The algorithm recognizes that certain source colours (like the intense Prussian blue or extreme tones of yiellow) entail large transport cost, and thus prevents them from distorting the overall color mapping.  This robustness is visually apparent in the  transformed image.  For instance, the sky tones transition naturally from Hokusai's distinctive palette to the target's atmospheric blues without the artificial color casts that plague the OT result via a number of grey untied dots. Moreover, thanks to the regularization, the E-ROBOT features smoothness in the final picture, confirming its statistical superiority for tasks involving real-world images with complex color distributions and inherent outliers.

\section{Conclusion and possible developments} \label{conclusion}

We introduce the E-ROBOT framework, which combines the outlier robustness of ROBOT with the computational efficiency of entropic regularization via the Schr\"odinger bridge. The resulting robust Sinkhorn divergence \(\overline{W}_{\varepsilon,\lambda}\) achieves a dimension-free sample complexity of \(\mathcal{O}(n^{-1/2})\), overcoming the curse of dimensionality. We demonstrate its applicability in high-dimensional, heavy-tailed settings for tasks like goodness-of-fit testing, barycenter computation, and gradient flows. E-ROBOT can be implemented with simple modifications to existing optimal transport algorithms.

Besides these theoretical and methodological results, other future developments can be envisioned. Hereunder we mention some of them.

(i) \textit{Parametric inference.} The robust Sinkhorn divergence \(\overline{W}_{\varepsilon,\lambda}\) can be applied as loss function to conduct parametric inference in statistics similarly to MKE (see \cite{bassetti2006minimum}) or in generative modeling (see \cite{genevay2018learning}). Specifically, as in \cite{Maetal25}, one may define a parametric model as  $ \{\mu_{\theta}, \ \theta \in \Theta \subset \mathbb{R}^q, \ q\geq1\}$. Then, the 
{minimum robust Sinkhorn estimator} is defined as
$$	
\hat{\theta}_n^{\lambda,\varepsilon} =\underset{\theta \in \Theta}{\operatorname{argmin}} \ \overline{W}_{\varepsilon,\lambda}\left(\hat{\mu}_n, \mu_\theta\right). 
$$
When there is no explicit expression for the probability measure characterizing the  parametric model (e.g. in complex generative models), the computation of $\hat{\theta}_n^{\lambda,\varepsilon} $  can be difficult. To cope with this issue,  one may think of using the {minimum expected robust Sinkhorn estimator} 
defined as
\begin{equation*}\label{MERWE}
	\hat{\theta}_{n, m}^{\lambda,\varepsilon} =\underset{\theta \in \Theta}{\operatorname{argmin}} \ {\mathbb E}_m \left[  \overline{W}_{\varepsilon,\lambda}\left(\hat{\mu}_n, \hat{\mu}_{\theta, m}\right)\right],
\end{equation*}
where the expectation $ \mathbb{E}_m $  is taken over  the distribution $ \mu^{(m)}_{\theta} $.  
To implement $\hat{\theta}_{n, m}^{\lambda}$ one can rely on  Monte Carlo methods and approximate numerically
$ {\mathbb E}_m [\overline{W}_{\varepsilon,\lambda}\left(\hat{\mu}_n, \hat{\mu}_{\theta, m}\right)] $. The existence, measurability, and
consistency of the resulting estimator should be proved as in \cite{bernton2019parameter}. Moreover, since the sample complexity of $\overline{W}_{\varepsilon,\lambda}$ scales with $n^{-1/2}$, we conjecture that it is possible to prove  root-$n$ consistency and asymptotic normality of $\hat{\theta}_{n, m}^{\lambda}$, similarly to the results for $W_2$ in \cite{del2019central}---a conjecture that does not make sense for OT- and ROBOT-based estimators in multivariate setting.  For the resulting estimators, one may think also of deriving small-sample approximation to their distributions via saddlepoint techniques, exploring connections in the setting of dependent data (as in \cite{LVR19,JLVRS23}) or of independent data (as in \cite{la2022some}).


(ii) \textit{GoF via Bregman-type divergence and MMD.}  Following the same logic as the GoF test described in \S \ref{GoF}, one can also define  test statistics using either the Hausdorff divergence in \eqref{Hd} or the MMD using $k_{\lambda,\varepsilon}$. In the case of simple hypothesis, the distribution of these statistics
can be obtained via Monte Carlo methods, as in \S \ref{GoF}. Moreover, one may think of considering also composite hypotheses and derive the distribution of the test statistics via bootstrap methods, whose statistical guarantees may be proved building on the results in Theorem \ref{Thm1} and on \cite{klatt2020empirical}. The resulting approach based on $\overline{W}_{\varepsilon,\lambda}$ may offer an alternative to the recent developments in  \cite{hu2025two}, where the max-sliced Wasserstein distance in applied.

(iii)  \textit{Relaxing the assumptions.} Some of our results of this paper assume compact support: this is common in the OT literature for technical convenience.  With this regard, we notice that some of the results in the paper (like Proposition \ref{Prop1} and \ref{PropDual}) already hold for noncompact spaces. Thus, one may think of relaxing the compactness assumption. In addition, we highlight that many of our theoretical and methodological results  can be extended to the case where we use a generic cost function is $\varepsilon^{-1} c_{\lambda}$, with $c_{\lambda}=\tilde{d}^p$ and $\tilde{d}_{\lambda} = \min(d,2\lambda)$. The use of this cost function could lead to an interesting research topic: extending ${W}_{\varepsilon,\lambda}$ to ${W}_{\varepsilon,\lambda, p}$, analogous to the order-$p$ Wasserstein distance $W_p$ in classical optimal transport. Both these extensions require modified assumptions and suitable modifications of our proofs.


(iv) \textit{Selection of $\lambda$ and $\varepsilon$.} A key challenge for practitioners is the joint selection of the robustness parameter $\lambda$ and the entropic regularization parameter $\varepsilon$. We note that this problem is not only open but also fundamentally new, as the literature on OT offers limited guidance even for choosing these parameters individually. The selection of $\varepsilon$ in E-OT is often heuristic (see e.g. \cite{goldfeld2020gaussian}), and robust OT methods frequently lack a general data-driven procedure for $\lambda$ (see e.g. \cite{NGC22}). Therefore, the joint calibration required for E-ROBOT operates on new ground and highlights a clear gap in the current literature that we are planning to fill. While beyond the scope of this paper, which introduces and validates the E-ROBOT framework itself, we conjecture that a solution could be based on a concentration inequality derived from the principles in Theorem~\ref{Thm1}. We posit this as a critical avenue for future work.

\bibliography{biblio}
\end{document}


\title{Supplementary material for:``E-ROBOT: a dimension-free method for  robust statistics and machine learning via  Schrödinger bridge" }


\author*[1]{\fnm{Davide} \sur{La Vecchia}}\email{davide.lavecchia@unige.ch}
\author[2]{\fnm{Hang} \sur{Liu}}\email{hliu01@ustc.edu.cn}


\affil*[1]{\orgdiv{Geneva School of Economics and Management}, \orgname{University of Geneva}, \orgaddress{\street{Bld. du Pont d'Arve}, \city{Geneva}, \postcode{CH-1211}, \state{Switzerland}}}

\affil[2]{\orgdiv{Department of Statistics and Finance, School of Management}, \orgname{University of Science and Technology of China}, \orgaddress{\street{Jinzhai Rd}, 
\city{Hefei}, \postcode{ 230026}, \state{Anhui Province}, \country{China}}}

\maketitle

\begin{appendices}
\section{Proofs } \label{App}

\subsection{Preliminary lemmas}

Let us recall first some basic definitions from \cite{CP19}. 
A symmetric function $k$ (resp., $\varphi$ ) defined on a set ${X} \times {X}$ is said to be positive (resp., negative) definite if for any $n \geq 0$, $x_1, \ldots, x_n \in X$, and vector $r \in \mathbb{R}^n$ the following inequality holds:
\begin{equation}
\sum_{i, j=1}^n r_i r_j k\left(x_i, x_j\right) \geq 0, \quad\left(\text { resp. } \quad \sum_{i, j=1}^n r_i r_j \varphi\left(x_i, x_j\right) \leq 0\right). \label{Eqpos}
\end{equation}
The kernel is said to be conditionally positive if positivity only holds in (\ref{Eqpos}) for zero mean vectors $r$ (i.e. such that $\left\langle r, {1}_n\right\rangle=0$), where ${1}_n$ is a n-dimensional vector of ones. If $k$ is conditionally positive, one defines the following norm:
\begin{equation}
\|\mu \|_k^2 \stackrel{\text { def. }}{=} \int k(x, y) \mathrm{d} \mu(x) \mathrm{d} \mu(y). \label{MMDnorm}
\end{equation}

These norms are often referred to as Maximum Mean Discrepancy (MMD). 
%
 Let us recall that the reproducing kernel Hilbert space (RKHS) \( \mathcal{H}_{k} \) associated with \( k \) is defined as the completion of the linear span of functions \( \{ k(x, \cdot) : x \in X \} \) under the inner product
$
\langle f, g \rangle_{\mathcal{H}_{k}} := \sum_{i,j} \alpha_i \beta_j k (x_i, x_j),
$
for \( f = \sum_i \alpha_i k (x_i, \cdot) \), \( g = \sum_j \beta_j k (x_j, \cdot) \). Moreover, we recall that,  according to \cite{sriperumbudur2011universality} (Section 3.2, p. 2399), a continuous kernel $k$ on a compact metric space $X$ is {\it $c$-universal} if the following hold: (a) $k(x,x) > 0$ for all $x \in X$, (b) there exists an injective feature map $\Phi : X \to \mathcal{H}$ such that $k(x,y) = \langle \Phi(x), \Phi(y) \rangle$. Finally, we recall that
a kernel $k : X \times X \to \mathbb{R}$ is said to be \emph{characteristic} if the associated RKHS embedding of probability measures is injective. That is, for any two Borel probability measures $\mu$ and $\nu$ on $X$,
\[
\mu \neq \nu \quad \iff \quad \int k(x, \cdot) \, d\mu(x) \neq \int k(x, \cdot) \, d\nu(x).
\]
Characteristicness ensures that the RKHS embedding of probability measures is injective, i.e., the kernel can distinguish between different probability distributions. 

\begin{lemma}  \label{LemmaPosUniv}
Let $c_\lambda$ be the ROBOT cost function defined on $X \times X$, with $X$ being a compact subset of $\mathbb{R}^d$. Then for $\epsilon >0$, the kernel $k_{\epsilon,\lambda}(x,y) = \exp(-c_\lambda(x,y)/\epsilon)$ is  positive $c$-universal and characteristic.
\end{lemma}

\begin{proof}
Let $X$ be compact and assume that $c_\lambda : X \times X \to [0, \infty)$ is continuous, symmetric, and satisfies $c_\lambda(x,y) = 0$ if and only if $x = y$.  It is trivial to show that the function induces a kernel $c_\lambda$ which is conditionally positive definite---indeed, this is obtained by a symmetric truncation of the energy distance kernel, see e.g. \cite{feydy2019}, \S 1.1. Moreover, the kernel
$k_{\epsilon,\lambda}(x,y) = \exp(-c_\lambda(x,y)/\epsilon)$
is continuous and strictly positive on $X \times Y$, since $c_\lambda(x,y) \geq 0$ and $c_\lambda(x,y) = 0$ if and only if $x = y$ implies $k_{\epsilon,\lambda}(x,y) = 1$ if and only if $x = y$, and $k_{\epsilon,\lambda}(x,y) < 1$ otherwise.. So, we define the canonical feature map \( \Phi : X \to \mathcal{H}_{k_\epsilon} \) by \( \Phi(x) := k_\epsilon(x, \cdot) \). To show that \( \Phi \) is injective, suppose \( x \neq y \). Then \( c_\lambda(x,y) > 0 \) implies \( k_\epsilon(x,y) < 1 \), while \( k_\epsilon(x,x) = k_\epsilon(y,y) = 1 \). Therefore, \( k_{\epsilon,\lambda} (x, \cdot) \neq k_\epsilon(y, \cdot) \) in \( \mathcal{H}_{k_\epsilon} \), and hence \( \Phi(x) \neq \Phi(y) \). Thus, \( \Phi \) is injective. Next,
we verify that the conditions for $c$-universality are satisfied by  $k_{\epsilon,\lambda}$: since $k_{\epsilon,\lambda}(x,x) = \exp(-c_\lambda(x,x)/\epsilon) = \exp(0) = 1$ for all $x \in X$, we have $k_{\epsilon,\lambda}(x,x) > 0$; as shown above, the canonical feature map $\Phi(x) := k_{\epsilon,\lambda}(x, \cdot)$ is injective.
Therefore, $k_{\epsilon,\lambda}$ is $c$-universal, namely, by definition, this means the RKHS \( \mathcal{H}_{k_\epsilon} \) is dense in \( C(X) \), the space of continuous real-valued functions on \( X \), equipped with the uniform norm. 
Moreover, as stated still in section 3.2. of \cite{sriperumbudur2011universality}, when $X$ is compact, $c$-universality implies that the kernel is characteristic. Hence, the RKHS induced by $k_{\epsilon,\lambda}$ is characteristic and dense in $C(X)$, thus it  is positive $c$-universal.
\end{proof}

%


Moreover, we state the following
\begin{lemma} \label{PropAlternRepr}
Let \( X \subset \mathbb{R}^d \) 
and \( c_\lambda : X \times X \to \mathbb{R} \) be the ROBOT cost function. Let \( \mu \in \mathcal{P}(X) \) be a probability measure. Then, for \( \varepsilon > 0 \), the E-ROBOT negentropy
$F_{\varepsilon, \lambda}(\mu)$
admits the  representation:
\begin{equation}
\frac{1}{\varepsilon} F_{\varepsilon, \lambda}(\mu) + \frac{1}{2} = \inf_{\xi \in \mathcal{P}(X)} \left\{ \int \ln \left( \frac{d\mu}{d\xi} \right) d\mu + \frac{1}{2} 
\Vert \mu \Vert^2_{k_{\varepsilon,\lambda}} \right\}
\label{my18}
\end{equation}
\end{lemma}

\begin{proof} The proof follows along the same lines as in Appendix B.5 of \cite{feydy2019}, to which we refer. Here we sketch its main steps, illustrating the pivotal role of $c_\lambda$. By definition, the E-ROBOT negentropy is
\[
F_{\varepsilon, \lambda}(\mu) := - \frac{1}{2} \inf_{\pi \in \Pi(\mu, \mu)}  C_\varepsilon(\mu, \mu, c_\lambda, \pi),
\]
where the entropic cost is given by
\begin{equation*}
C_\varepsilon(\mu, \mu, c_\lambda, \pi) = \int c_\lambda(x,y) \, d\pi(x,y) + \varepsilon H(\pi \| \mu \otimes \mu). \label{DualFormProof}
\end{equation*}
By standard duality results and symmetry of the potentials, we have that for $\mu \equiv \nu \Rightarrow \varphi^*=\psi^\ast =: f$, this cost admits the dual formulation:
\begin{equation}
C_\varepsilon(\mu, \mu, c_\lambda, \pi) = \sup_{f \in L^1(\mu)} \left\{ 2 \int f(x) \, d\mu(x) - \varepsilon \iint \exp\left( \frac{f(x) + f(y) - c_\lambda(x,y)}{\varepsilon} \right) d\mu(x) d\mu(y) + \varepsilon \right\}. \label{DualFormProof}
\end{equation}
Hence, the negentropy admits the dual formualtion:
\[
F_{\varepsilon, \lambda}(\mu) = -\frac{1}{2} \sup_{f \in  L^1(\mu)} \left\{ 2 \int f(x) \, d\mu(x) - \varepsilon \iint \exp\left( \frac{f(x) + f(y) - c_\lambda(x,y)}{\varepsilon} \right) d\mu(x) d\mu(y) + \varepsilon \right\}.
\]
Now, we consider the kernel $k_{\varepsilon,\lambda}(x,y)$ 
and perform the change of variable
\begin{equation}
f(x) = \varepsilon \ln \left( \frac{d\xi}{d\mu}(x) \right)
\quad \Rightarrow \quad d\mu(x) = e^{-f(x)/\varepsilon} d\xi(x). \label{ChVar}
\end{equation}
Then,
\[
\int f(x) \, d\mu(x) = \varepsilon \int \ln \left( \frac{d\xi}{d\mu}(x) \right) d\mu(x)
= -\varepsilon \int \ln \left( \frac{d\mu}{d\xi}(x) \right) d\mu(x).
\]
Next, we compute the term 
\begin{align*}
\iint \exp\left( \frac{f(x) + f(y) - c_\lambda(x,y)}{\varepsilon} \right) d\mu(x) d\mu(y)
&= \iint \exp\left( \frac{f(x)}{\varepsilon} \right) \exp\left( \frac{f(y)}{\varepsilon} \right) k_{\varepsilon,\lambda}(x,y) \, d\mu(x) d\mu(y) \\
&= \iint k_{\varepsilon,\lambda}(x,y) \, d\xi(x) d\xi(y),
\end{align*}
where we use (\ref{ChVar}). Therefore the dual expression \eqref{DualFormProof} becomes:
\[
F_{\varepsilon, \lambda}(\mu) = -\frac{1}{2} \sup_{\xi \in \mathcal{P}(X)} \left\{ -2 \varepsilon \int \ln \left( \frac{d\mu}{d\xi}(x) \right) d\mu(x) - \varepsilon \iint k_{\varepsilon,\lambda}(x,y) \, d\xi(x) d\xi(y) + \varepsilon \right\}.
\]
Dividing both sides by \( \varepsilon \) and rearranging the terms yields: 
\[
\frac{1}{\varepsilon} F_{\varepsilon, \lambda}(\mu) + \frac{1}{2}
= \inf_{\xi \in \mathcal{P}(X)} \left\{ \int \ln \left( \frac{d\mu}{d\xi}(x) \right) d\mu(x) + \frac{1}{2} \iint k_{\varepsilon,\lambda}(x,y) \, d\xi(x) d\xi(y) \right\}.
\]
This completes the proof. 
\end{proof}

\subsection{Proof of Proposition 3} 
\begin{proof}
(i) From Proposition 1, 
the optimal potentials satisfy the fixed-point equations:
\[
\varphi^*(x) = -\varepsilon \ln \int e^{\psi^*(y) - c_\lambda(x, y)/\varepsilon} \, d\nu(y),
\]
\[
\psi^*(y) = -\varepsilon \ln \int e^{\varphi^*(x) - c_\lambda(x, y)/\varepsilon} \, d\mu(x).
\]

Fix \( x, x' \in X \). Using the expression for \( \varphi^* \), we write:
\begin{equation}
|\varphi^*(x) - \varphi^*(x')| = \varepsilon \left| \ln \int e^{\psi^*(y) - c_\lambda(x, y)/\varepsilon} \, d\nu(y) - \ln \int e^{\psi^*(y) - c_\lambda(x', y)/\varepsilon} \, d\nu(y) \right|. \label{intlog}
\end{equation}

By the mean value theorem for the logarithm and the boundedness of \( \psi^* \), we can bound this difference using the Lipschitz continuity of \( c_\lambda \) in its first argument. Clearly, \( c_\lambda \) is uniformly continuous on \( X \times Y \) and Lipschitz, with Lipschitz constant  $L$. 
Then:
$|\varphi^*(x) - \varphi^*(x')| \le L \|x - x'\|.$
A symmetric argument applies to \( \psi^* \), using the analogous expression and the Lipschitz continuity of \( c_\lambda \) in the second variable.
Therefore, both \( \varphi^* \) and \( \psi^* \) are Lipschitz continuous.

(ii) To prove boundedness, note that the integrals inside the logarithms in \eqref{intlog} are bounded above and below due to the boundedness of \( c_\lambda \). Since the exponential of a bounded function is bounded, and the logarithm of a bounded positive function is also bounded, it follows that \( \varphi^*  \in L^\infty(X) \) and $\psi^* \in L^\infty(Y)$.
\end{proof}

\subsection{Proof of Proposition 4} 

\begin{proof}
From Proposition 1, 
the Schrödinger potentials satisfy
$$
\phi(x) = -\varepsilon \ln \int e^{\psi(y) - \frac{1}{\varepsilon} c_\lambda(x,y)} \, d\nu(y),$$ and  $$
\phi_n(x) = -\varepsilon \ln \int e^{\psi_n(y) - \frac{1}{\varepsilon} c_\lambda(x,y)} \, d\nu_n(y).
$$ 
Looking at the integrand, let us define the functions 
$$f(x,y) := e^{\psi(y) - \frac{1}{\varepsilon} c_\lambda(x,y)},$$ and 
$$f_n(x,y) := e^{\psi_n(y) - \frac{1}{\varepsilon} c_\lambda(x,y)}.$$ 
We first establish some regularity properties of these functions. Since \( \psi \) and \( \psi_n \) are uniformly Lipschitz and bounded (by Proposition 3)
, and \( c_\lambda \) is continuous and bounded, it follows that for each \( x \in X \), the maps \( y \mapsto f(x,y) \) and \( y \mapsto f_n(x,y) \) are uniformly bounded and equicontinuous. The exponential of a bounded Lipschitz function is again bounded and Lipschitz, so both \( f(x,\cdot) \) and \( f_n(x,\cdot) \) are uniformly Lipschitz in \( y \), uniformly over \( x \). Next, consider the function class $
\mathcal{F} := \left\{ f_x(y) := f(x,y) \mid x \in X \right\}.$
This class is uniformly bounded, uniformly Lipschitz in \( y \), and indexed by the compact set \( X \subset \mathbb{R}^d \). As a result, it has finite uniform entropy\footnote{In this context, finite uniform entropy means that the class \( \mathcal{F} = \{ f_x(y) := f(x,y) \mid x \in X \} \), where each \( f_x \) is Lipschitz and bounded on a compact domain, can be covered by finitely many functions in the \( L^2(P) \) norm, uniformly over all probability measures \( P \). Here, the measures \( P \) are Borel probability measures on the space \( X \). This ensures that the class  supports uniform convergence of empirical integrals. See \cite[Corollary~2.7.10]{vandervaart1996weak}.} and is a Glivenko–Cantelli class (see \cite[Section 2.4, Theorem 2.4.1]{vandervaart1996weak}). Therefore:
\[
\sup_{f \in \mathcal{F}} \left| \int f(y) \, d\nu_n(y) - \int f(y) \, d\nu(y) \right| \to 0.
\]

This supremum over \( \mathcal{F} \) can be rewritten as a supremum over \( x \in X \). Indeed, each \( f \in \mathcal{F} \) is of the form \( f_x(y) = f(x,y) \) for some \( x \in X \), and the map \( x \mapsto f_x \) is continuous in the uniform topology on \( \mathcal{C}(X) \), due to the regularity of \( \psi \) and \( c_\lambda \). Moreover, the class \( \mathcal{F} \) is uniformly bounded and equicontinuous, and \( X \) is compact. Hence, we can equivalently write:
\[
\sup_{f \in \mathcal{F}} \left| \int f(y) \, d\nu_n(y) - \int f(y) \, d\nu(y) \right|
= \sup_{x \in X} \left| \int f(x,y) \, d\nu_n(y) - \int f(x,y) \, d\nu(y) \right|.
\]

To incorporate the approximating functions \( f_n(x,y) \), we observe that \( f_n(x,y) \to f(x,y) \) pointwise as \( n \to \infty \), due to pointwise convergence \( \psi_n(y) \to \psi(y) \) and continuity of \( c_\lambda \). As established earlier, both \( f_n(x,\cdot) \) and \( f(x,\cdot) \) are uniformly bounded and equicontinuous in \( y \), uniformly in \( x \), so the family \( \{ f_n(x,\cdot) \} \) is uniformly integrable. An application the triangle inequality yields
\begin{eqnarray}
\left| \int f_n(x,y) \, d\nu_n(y) - \int f(x,y) \, d\nu(y) \right| \leq 
\left| \int f_n(x,y) \, d\nu_n(y) - \int f(x,y) \, d\nu_n(y) \right| \nonumber \\ +
\left| \int f(x,y) \, d\nu_n(y) - \int f(x,y) \, d\nu(y) \right|.
\end{eqnarray}

The second term vanishes uniformly in \( x \) by the Glivenko–Cantelli result. For the first term, we note that for each fixed \( x \in X \), the sequence \( f_n(x,\cdot) \to f(x,\cdot) \) converges pointwise in \( y \), and the functions are uniformly bounded and equicontinuous in \( y \), uniformly over \( x \). Therefore, by  the dominated convergence theorem, we obtain:
\[
\sup_{x \in X} \left| \int f_n(x,y) \, d\nu_n(y) - \int f(x,y) \, d\nu_n(y) \right| \to 0.
\]
Since the logarithm is Lipschitz on compact subsets of \( (0,\infty) \), and the integrals of \( f_n(x,\cdot) \) and \( f(x,\cdot) \) are bounded away from zero and infinity, we conclude $
\sup_{x \in X} |\phi_n(x) - \phi(x)| \to 0.$

The same argument applies symmetrically to \( \psi_n(y) \) and \( \psi(y) \) using the fixed-point equations:
\[
\psi(y) = -\varepsilon \ln \int e^{\phi(x) - \frac{1}{\varepsilon} c_\lambda(x,y)} d\mu(x), \quad
\psi_n(y) = -\varepsilon \ln \int e^{\phi_n(x) - \frac{1}{\varepsilon} c_\lambda(x,y)} d\mu_n(x).
\]
Hence, both potentials converge uniformly.
\end{proof}

\subsection{Proof of Proposition 5} 
 \begin{proof}
 (i) For a fix $\lambda$, the associated $c_\lambda$ is   continuous in $(x,y)$ and bounded.
Lemma 5.3 in \cite{N22} implies that, given \( \eta > 0 \) and \( \pi_0 \in \Pi(\mu,\nu) \), there exists \( \tilde{\pi} \in \Pi(\mu,\nu) \) such that
$\left|
\int c_\lambda \, d\tilde{\pi} - \int c_\lambda \, d\pi_0 \right| \leq \eta$
and \( \frac{d\tilde{\pi}}{d(\mu \otimes \nu)} \) is bounded.
So, 
$\int c_\lambda \, d\tilde{\pi}  \leq  C_0 + \eta$ and  $ H(\tilde{\pi} \| \mu \otimes \nu) < \infty.
$ Thus, 
$$
  C_0   \leq \int c_\lambda \, d\tilde\pi +   \varepsilon H (\tilde\pi \vert P) \leq  C_0 + \eta + \varepsilon H (\tilde\pi \vert P)
$$
and (16) follows from the arbitrariness  of $\eta>0$.

(ii) Since, $c_\lambda$ is lower semicontinuous, the statement follows from Proposition 5.9 in \cite{N22}.
 \end{proof}

\subsection{Proof of Proposition 6} 
\begin{proof}
Since \( X \) is compact and \( c_\lambda \) is continuous and bounded on \( X \times Y \), the entropic cost 
$C_\varepsilon(\mu, \nu, c_\lambda, \pi) = \int c_\lambda(x,y) \, d\pi(x,y) + \varepsilon H(\pi \| \mu \otimes \nu)$
is well-defined and finite by assumption. Moreover, it is lower semicontinuous with respect to weak convergence of \( \pi \), meaning that if \( \pi_n \to \pi \) weakly, then
\[
\liminf_{n \to \infty} C_\varepsilon(\mu_n, \nu_n, c_\lambda, \pi_n) \geq C_\varepsilon(\mu, \nu, \lambda, \pi).
\]
This ensures that the cost of the limiting plan \( \pi \) is not smaller than the limiting cost of the approximating sequence. In particular, it guarantees that any weak limit of a minimizing sequence remains a valid candidate minimizer for the limiting problem. Additionally, the functional is strictly convex in \( \pi \) for fixed \( \varepsilon > 0 \), due to the strict convexity of the relative entropy term. This guarantees uniqueness of the minimizer \( \pi^* \), which is crucial for concluding convergence of the entire sequence rather than just a subsequence. Moreover, by Proposition~4, 
the Schrödinger potentials \( \varphi_n^*, \psi_n^* \) associated with \( \pi_n^* \) converge uniformly to \( \varphi^*, \psi^* \), the potentials associated with \( \pi^* \). This implies that the densities
\[
\frac{d\pi_n^*}{d(\mu_n \otimes \nu_n)}(x,y) = \exp\left( \varphi_n^*(x) + \psi_n^*(y) - \frac{1}{\varepsilon} c_\lambda(x,y) \right)
\]
converge uniformly to the density of \( \pi^* \) with respect to \( \mu \otimes \nu \). Since \( \pi_n^* \in \mathcal{P}(X \times Y) \) and \( X,Y \) are compact, the sequence \( \{\pi_n^*\} \) is tight. By Prokhorov’s theorem, it admits a weakly convergent subsequence. Let \( \pi_\infty \) be any such limit. Since the marginals \( \mu_n \to \mu \) and \( \nu_n \to \nu \) weakly, and the potentials converge uniformly, the limit \( \pi_\infty \) must minimize the entropic cost for \( \mu, \nu \). By uniqueness of the minimizer due to strict convexity, we conclude \( \pi_\infty = \pi^* \). Therefore, every subsequence of \( \{\pi_n^*\} \) has a further subsequence converging to \( \pi^* \), which implies that the full sequence converges:
$\pi_n^* \to  \pi^* $ weakly, as  $n \to \infty.$
\end{proof}

\subsection{Proof of Proposition 7}
\begin{proof} Lemma \ref{LemmaPosUniv} shows  that the  Lipschitz function $c_\lambda$ induces,
for $\varepsilon > 0$, a positive $c$-universal kernel $k_{\varepsilon, \lambda}(x,y)$. Moreover,  the dual expression in Eq. (14) of the main text 
as a maximization of linear forms ensures that  $W_{\varepsilon,\lambda}(\mu, \nu)$ is convex with respect to $\mu$ and with respect to $\nu$ (but not jointly convex if $\varepsilon > 0$). Thus, Proposition  3 and Proposition 4  in \cite{feydy2019} imply that $\overline{W}_{\varepsilon,\lambda}(\mu, \nu) $ is convex with respect to both inputs. Statements (i), (ii), and (iii) follow from Theorem 1 in 
 \cite{feydy2019}, while (iv) follows from our Proposition 5. 
 \end{proof}

\subsection{Proof of Theorem 9} 
\begin{proof}
We aim to bound the quantity:
$
\mathbb{E}\left[ \left| \overline{W}_{\varepsilon,\lambda}(\mu_n, \nu_n) - \overline{W}_{\varepsilon,\lambda}(\mu, \nu) \right| \right]. 
$
To this end, we decompose it as:
\begin{equation}
\left| \overline{W}_{\varepsilon,\lambda}(\mu_n, \nu_n) - \overline{W}_{\varepsilon,\lambda}(\mu, \nu) \right| \leq \left| W_{\varepsilon,\lambda}(\mu_n, \nu_n) - W_{\varepsilon,\lambda}(\mu, \nu) \right| + \frac{1}{2} \left| \Delta_n^\mu + \Delta_n^\nu \right| \label{Decomp}
\end{equation}
where
$
\Delta_n^\mu := W_{\varepsilon,\lambda}(\mu_n, \mu_n) - W_{\varepsilon,\lambda}(\mu, \mu),$ and $
\Delta_n^\nu := W_{\varepsilon,\lambda}(\nu_n, \nu_n) - W_{\varepsilon,\lambda}(\nu, \nu).$

Let $\varphi^*, \psi^*$ be the optimal Schrödinger potentials for $W_{\varepsilon,\lambda}(\mu, \nu)$. These are bounded, Lipschitz functions due to the regularity of the cost and compactness of the domain---see Proposition  3. 
Define the suboptimal estimator:
\[
\widehat{W}_{\varepsilon,\lambda}(\mu_n, \nu_n) := \int \varphi^* \, d\mu_n + \int \psi^* \, d\nu_n.
\]
Then, we consider the identity:
\[
 \int \varphi^*_n \, d\mu_n + \int \psi^*_n \, d\nu_n=:W_{\varepsilon,\lambda}(\mu_n, \nu_n) = \widehat{W}_{\varepsilon,\lambda}(\mu_n, \nu_n) + \text{bias}_n
\]
where 
$$\text{bias}_n :=  W_{\varepsilon,\lambda}(\mu_n, \nu_n) - \widehat{W}_{\varepsilon,\lambda}(\mu_n, \nu_n) = \int (\varphi_n^* - \varphi^*) \, d\mu_n + \int (\psi_n^* - \psi^*) \, d\nu_n.
$$ From uniform convergence of Schrödinger potentials stated in Proposition 4, 
we have
$
  \sup_{x \in X} |\varphi_n^*(x) - \varphi^*(x)| = \mathcal{O}(n^{-1/2})$ and $  \sup_{y \in X} |\psi_n^*(y) - \psi^*(y)| = \mathcal{O}(n^{-1/2})$,
and since $\mu_n, \nu_n$ are probability measures, we have the upper bound:
\[
\mathbb{E} |\text{bias}_n| \leq  \sup_{x} |\varphi_n^*(x) - \varphi^*(x)| + \sup_{y} |\psi_n^*(y) - \psi^*(y)| = \mathcal{O}(n^{-1/2}).
\]
Therefore:
\[
\mathbb{E} \left| \widehat{W}_{\varepsilon,\lambda}(\mu_n, \nu_n) - W_{\varepsilon,\lambda}(\mu, \nu) \right| \leq  \mathbb{E} \left| \int \varphi^* \, d(\mu_n - \mu) \right| + \mathbb{E} \left| \int \psi^* \, d(\nu_n - \nu) \right| + \mathcal{O}(n^{-1/2})
\]

Since \( \varphi^*, \psi^* \) are Lipschitz, we apply bounded-Lipschitz norm duality:
\begin{equation}
\mathbb{E} \left| \int \varphi^* \, d\mu_n - \int \varphi^* \, d\mu \right| \le \|\varphi^*\|_{\mathrm{Lip}} \mathbb{E} \|\mu_n - \mu\|_{\mathrm{BL}},
\quad
\left| \int \psi^* \, d\nu_n - \int \psi^* \, d\nu \right| \le \|\psi^*\|_{\mathrm{Lip}} \mathbb{E} \|\nu_n - \nu\|_{\mathrm{BL}}, \label{LipB}
\end{equation}
where  the {Lipschitz norm} of $\varphi^*$ is defined as:
\[
\|\varphi^*\|_{\mathrm{Lip}} := \sup_{x \neq y} \frac{|\varphi^*(x) - \varphi^*(y)|}{d(x, y)},
\]
and similarly for $\psi^*$.
From empirical process theory (see e.g. Section 2.1.4, p.~91; 2.2, pp.~95--104; and  Section 2.5.1, p.~127] in \cite{vandervaart1996weak}), we have:
\begin{equation}
\mathbb{E}[\|\mu_n - \mu\|_{\mathrm{BL}}] = \mathcal{O}(n^{-1/2}), \quad \mathbb{E}[\|\nu_n - \nu\|_{\mathrm{BL}}] = \mathcal{O}(n^{-1/2}). \label{BLbounds}
\end{equation}
Thus, from \eqref{LipB} and \eqref{BLbounds}, we have
\[
\mathbb{E}\left[ \left| \int \varphi^* \, d(\mu_n - \mu) \right| \right] = \mathcal{O}(n^{-1/2}), \quad
\mathbb{E}\left[ \left| \int \psi^* \, d(\nu_n - \nu) \right| \right] = \mathcal{O}(n^{-1/2}).
\]
Hence:
\[
\mathbb{E}\left[ \left| W_{\varepsilon,\lambda}(\mu_n, \nu_n) - W_{\varepsilon,\lambda}(\mu, \nu) \right| \right] = \mathcal{O}(n^{-1/2}).
\]

Similar arguments allow to state:
$\mathbb{E}\left[ \left| \Delta_n^\mu \right| \right] = \mathcal{O}(n^{-1/2}),$ and $\mathbb{E}\left[ \left| \Delta_n^\nu \right| \right] = \mathcal{O}(n^{-1/2}),
$
so, the triangle inequality implies that 
$\mathbb{E}\left[ \left| \Delta_n^\mu + \Delta_n^\nu \right| \right] \leq \mathbb{E}\left[ \left| \Delta_n^\mu \right| \right] + \mathbb{E}\left[ \left| \Delta_n^\nu \right| \right] = \mathcal{O}(n^{-1/2})$.
Combining the results into \eqref{Decomp} yields
$
\mathbb{E}\left[ \left| \overline{W}_{\varepsilon,\lambda}(\mu_n, \nu_n) - \overline{W}_{\varepsilon,\lambda}(\mu, \nu) \right| \right] = \mathcal{O}(n^{-1/2}),
$
as claimed.
\end{proof}

\subsection{Proof of Corollary 10} 
\begin{proof}
Let $\mu_n$ and $\nu_n$ be two independent empirical measures, each based on $n$ i.i.d.\ samples from $\mu$.  We aim to bound $\mathbb{E}\left[ \overline{W}_{\varepsilon,\lambda}(\mu_n, \mu) \right]$, where the expected value is taken w.r.t. distribution of  $\mu_n$. The Sinkhorn loss \( \overline{W}_{\varepsilon,\lambda} \) is convex in each argument (Proposition~6). Fixing \( \mu_n \), the map \( \nu \mapsto \overline{W}_{\varepsilon,\lambda}(\mu_n, \nu) \) is convex.
Since \( \mathbb{E}[\nu_n] = \mu \), so $\overline{W}_{\varepsilon,\lambda}(\mu_n, \mathbb{E}[\nu_n] ) = \overline{W}_{\varepsilon,\lambda}(\mu_n, \mu)$. Then, Jensen's inequality implies
$\overline{W}_{\varepsilon,\lambda}(\mu_n, \mu) \leq \mathbb{E}\left[ \overline{W}_{\varepsilon,\lambda}(\mu_n, \nu_n) \right]$. Taking expectation over \( \mu_n \) yields 
\begin{eqnarray*}
\mathbb{E}_{\mu_n}\left[ \overline{W}_{\varepsilon,\lambda}(\mu_n, \mu) \right] \leq 
\mathbb{E} \left[ \overline{W}_{\varepsilon,\lambda}(\mu_n, \nu_n) \right] &\leq& \mathbb{E}  \left[ \overline{W}_{\varepsilon,\lambda}(\mu_n, \nu_n) \right]  -  \mathbb{E} \left[ \overline{W}_{\varepsilon,\lambda}(\mu, \mu) \right] \\
&\leq& \mathbb{E}\left[ \left| \overline{W}_{\varepsilon,\lambda}(\mu_n, \nu_n) - \overline{W}_{\varepsilon,\lambda}(\mu, \mu) \right| \right] 
\end{eqnarray*}
where we make use of $ \overline{W}_{\varepsilon,\lambda}(\mu, \mu)  =0$. 
By Theorem~10, we have
$$
 \mathbb{E}\left[ \left| \overline{W}_{\varepsilon,\lambda}(\mu_n, \nu_n) - \overline{W}_{\varepsilon,\lambda}(\mu, \mu) \right| \right] = \mathcal{O}\left(  {n}^{-1/2} \right),
$$
so,
$$
\mathbb{E}_{\mu_n}\left[ \overline{W}_{\varepsilon,\lambda}(\mu_n, \mu) \right] =   \mathcal{O}\left(  {n}^{-1/2} \right).
$$
\end{proof}

\subsection{Proof of Proposition 11}
\begin{proof} 
	The proof is an immediate application of Proposition 5.2 from \cite{RigolletWeeds2018}. We simply need to verify that the E-ROBOT setup fits their general framework.
The generative model is $Y = X + Z$, where the noise $Z$ has a known density $f$ with respect to the Lebesgue measure. In the E-ROBOT case, this density is defined by the truncated cost function:
\[
f(z) = \frac{1}{\beta} \exp\left( -\frac{1}{\varepsilon} c_{\lambda}(0, z) \right),
\]
where $\beta = \int \exp\left( -\frac{1}{\varepsilon} c_{\lambda}(0, z) \right) dz$ is the normalization constant. This is a {truncated Laplace distribution}, a well-defined probability density function because $c_\lambda$ is bounded and continuous. From  \cite{RigolletWeeds2018}, it follows  for this noise model, we have:
\begin{align*}
W_f(\mu, \nu) &:= \min_{\gamma \in \Pi(\mu, \nu)} \left\{ -\int \ln f(x - y)  d\gamma(x, y) + H(\gamma\Vert \mu \otimes \nu) \right\} \\
&= \min_{\gamma \in \Pi(\mu, \nu)} \left\{ \int \left( \frac{1}{\varepsilon} c_{\lambda}(x, y) + \ln \beta \right)  d\gamma(x, y) + H(\gamma\Vert \mu  \otimes \nu)  \right\} \\
&= \frac{\ln \beta}{\varepsilon} + \frac{1}{\varepsilon} \min_{\gamma \in \Pi(\mu, \nu)} \left\{ \int c_{\lambda}(x, y)  d\gamma(x, y) + \varepsilon H(\gamma\Vert \mu  \otimes \nu)  \right\} \\
&= C + \frac{1}{\varepsilon} W_{\varepsilon, \lambda}(\mu, \nu),
\end{align*}
where $C = \ln \beta$ is a constant independent of $\mu$ and $\nu$. Since $C$ and the factor $1/\varepsilon$ are constants with respect to the minimization over $\mu \in \mathcal{P}$, we have:
\[
\arg \min_{\mu \in \mathcal{P}} W_f(\mu, \nu_n) = \arg \min_{\mu \in \mathcal{P}} W_{\varepsilon, \lambda}(\mu, \nu_n).
\]
From Proposition 5.2 in \cite{RigolletWeeds2018}, under the specified generative model, the maximum-likelihood estimator is such that:
\[
\hat{\mu}_n = \arg \min_{\mu \in \mathcal{P}} W_f(\mu, \nu_n). = \arg \min_{\mu \in \mathcal{P}} W_{\varepsilon, \lambda}(\mu, \nu_n).
\]
This completes the proof.
\end{proof}

\section{Additional numerical exercises} \label{AppMMDs}

\begin{figure}[!h]
     \centering
     \begin{subfigure}
         \centering
         \includegraphics[width=0.97\textwidth, height=0.6\textwidth]{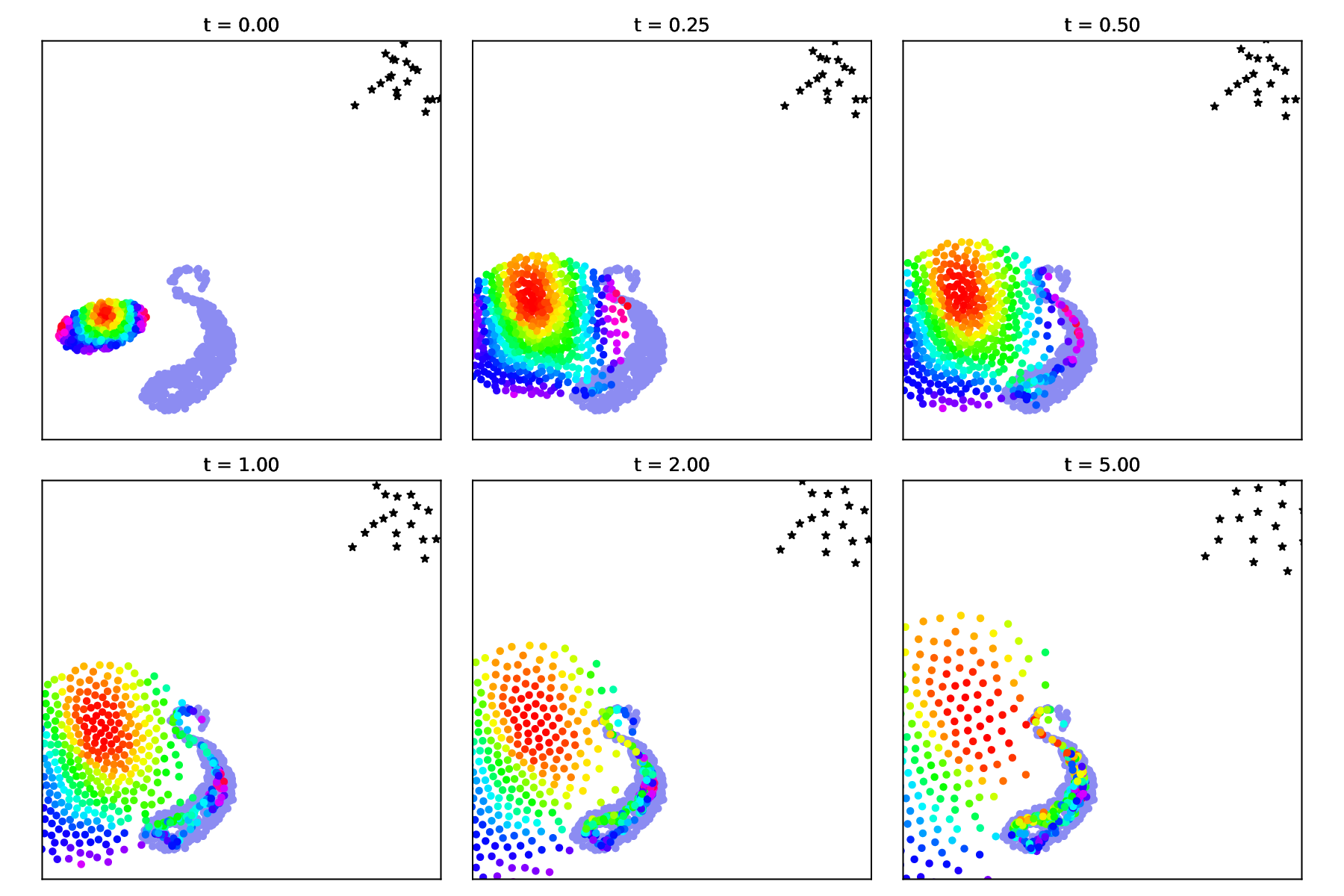}
     \end{subfigure} \\
     \begin{tabular}{cc}
         \includegraphics[width=0.45\textwidth, height=0.35\textwidth]{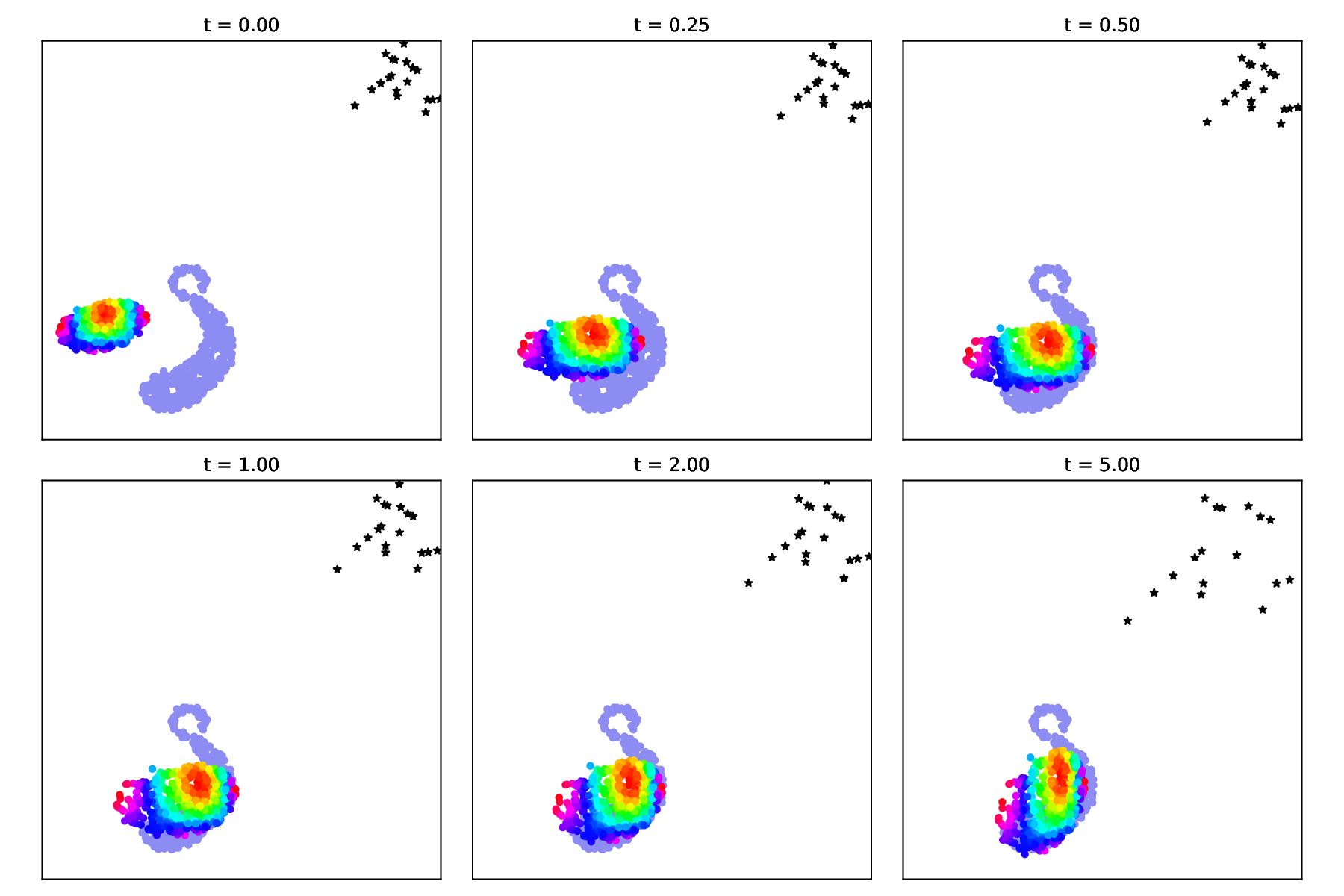} &
         \includegraphics[width=0.45\textwidth, height=0.35\textwidth]{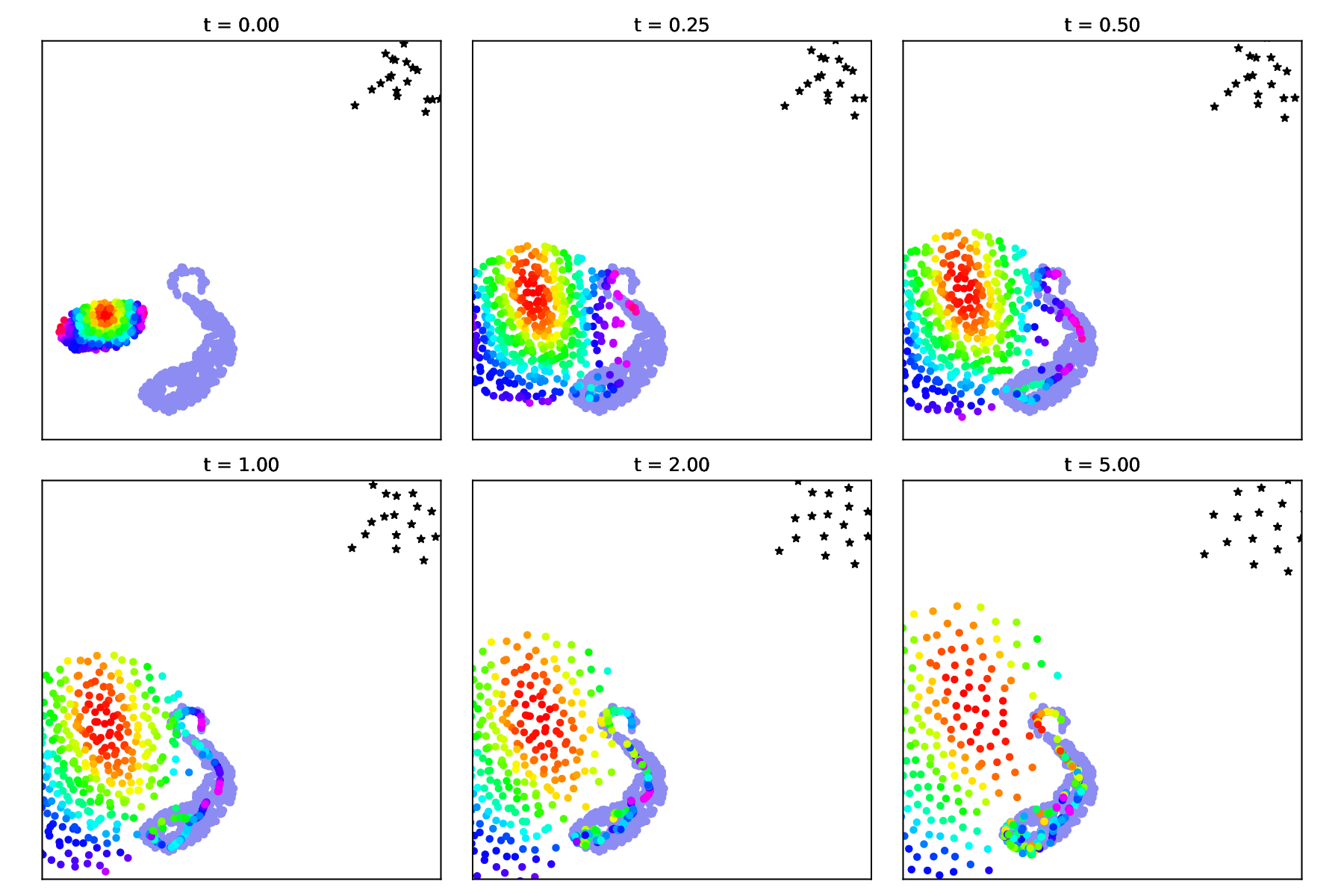}
     \end{tabular}        
           \caption{Gradient flows for 2D shapes via MMDs with different kernels. Top  panels:  MMD with kernel $k_{\lambda,\varepsilon}$, $\lambda=0.6$, $\varepsilon=0.05$ and learning rate, 
           i.e. time step $\tau$, is set equal to $0.05$. Bottom  
            left 6 panels: MMD  with Gaussian kernel having $\sigma=0.65$. Bottom right 6 panels: 
        MMD  with Gaussian kernel having $\sigma=0.05$.}
       \label{FigFlowsMMD}
\end{figure}

\cite{arbel2019maximum} propose the use of MMD for gradient flow. Therefore, we repeat the above exercise using three different MMDs:
one is obtained using the kernel $k_{\lambda,\varepsilon}$  and two are obtained using a Gaussian kernel, with small and large variance ($\sigma=0.05$ and $\sigma=0.65$, respectively). We display the results in Figure 
 \ref{FigFlowsMMD}.  Comparing the 6 top plots in Figure 6 of the paper 
 to those obtained via MMDs, we notice that 
the methods slow down the movement of some  points far from the target distribution. This is due to the fact that at these points
 the kernel values become negligible and this, in turn, implies
that the gradients vanish. As a consequence, there is a region in space where the gradient signals are strong enough to move particles, whereas in other regions there are so weak and they entail no movements.  Because of these considerations, we conclude that use of $\overline{W}_{\lambda,\varepsilon}$ is preferable to the MMDs for gradient flow: it provides a more reliable, more geometry aware flow and more robust gradient signal for moving probability distributions toward each other, especially when they are initially far apart and contain outliers.

\begin{figure}[h!]
\centering
\includegraphics[width=0.8\textwidth, height=0.6\textwidth]{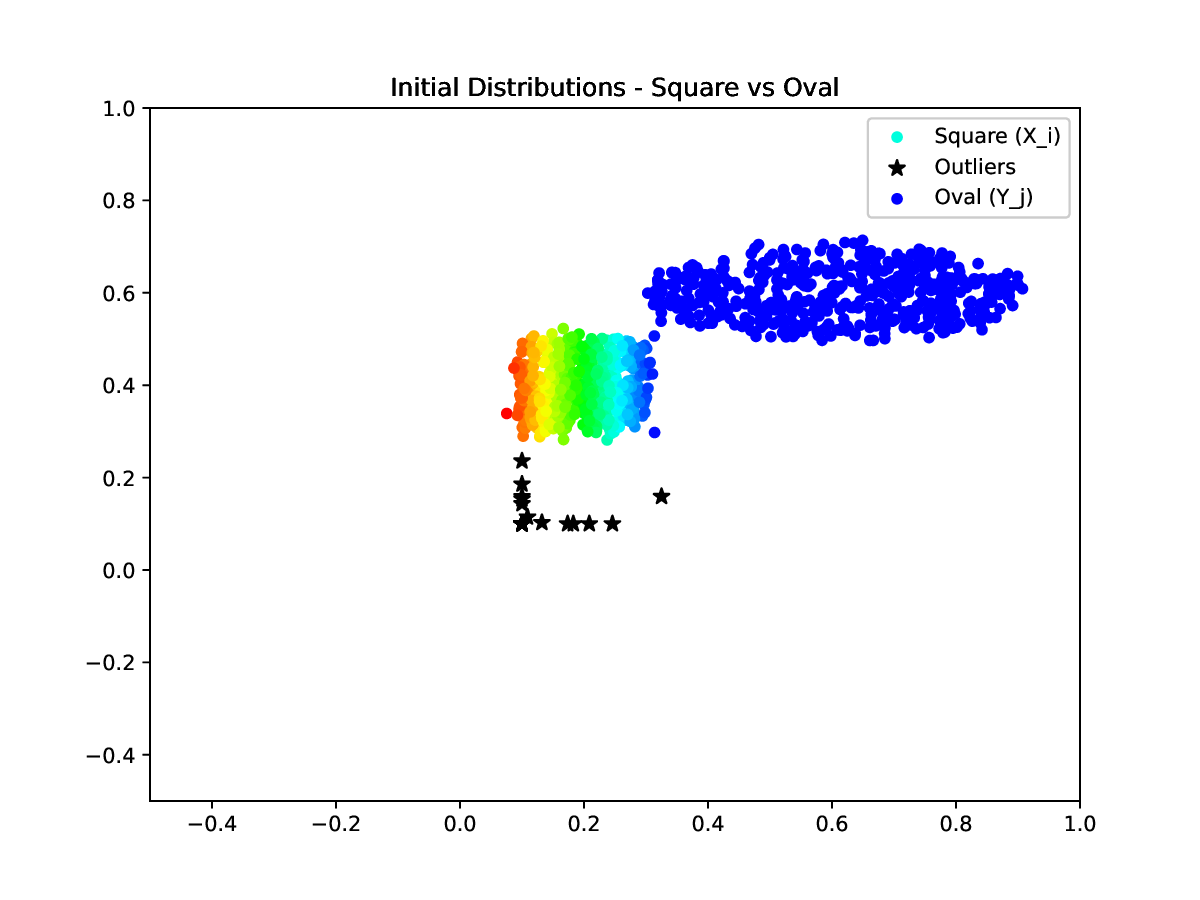}
\caption{Square (source shape) and an oval (target shape), in the presence of outlying values.}
 \label{FigShapes1}
\end{figure}

Different numerical experiments made us understand that the performance of MMDs depends on many aspects of the numerical design, like the type of shapes, their overlapping, and the position of outliers.  To elaborate further, we illustrate that MMD-based gradient flow depends on the positions on both  the underlying shapes and on the locations of outliers. To this end, we consider the gradient flow between square (source shape) and an oval (target shape), in the presence of outlying values, as depicted in Figure \ref{FigShapes1}.

\begin{figure}[h!]
     \begin{tabular}{cccc}
         \includegraphics[width=0.475\textwidth, height=0.45\textwidth]{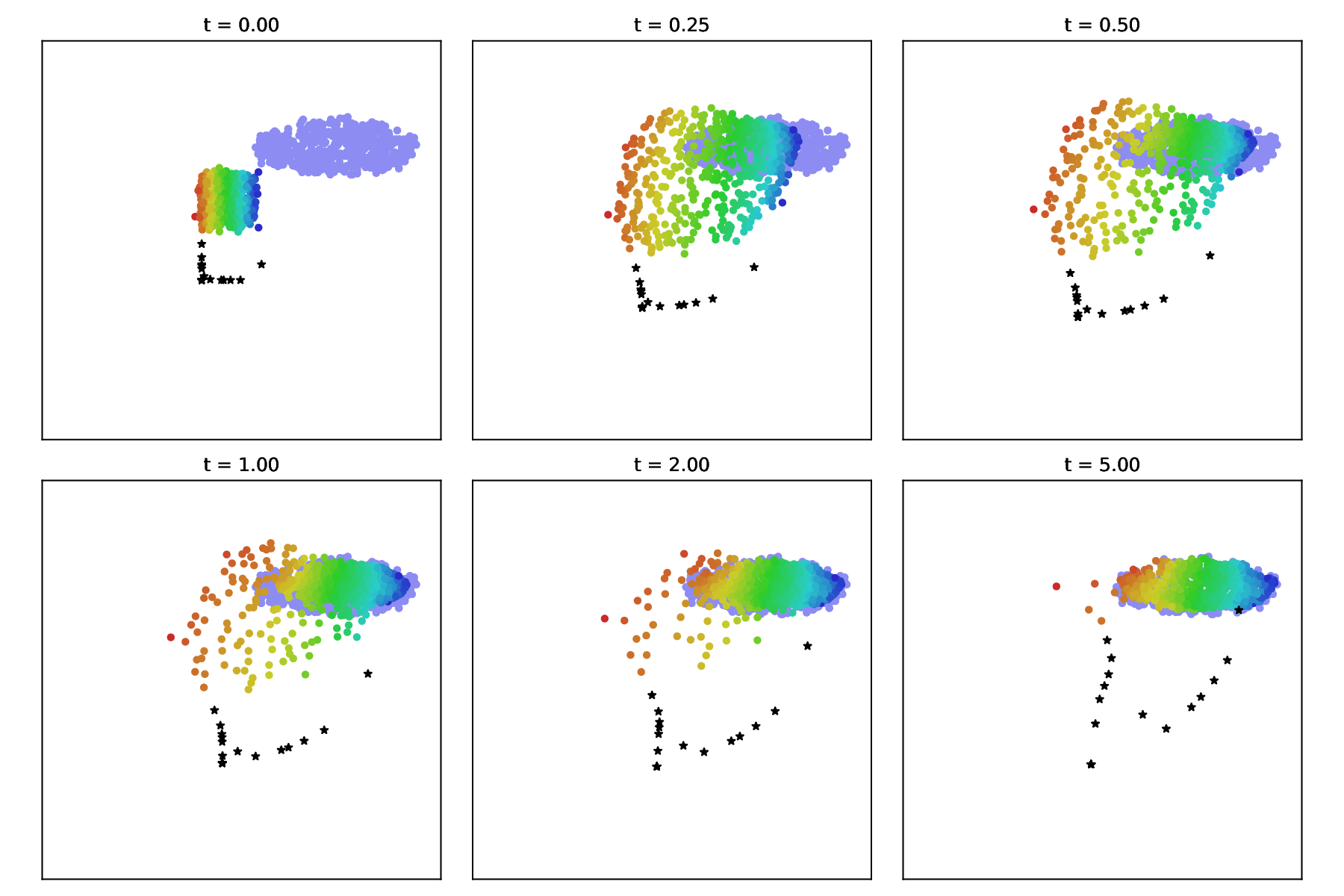} &
         \includegraphics[width=0.475\textwidth, height=0.45\textwidth]{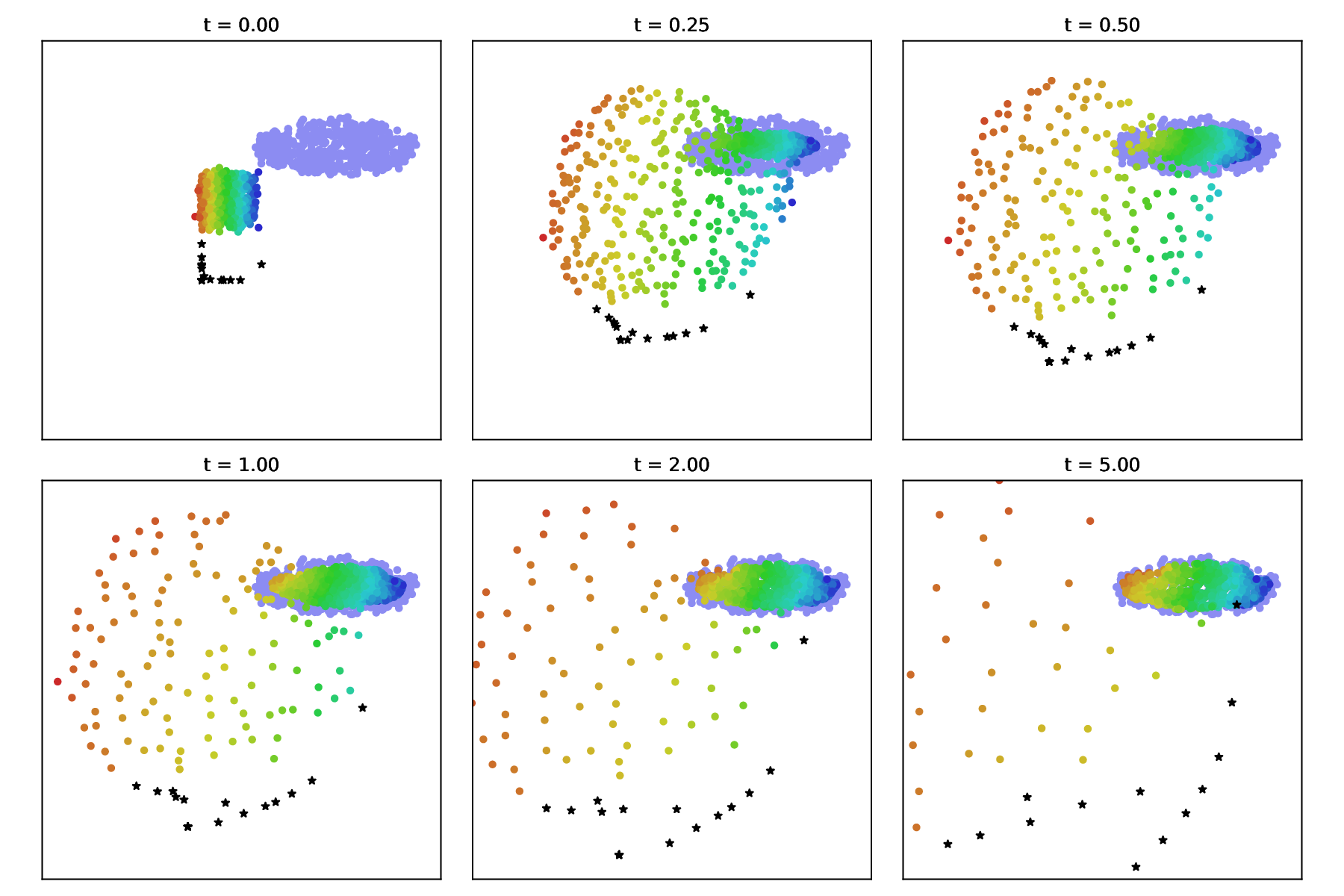} \\
         \includegraphics[width=0.475\textwidth, height=0.45\textwidth]{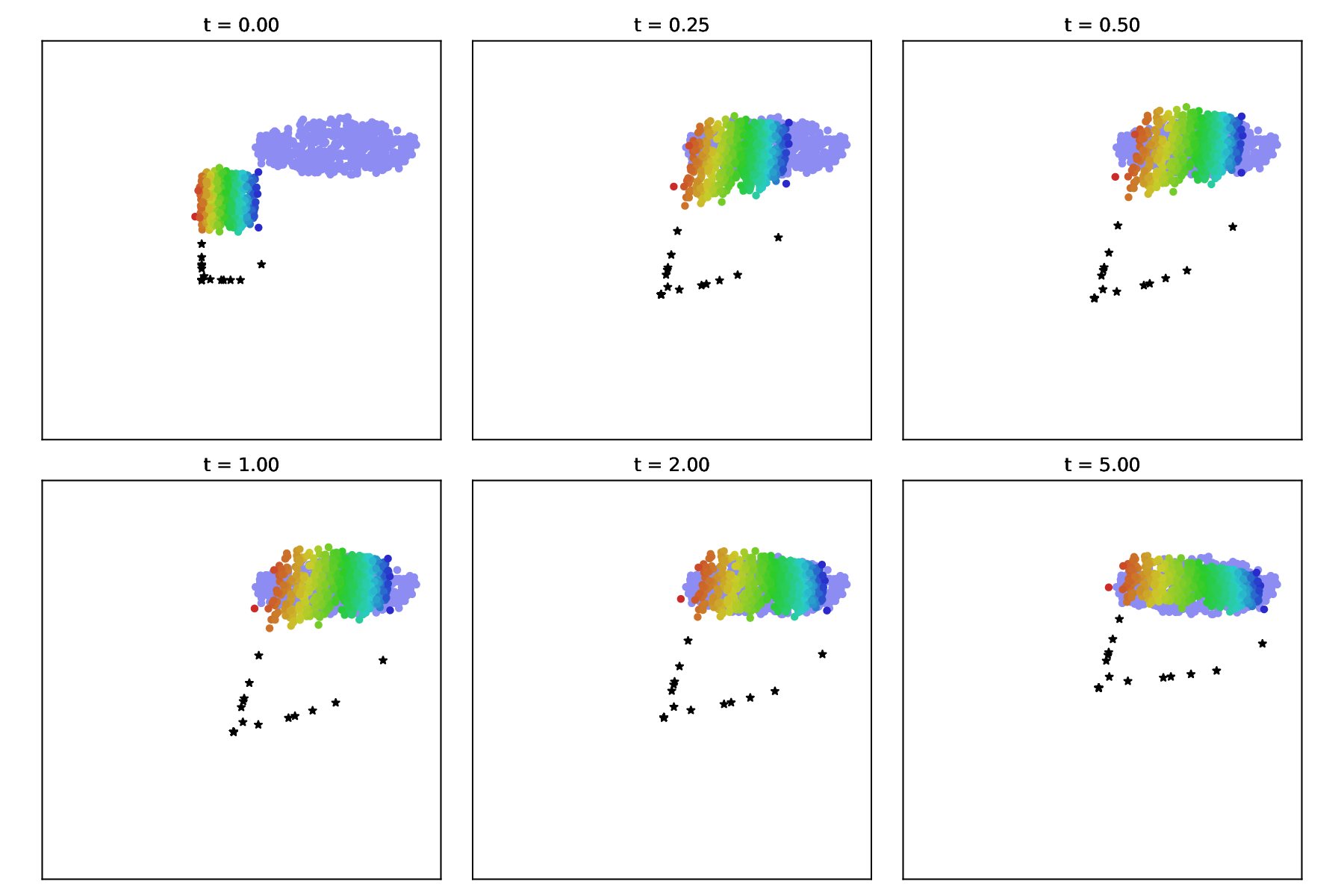} &
         \includegraphics[width=0.475\textwidth, height=0.45\textwidth]{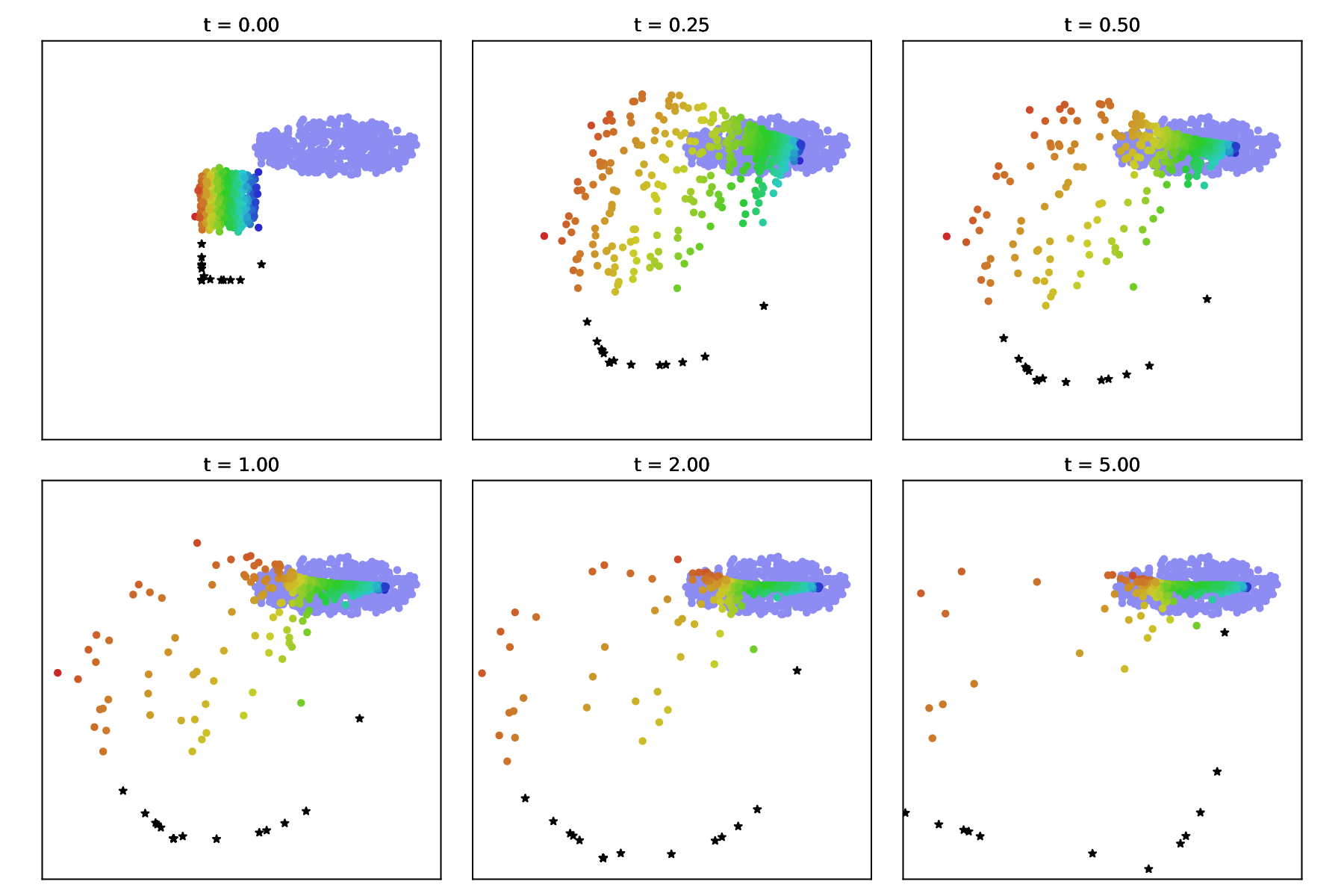} \\
     \end{tabular}        
           \caption{Gradient flows for 2D shapes via MMDs with different kernels. Top left  panels:  MMD with kernel $k_{\lambda,\varepsilon}$, $\lambda=4$, $\varepsilon=1$. Top right panels:  MMD with kernel $k_{\lambda,\varepsilon}$, $\lambda=4$, $\varepsilon=0.25$. Bottom  
            left 6 panels: MMD  with Gaussian kernel having $\sigma=0.25$. Bottom left 6 panels: 
        MMD  with Gaussian kernel having $\sigma=0.55$. Bottom left 6 panels: 
        MMD  with Gaussian kernel having $\sigma=0.25$. For all plots the  learning rate in the gradient flow is 
         $\tau = 0.05$}
       \label{FigFlowsMMD2}
\end{figure}

The numerical experiments in Figure \ref{FigFlowsMMD2} provide a nuanced view of how kernel choice and parameters influence gradient flow performance. The top row demonstrates the effect of the scale parameter $\varepsilon$  within the truncated Laplace MMD ($k_{\lambda,\varepsilon}$). A larger $\varepsilon$ (e.g., $\varepsilon=1$, top left) increases the smoothing effect, leading to a more diffuse and stable but potentially less precise flow. Crucially, the flow for the Laplace kernel with a larger scale ($\varepsilon=1$) demonstrates the most effective overall performance: it successfully merges the central cloud of points into a coherent oval and, despite it transports one outlier star, meaningfully it slowly transports the other outliers towards the target. This effective regularization highlights a potential advantage of this kernel's structure.

The comparison with the Gaussian MMD (bottom row) reveals a more fundamental sensitivity. The Gaussian kernel with a larger bandwidth ($\sigma=0.55$, bottom left) performs reasonably by preventing vanishing gradients, but it fails to fully resolve the target shape. The flow lacks the necessary precision to fully contract all rainbow square points into a tight oval, and consequently, it also fails to fully integrate the outliers, leaving them stranded. The result is a blurred and incomplete registration. The Gaussian with a smaller bandwidth ($\sigma=0.25$) performs worse, as expected, with severe vanishing gradients stalling the flow for distant points.

These different behaviours underscore a key potential advantage of the truncated Laplace kernel: its parameters $\lambda$ (robustness) and $\varepsilon$ (scale) offer a more interpretable and effective mechanism for balancing smoothness with precision. The Laplace kernel's built-in robustness, which bounds the influence of distant outliers, often makes it a more reliable and easier-to-tune choice than the Gaussian kernel for gradient flow applications, as it provides a more uniform and effective gradient signal across the space.

\end{appendices}

\newpage
\bibliography{biblio}